\documentclass[12pt,table,xcdraw]{article}
\usepackage{amsmath}
\usepackage{amsfonts}
\usepackage[letterpaper]{geometry}
\usepackage{natbib}

\usepackage{float}
\usepackage{caption}
\usepackage{subcaption}

\usepackage{url} 
\usepackage{hyperref}
\usepackage{booktabs} 
\usepackage{xcolor}
\usepackage{graphicx}
\usepackage{multirow}
\usepackage{enumitem}
\usepackage{tikz}
\usetikzlibrary{shapes, positioning, calc, fit, backgrounds}

\usepackage{epstopdf}

\usepackage[normalem]{ulem}
\newcommand{\blind}{1}

\useunder{\uline}{\ul}{}
\newcommand{\bm}[1]{\mathbf{#1}}
\newcommand{\btheta}{\mathbf{\theta}}

\begin{document}

\if1\blind
{
  \title{\bf From Small to Large Language Models:\\ Revisiting the Federalist Papers}
  \author{So Won Jeong\thanks{2nd year doctoral student at the Booth School of Business of the University of Chicago. \hspace{5cm}
  This review paper was completed to fulfill the ``summer paper"   requirement for $1^{st}$ year doctoral students at Booth.  It serves as supplementary reading for students enrolled in  the Big Data course  at Booth as it illustrates several techniques taught in this class.} \,\, and Veronika Rockova \thanks{Bruce Lindsay Professor of Econometrics and Statistics at the Booth School of Business of the University of Chicago}}
  \maketitle
} \fi

\if0\blind
{
  \bigskip
  \bigskip
  \bigskip
  \begin{center}
    {\LARGE\bf From Small to Large Language Models:\\ Revisiting the Federalist Papers}
\end{center}
  \medskip
} \fi

\bigskip

\begin{abstract}
For a long time, the authorship of the Federalist Papers had been  a subject of  inquiry and debate,  not only by linguists and historians but also by statisticians. In what was arguably the first Bayesian case study, \cite{Mosteller1963Inference} provided the first statistical evidence for attributing all disputed papers to Madison. 
Our paper revisits this historical dataset but from a lens of modern language models, both small and large.
We review some of the more popular Large Language Model (LLM) tools and examine them from a statistical point of view  in the context of text classification.
We investigate whether, without any attempt to fine-tune, the general embedding constructs
can be useful for stylometry and attribution. We explain differences between various word/phrase
embeddings and discuss how to aggregate them in a document. 
Contrary to our expectations, we exemplify that dimension {\em expansion} with word embeddings may not always be beneficial for attribution relative to dimension {\em reduction} with topic embeddings.
Our experiments demonstrate that default LLM embeddings  (even after manual fine-tuning) may not consistently improve authorship attribution accuracy. Instead, Bayesian analysis with  topic embeddings   trained on ``function words''  yields superior out-of-sample classification performance. This suggests that traditional (small) statistical language models, with their interpretability and solid theoretical foundation, can offer significant advantages in authorship attribution tasks. The code used in this analysis is available at \href{https://github.com/sowonjeong/slm-2-llm}{github.com/sowonjeong/slm-to-llm}.

\end{abstract}

\noindent%
\vfill

\newpage
\section{Introduction}

The rise of generative AI marked a pivotal shift in how society approaches knowledge acquisition
and task automation. Today, text-generating systems have become one of the most recognizable
representations of AI.
These systems are built on Large Language Models (LLMs) which  are trained on a vast corpus of text data. In order to obtain more tailored answers, LLMs can be fine-tuned using  specialized datasets.
Unlike traditional statistical models defined through a likelihood,  LLMs are 
an example of a black-box simulation-based model  defined implicitly through a mapping (transformer) that transforms prompts into a text output.
While  LLMs are not probabilistic models per-se, these massive architectures do rely on several tools from
probability theory, statistics and machine learning (latent embeddings, next token prediction through conditional distributions, deep learning etc).
The purpose of this paper is to review aspects of language models (both small and large) from the point of view of an applied statistician. 


Several large-scale language models, such as BERT \citep{devlin2019bert, liu2019robertarobustlyoptimizedbert}, GPT \citep{brown2020gpt3} and LlaMa 
\citep{touvron2023llama}, have shown unprecedented humanlike  performance in various natural language processing tasks (read/write skills, sentiment analysis or question answering). However, these models are also prone to hallucinations and can provide contradictory answers to similar queries. 
As the AI technology proliferates, so do concerns about the accuracy of information provided by these tools and how reliable they might be.
Our analysis focuses on  statistical performance of  word/phrase embeddings as features for stylometric analysis and text classification.
We investigate whether general-purpose embeddings generated by these LLMs can be used effectively in authorship attribution  without fine-tuning.
Authorship attribution is closely related stylometry in linguistics, which consists of identifying subtle syntactic or lexical patterns unique to individual authors, and has been tackled with numerous statistical approaches \citep{yule1939sentence,  Williams1975Mendenhall, holmes1995federalist, Seroussi2014Authorship, sari2018topic}.
 Research has demonstrated that unsupervised embeddings from models like BERT can capture deep semantic relationships between words \citep{reimers2019sentencebert}. Our comparison focuses on determining whether embeddings from LLMs can distinguish authors' styles as effectively as (or better than) traditional small   language models.  By a small language model, we understand  a traditional likelihood-based model that aims at dimension reduction (either through latent variables or variable selection).

There are not many  benchmark text datasets that have received more attention than  the Federalist Papers.
With the emergence of LLMs, our paper has aimed at shedding some new light onto this classic dataset, either confirmatory or contradictory. 
We deploy general-purpose (and fine-tuned) embeddings from BERT, GPT and LlaMa, alongside  with traditional latent variable models such as Latent Dirichlet Allocation (LDA) and Principal Component Analysis (PCA), inside statistical classification techniques such as LASSO \citep{tibshirani1996lasso} and Bayesian Additive Regression Trees (BART) \citep{Chipman2010bart}. We also explain the differences between various word/phrase embeddings and discuss how to aggregate them in a document. The results from the Federalist Papers analysis are mixed.
Our findings suggest that  {\em dimension expansion} (with generic embeddings through LLMs) may not always result in better out-of-sample performance for authorship attribution. While LLMs are known to excel at capturing semantic similarity and next token prediction, they may overlook key syntactic or stylistic features that are crucial for authorship detection. These results are consistent with recent research \citep{fabien2020bertaa} that highlights the importance of specialized task-specific fine-tuning for LLMs. On the other hand, {\em dimension reduction} techniques (such as  LDA)  provide tailored low-dimensional embeddings that lead to more robust results and that seem to capture topics and stylistic patterns.
Contrary to the assumption that larger models always outperform traditional approaches, our results suggest that dimension reduction through latent topics remains a valuable strategy for text analysis. The clear winner in our out-of-sample comparisons was a small language model built with  topic embeddings by Bayesian tree classifier. This model points at Madison being the author of the disputed papers, adding to the evidence obtained by \cite{Mosteller1963Inference}. 

Due to its ability to synthesize information from various sources, it is tempting to use text-generating systems for information acquisition. For example, a user might like to seek a diagnosis for a medical condition based on self-reported symptoms \citep{kim2024adaptive}. However, many users may not consider the possibility of asking the same question multiple times and seeing how much the answer varies.  A statistician would typically want to understand underlying variation in the answers to be able to determine how reliable the answer is. We show an anecdotal example in the context the Federalist No. 52, asking ChatGPT4 for its authorship where we see contradictory evidence, human-like confusion and even humility admitting a contradictory answer. See the snapshot of the conversation in Figure~\ref{fig:intro-gpt-uncertain} in the Supplement. 
Depending on the quality of the input query, the output of a text-generating system can be a predictable paraphrase, or a unique creation not replicable after repeated prompting. There is inherent uncertainty inside the generative model extent to which depends heavily on the type of question asked.  

\begin{figure}
    \centering
    \begin{subfigure}{.40\textwidth}
    \centering
    \includegraphics[width=.95\linewidth]{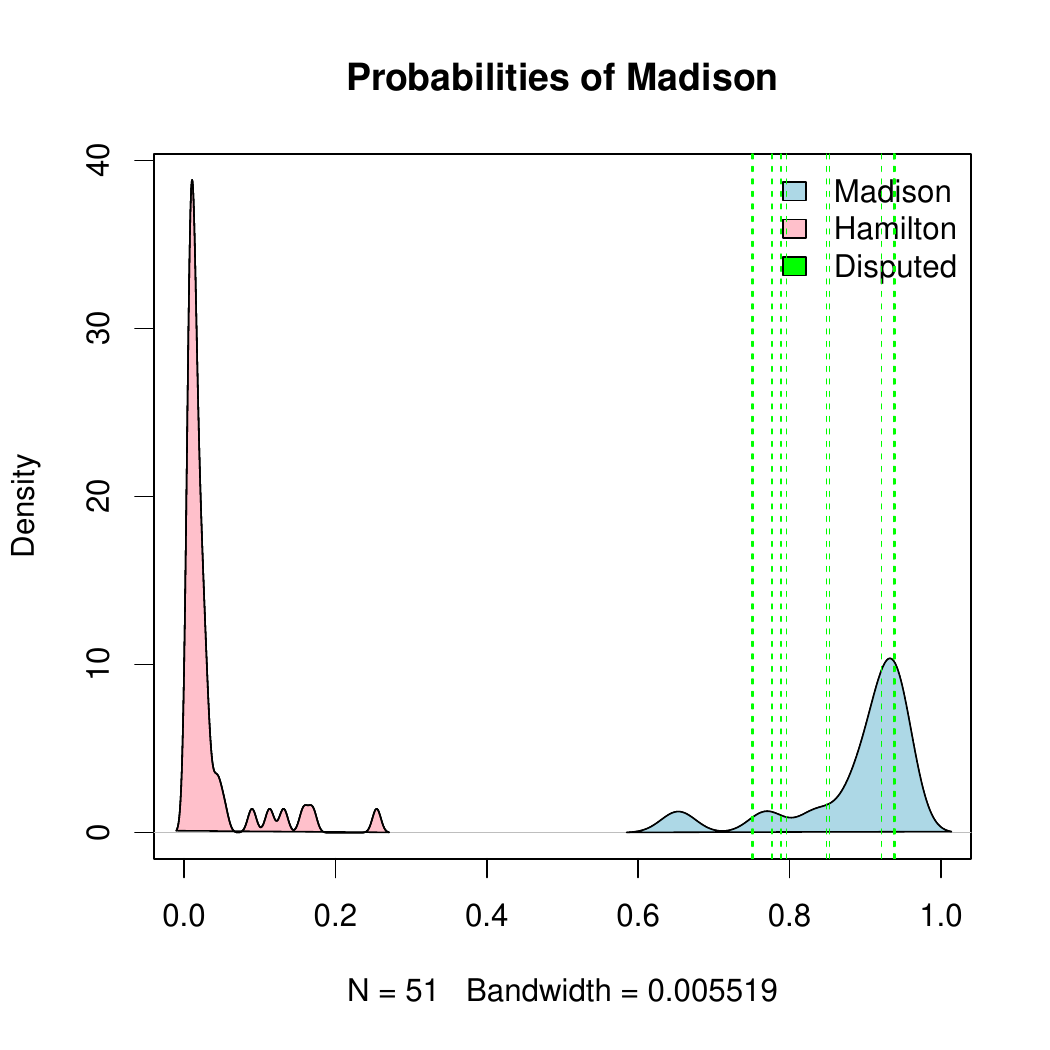}
    \caption{LDA}
    \end{subfigure}
    \begin{subfigure}{.40\textwidth}
    \centering
    \includegraphics[width=.95\linewidth]{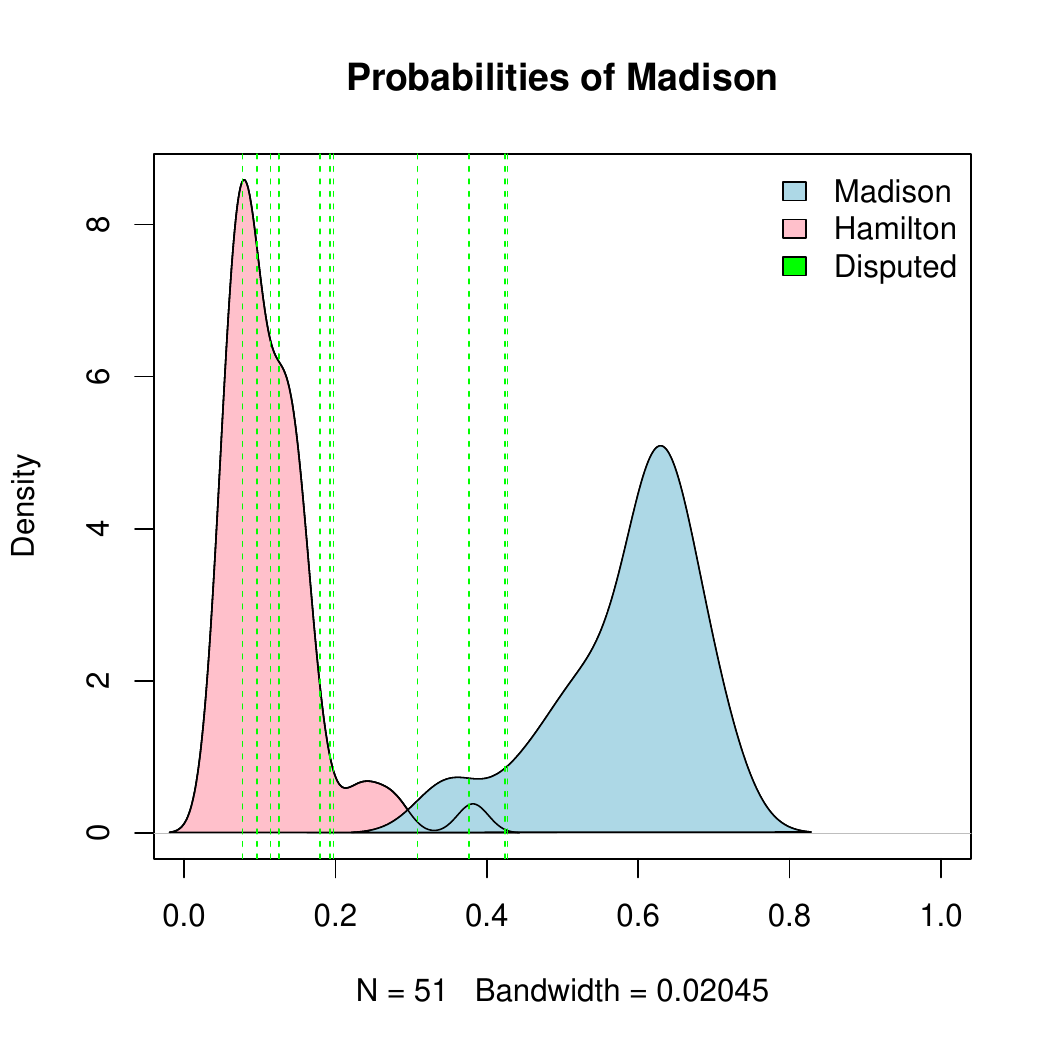}
    \caption{GPT4}
    \end{subfigure}
    \caption{BART classification probability based on document embeddings. The red density is the kernel density estimate of predicted probabilities of BART for papers authored by Hamilton, and the blue density is the kernel density estimate of the ones by Madison. The predicted probabilities of disputed papers are denoted as green vertical lines. For LDA, the results are based on the word counts of ``functions words" as an input. The well-separated densities between Hamilton and Madison indicate less uncertainty for the prediction on the disputed papers.}
    \label{fig:uq-prediction}
\end{figure}

One of the fundamental goals of Statistics is to disentangle and quantify such uncertainty. In the simplified context of authorship attribution, statistical models can provide not only answers but also surrounding uncertainty. Our classification model emits an estimate of the probability that  Madison (as opposed to Hamilton) wrote each disputed paper based on extracted language features. Attribution can then be based on the relative location of this estimate compared to the densities of estimated probabilities of Hamilton versus Madison based on the labeled papers. The gap between the two density estimates signifies the discriminative ability of the features. As shown in Figure~\ref{fig:uq-prediction}, the probability distributions obtained from training data reveal differences across embeddings. Specifically, the densities generated by the LDA embeddings for Madison and Hamilton are entirely separated, whereas the GPT embeddings display substantial overlap. This suggests that with LDA, we are able to better separate the labeled papers and attribute all disputed papers to Madison without too much hesitation. In contrast, although GPT embeddings also predict Madison for Nos. 49, 51, 63, as the estimated probabilities fall within the overlapping regions, the uncertainty for the prediction is much higher.
We assess the quality of  various classifiers using various features (embeddings) with leave-one-out classification.  

We hope that our paper will educate practitioners curious about deploying generic embeddings from LLMs in their everyday statistical text analyses.

The paper is structured as follows. Section~\ref{sec:original-answer} introduces the disputes on authorship of the Federalist Papers and classical statistical approaches to addressing the problem. In Section~\ref{sec:review}, we present a comprehensive overview of traditional statistical language models and modern Large Language Models (LLMs), drawing connections between them. Section~\ref{sec:proposed-approach} introduces our approach, which integrates statistical models with LLMs for addressing the Federalist authorship problem. Section~\ref{sec-res} analyzes the application of various language models to authorship prediction, word selection, and interpretability. Based on our case study, Section~\ref{sec:advice} offers practical recommendations for using vanilla LLMs and improve their usage by statistical models. Finally, Section~\ref{sec:conclusion} summarizes our findings and key insights from the case study.

\section{Statisticians Tackling Disputed Federalist Papers}\label{sec:original-answer}
The authorship distribution of the Federalist Papers, a collection of 85 essays advocating for the ratification of the United States Constitution, has been a subject of historical inquiry and debate. Published between 1787 and 1788 under the pseudonym ``Publius", the papers were instrumental in shaping public opinion and garnering support for the proposed Constitution. While the authorship of the Federalist Papers has traditionally been attributed to Alexander Hamilton, James Madison, and John Jay, the exact division of labor among these three founding fathers remains a topic of scholarly discussion \citep{authorship1944adair}.

According to Douglass Adair, Alexander Hamilton wanted to keep the authorship of ``The Federalist Papers" secret due to the politically sensitive nature of the essays and the potential repercussions for openly claiming authorship. The day before his deadly duel with Aaron Burr in 1804, Hamilton handed the list of authorship to his lawyer, Egbert Benson. This list attributed 63 out of the 85 papers to Hamilton himself. In 1818, James Madison contradicted Hamilton's claim of authorship through Jacob Gideon's edition, asserting that he had written 29 essays instead of the 14 essays listed in Benson's account (See Table~\ref{tab:summary-authorship} for detailed division of labor). Other lists, such as those by Kent and Washington, also exist. This discrepancy has fueled scholarly debate and analysis.

Historians generally agree on the primary authors: Hamilton wrote the majority of the papers, Madison contributed significantly, and Jay authored a few. Despite these general attributions, determining the precise authorship of each paper has proven challenging due to the conflicted evidence and the secretive nature of the project. 
For example, simple statistics like average sentence length cannot distinguish the style of Hamilton from that of Madison. This necessitated more sophisticated approach to resolve the authorship question.

%
\begin{table}
\centering
\resizebox{0.75\textwidth}{!}{%
\begin{tabular}{@{}c c c c l c@{}}
\toprule
\textbf{Number} & \textbf{Benson}     & \textbf{Gideon}    & \textbf{Mosteller \& Wallace}       \\ \midrule
1     & Hamilton      & Hamilton  & Hamilton \\
2–5   & Jay           & Jay       & Jay      \\
6–9   & Hamilton      & Hamilton  & Hamilton \\
10    & Madison       & Madison  & Madison  \\
11–13 & Hamilton      & Hamilton  & Hamilton \\
14    & Madison       & Madison    & Madison  \\
15–17 & Hamilton      & Hamilton  & Hamilton \\
18–20 & Madison \& Hamilton         & Madison    & Madison \& Hamilton \\
21–36 & Hamilton      & Hamilton  & Hamilton \\
37–48 & Madison       & Madison   & Madison  \\
49–53 & Hamilton      & Madison   & Madison  \\
54    & Jay & Madison  & Madison  \\
55–58 & Hamilton      & Madison  & Madison  \\
59–61 & Hamilton      & Hamilton  & Hamilton \\
62–63 & Hamilton      & Madison & Madison  \\
64    & Hamilton & Jay     & Jay      \\
65–85 & Hamilton      & Hamilton & Hamilton \\ \bottomrule
\end{tabular}%
}
\caption{Summary of authorship for 85 Federalist papers \citep{authorship1944adair}. Hamilton left a note to Egbert Benson before his death about the authorship of the papers in 1804. Madison claimed the authorship of each paper through Jacob Geideon's edition, published in 1818.}
\label{tab:summary-authorship}
\end{table}

The first rigorous statistical attempt to address the authorship of the disputed Federalist Papers was undertaken by \cite{Mosteller1963Inference}. Their approach applied Bayesian inference to the frequency of function words -- articles, prepositions, and conjunctions -- on the grounds that these words are stylistically neutral and less topic-dependent. The function words, which act as linguistic markers, were used to differentiate the authors based on patterns that remain stable across different contexts. Their analysis concluded that Madison authored all 12 disputed papers. Although the log odds strongly supported Madison's authorship for most of the papers, Nos. 55 and 56 presented somewhat weaker evidence due to the limited presence of marker words in these essays. The jointly authored papers -- Nos. 18, 19, and 20 -- also showed strong support for Madison's primary authorship, although No. 20 provided less definitive evidence, likely due to the shorter length of the text (see Appendix~\ref{sec:detailed-MW} for a detailed review).

Building on the work of \cite{Mosteller1963Inference}, subsequent researchers have employed different statistical techniques to address the authorship problem. \cite{holmes1995federalist} used vocabulary richness measures and a genetic rule-finder algorithm to analyze high-frequency words. \cite{Tweedie1996Neural} applied a two-layer neural network trained on the rate of occurrence of 11 key words, a subset derived from the function words used by \cite{Mosteller1963Inference}. Other approaches include \cite{Robert1998Seperating}, who used a separating hyperplane based on linear classifiers to distinguish between authors. This method relied on the 70 preselected function words identified in the original Bayesian analysis, creating a decision boundary to separate texts by authorship. \cite{Diederich2003SVM} introduced Support Vector Machines (SVM) for authorship attribution, while \cite{popescu2007kernel}) explored kernel-based methods for text classification. \cite{Collins2004Detecting} expanded the scope of analysis by examining collaborative patterns in rhetorical style and syntactic structure, aiming to capture unique stylistic features of each author. This research emphasized the potential for more nuanced modeling of joint authorship beyond simple binary classification. For a detailed discussion on joint authorship and the selection of significant words, see Section~\ref{sec:discussion}.

A more recent contribution by \cite{Kipnis2022Higher} approached the problem from a multiple hypothesis testing framework. Their method identifies sparse signals in large frequency tables, constructed from word counts for each author. This approach automatically selects significant discriminators through a Higher Criticism (HC) threshold and successfully attributes all 12 disputed papers to Madison. Additionally, their joint authorship analysis aligns with the findings of \cite{Mosteller1963Inference}, indicating that Madison was the primary author of Nos. 18 and 19. However, the contribution levels for No. 20 remain ambiguous due to the paper's brevity. A limitation of the HC measure is that it lacks a formal mechanism for quantifying the uncertainty associated with the decision rule (see Appendix~\ref{sec:hc-authorship} for further discussion).


\section{Small and Large Language Models}\label{sec:review}
This section  provides a roadmap for  language models  ranging from small (such as Latent Dirichlet Allocation (LDA)) to large  (such as GPT or LlaMa). Our distinction between  small versus Large Language Models is based on underlying statistical features.
We regard a language model small when it is trained on smaller available datasets without relying on pre-trained features from data that is not available to the user.
Small language models aim at dimension reduction (as opposed to dimension expansion as we will explain later in Section~\ref{sec:llm}) and focus on simplification and insight through probabilistic modeling. 

In contrast, Large Language Models are trained on extensive, non-task-specific datasets, often with high embedding dimensions (termed ``dimension expansion") --for example, GPT 3 has a maximum embedding dimension of 2048, while LlaMa reaches 4096.  Many widely recognized Large Language Models (LLMs) rely on the transformer architecture \citep{vaswani2017attention}, although this architecture does not define their core purpose of modeling the joint probability of word sequences. The definition of LLMs remains a subject of debate, which we provide further literature review in Appendix~\ref{sec:llm-def}.

We regard text as unstructured data $\mathbf{W} = (W_1, W_2, \ldots, W_T)'$ with layered information (e.g. semantic and syntactic) made up of tokens $w_j$ chosen from a vocabulary set $\mathcal{W}$ of size $N=|\mathcal W|$, where $j$ denotes the position  in the sequence of length $T$. Token refers to a single unit of text, which could either be a word, or a word stem. 

\subsection{Probabilistic Language Models}\label{plm}

The fundamental objective of language models is characterizing the probabilistic mechanism for $\mathbf{W}$ that captures how often tokens (in a particular order) occur in the language. This can be captured with a likelihood function

\begin{equation}
P(\bm W=\bm w| \btheta)=\pi_{\btheta}(\bm w)\quad \text{for}\quad \bm w\in \mathcal W^T,\label{eq:model}
\end{equation}
which explains how likely $\bm w$ is under different values of the   model parameters $\btheta$. In LLMs, $\btheta$ is entirely uninterpretable and has massive dimensionality,  parametrizing layers and layers of encoders and decoders. In small language models,   $\theta$ has moderate dimensionality and serves to provide insight into the language structure.
Training a language model on real data $\mathcal D$ involves estimating a parameter $\hat\theta$ for which the likelihood of observing aspects (sequences) of $\mathcal D$ is the largest.
 For LLMs, the training data $\mathcal D$ is a vast corpus of text (inaccessible to the user) while for small language models, $\mathcal D$ is a collection of observed text documents (such as our Federalist Papers). Training the language model \eqref{eq:model} can be facilitated by making some  simplifying structural assumptions about $\pi_{\btheta}(\bm w)$. We will  roughly group language models according to the assumption made.

{\em Bag-of-words Models} utilize the simplifying assumption that the order of the words in the sequence $\bm W$ does not matter. In statistics, this corresponds to the notion of exchangeability where $\pi_{\btheta}(\bm w)=\pi_{\btheta}(\bm w_{\sigma(1,\dots,T)})$ for any permutation $\sigma(1,\dots,T)$. 
Such language models very often operate on a specific summary statistic (word counts) as opposed to raw data (word sequences). For example, our analysis of the Federalist Papers involves training data documents $\mathcal D=\{D_{i}\}_{i=1}^n$ where $D_i=\{w_{ij}\}_{j=1}^{T_i}$. The word count matrix  $X=\{x_{ij}\} \in \mathbb{R}^{n \times N}$ can be then constructed by counting the occurrence of each word in each document  $x_{ij}=\sum_{i=1}^n\sum_{j=1}^N\mathbb I\{w_{j}\in D_i\}$, where $N = |\mathcal{W}|$ denotes the size of the vocabulary.  
We talk about bag-of-words models later in Section~\ref{sec:statistical-lm} in the context of Latent Dirichlet Allocation and Latent Semantic Analysis.


{\em Autoregressive models}  characterize the joint distribution using the chain rule of probability as follows:
\begin{equation}\label{eq-ar}
P(\bm W=\bm w| \btheta) =P(W_1=w_1|\btheta) \prod_{j=2}^T  P(W_j=w_j|\mathbf{W}_{[1:j-1]}=\mathbf{w}_{[1:j-1]},\btheta) 
\end{equation}
where $W_{[1:j]}=(W_1,\dots, W_j)'$.
This factorization  leverages an assumption that each word is generated sequentially, depending only on the words that came before.    One prominent example of an autoregressive statistical language model is the $n$-gram model, which is based on the Markov assumption that the occurrence of a word depends only on $k$ preceding words through a transition kernel  $P(W_j=w_j|\mathbf{W}_{[1:j-1]}=\mathbf{w}_{[1:j-1]},\btheta)=M_\btheta(W_{j-1} = w_{j-1},\dots, W_{j-k}=w_{j-k})$. Having learned this kernel, one immediately obtains a naive text generator by sliding this kernel left to right and generating the next word from this conditional distribution. As an example, the maximum entropy language model assumes
$ P(W_j=w_j|\mathbf{W}_{[1:j-1]}=\mathbf{w}_{[1:j-1]},\btheta)\propto \exp\left\{\btheta'f(\mathbf{w}_{[1:j]})\right\}$, where $f(\mathbf{w}_{[1:j]})$ is a ``feature vector", a.k.a embedding. 
Hidden Markov Models (HMMs) \citep{rabiner1989tutorial} in another extension of this framework by introducing latent variables that represent underlying structures or states in the data. In an HMM, the sequence of latent states $(\bm{S} = (S_1, \dots, S_T)$ evolves according to a first-order Markov process, and each word is generated based on the current latent state, $P(\bm{W} = \bm{w}, \bm{S}=\bm{s}) = P(S_1 = s_1)\prod_{t=2}^TP(S_t = s_t|S_{t-1}=s_{t-1})\prod_{t=1}^T P(W_t = w_t|S_t = s_t).$
HMMs allow for richer modeling of sequence dependencies compared to simple $n$-gram models by incorporating latent structures, but they still rely on simplifying independence assumptions and struggle with long-range dependencies.

The autoregressive structure prevalent in the decoder-only transformer models (discussed later in Section~\ref{sec:llm}) consist of a stack of transformer \textbf{decoder} layers that predict the next word in a sequence based on the previous context. These models are designed primarily for tasks that involve generating text. The GPT \citep{radford2018improvingLU, radford2019gpt2, brown2020gpt3} and LlaMA \citep{touvron2023llama} series are both based on autoregressive architectures, but they take distinct approaches: GPT models scale all dimensions simultaneously, while LLaMA optimizes architectural efficiency. Both seek to better approximate the true probability distribution of language sequence, but through different statistical trade-offs.

{\em Encoder-based models} like BERT \citep{devlin2019bert} employ a different strategy. These models do not assume a strict left-to-right (or right-to-left) dependence but instead use a bidirectional mechanism. In this approach, each word $w_i$ is conditioned on both past and future words, allowing the model to capture context from the entire sequence. Formally, this can be viewed as estimating conditional probabilities
\begin{equation}\label{eq-ae}
P(W_j=w_j| \btheta, W_i=w_i\,\,i\neq j )\quad\text{for}\quad 1\leq j\leq T.
\end{equation}
Unlike in the autoregressive case, where the conditionals uniquely define the joint distribution, 
it is possible that no joint
probability  function $P(\bm W=\bm w | \btheta)$ exists that is compatible with all the given conditionals  \eqref{eq-ae} \citep{besag1974spatial, arnold1989compatible}. 
Language models that  focusing purely on modeling conditionals \eqref{eq-ae} thus deviate from  purely statistical approaches that would start with $P(\bm W=\bm w | \btheta)$ and derive the conditionals from it.

\begin{figure}
   \centering
\includegraphics[width=0.8\linewidth]{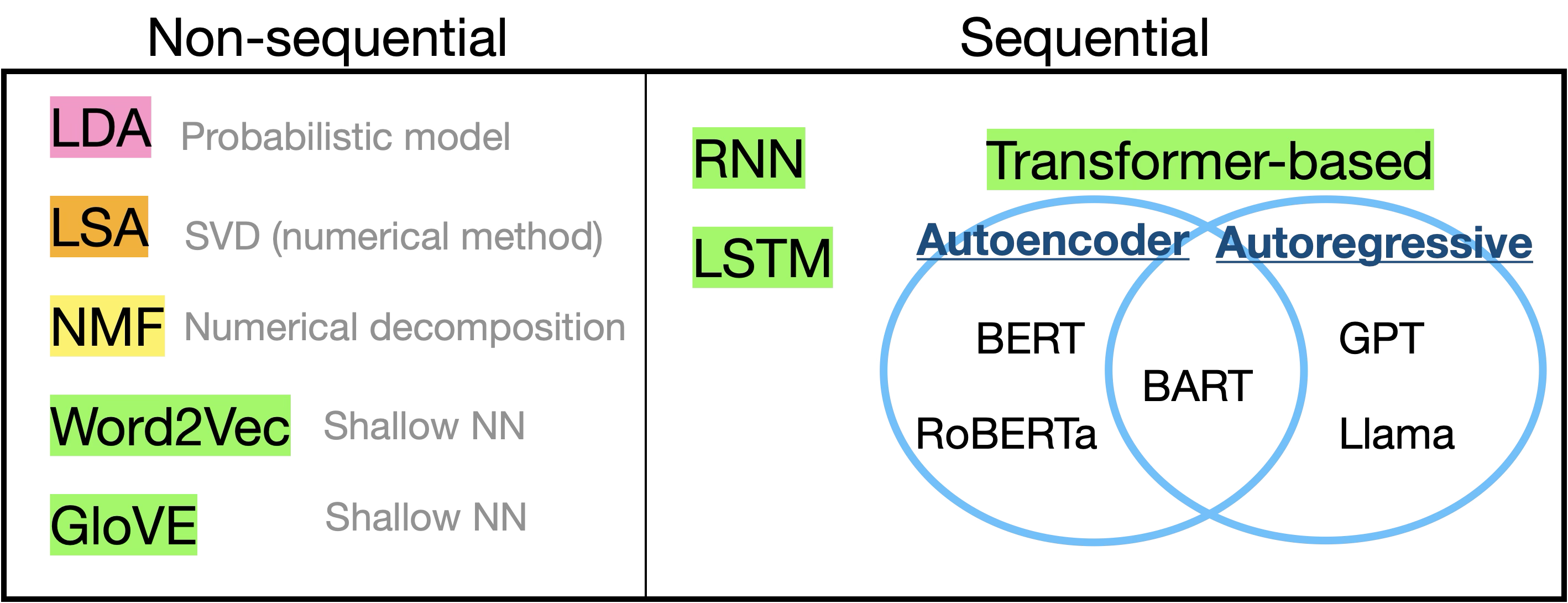}
    \caption{Stylized representation of the relationships among language models. Each model learns a function $f: \mathbf{W}_{[1:T]} \to \mathbb{R}^p$, mapping a word sequence to a latent representation. LDA (red) is a probabilistic model, while LSA and NMF (orange, yellow) use numerical methods like matrix factorization. Word2Vec and GloVe (green) introduce shallow neural networks, followed by recurrent models (RNN, LSTM) and transformer-based models (green), which capture long-range dependencies. Modern deep learning methods (blue) rely on large neural networks with autoencoding and autoregressive objectives. This spectrum illustrates the shift from probabilistic and numerical methods to neural architectures.}
    \label{fig:lm-cartoon}
\end{figure}
\subsection{Small Language Models: From Counting to Learning}\label{sec:small-lm}

We provide an historical overview of small language models, distinguishing between statistical and neural approaches. Statistical language models are small models rooted directly in probabilistic assumptions for both model structure and parameter estimation. Neural language models, though still grounded in probabilistic principles, differ by estimating parameters $\bm{\theta}$ through training neural networks. 
Neural language models are a precursor to the Large Language Models departing from purely statistical model. 

\subsubsection{Statistical Language Models}\label{sec:statistical-lm}

Latent Semantic Analysis (LSA) \citep{deerwester1990indexing} and Probabilistic Latent Semantic Analysis (pLSA) \citep{hofmann1999learning} aim to uncover semantic relationships across terms and documents. LSA leverages Singular Value Decomposition (SVD) on the term-document matrix $X \in \mathbb{R}^{n \times N}$, where $n$ is the number of documents and $N$ is the vocabulary size, to reduce dimensionality and reveal underlying semantic structure. This reduced representation yields a compact, $p$-dimensional form with $p \leq n \wedge N$, either on term space or document space. This method, however, is not explicitly probabilistic, which limits its capacity to incorporate uncertainty.

In response to the limitations of LSA, Probabilistic Latent Semantic Analysis (pLSA) \citep{hofmann1999learning} provides a probabilistic interpretation of LSA by modeling documents as mixtures of latent variables (e.g topics). For each document $D$, we have a distribution over topics, denoted $P(Z=z|D)$, where $z \in \{1,...,K\}$ (with $K$ being the number of topics). For each word $w_{ij}$ in a document $D_i$, pLSA models the probability of a word sequence in a document as
\begin{equation*}\label{eq:plsa}
    P(\mathbf{W}_{i\cdot} = \mathbf{w}_{i\cdot}) = \prod_{j=1}^N \sum_{z=1}^K P(W_{ij} = w_{ij}|Z = z) p(Z = z|D_i). 
\end{equation*} The objective of pLSA was shown to be compatible \citep{ding2006nonnegative} with the one of Nonnegative Matrix Factorization (NMF) \citep{lee1999learning}. However, pLSA lacks a full generative process for the corpus, meaning it cannot provide a unified probabilistic model across all documents or explicitly capture uncertainty at the corpus level. This limitation spurred the development of Latent Dirichlet Allocation (LDA) \citep{blei2003latent}, which employs Dirichlet priors to define document-topic and topic-word distributions. 

LDA, a fully probabilistic model, assumes that documents are mixtures of latent topics and generates words by sampling from these topics. Each document $D_i$ consists of up to $K$ topics and the proportion of topics covered can be represented by a vector  $\boldsymbol{\varphi}_i\in [0,1]^K$  where $\sum_{k=1}^K\varphi_{ik}=1$ which is assumed to be drawn from a Dirichlet distribution, $Dir(\alpha)$. Within each topic $k$, a word $w_{\cdot j}$ is sampled with probability $\eta_{kj}$, where $\sum_{j=1}^N\eta_{kj}=1$ and $\eta_k$ are drawn from another Dirichlet distribution $Dir(\beta)$. Then, for each document $D_i$, a topic $z_{ij}$ is drawn from $\boldsymbol{\varphi}_i$ and subsequently the word $w_{ij}$ is generated from the corresponding word-topic distribution parametrized by ${\eta}_{z_{ij}}$. The likelihood is given as 
\begin{equation*}
P({\boldsymbol {W}} = \mathbf{w},{\boldsymbol {Z}} = \mathbf{z},{\boldsymbol {\varphi }} ,{\boldsymbol {\eta }}|\alpha ,\beta )=
P(\boldsymbol{\eta}|\beta )P(\boldsymbol{\varphi}|\alpha )\prod_{i=1}^{n}\prod _{j=1}^{N}P(Z_{ij}\mid \varphi_{i})P(w_{ij}|\eta _{Z_{ij}})
\end{equation*}
where $z_{ij}$ is the latent topic assignment for the word $j$ in the document $i$, and $w_{ij}$ is the corresponding observed word count. 

The development of statistical language models began with a focus on language's sequential nature, seen in $n$-grams and HMMs, and then shifted toward modeling semantic relationships and thematic clustering, as with pLSA and LDA. However, the latter models rely heavily on summary statistics, such as word counts, rather than fully leveraging the sequential and contextual richness of language. This evolution highlights a trade-off: models tend to emphasize either sequence or semantics, but not both, leaving limitations in generating nuanced representations of language. This gap laid the groundwork for models that integrate both sequential and semantic aspects, leading to the development of more sophisticated language embeddings, discussed in Section~\ref{sec:llm}.

\subsubsection{Neural Language Model}

The transition from small language models to LLMs was significantly influenced by intermediary models such as Word2Vec \citep{mikolov2013efficient,mikolov2013distributed} and GloVe \citep{pennington2014glove}, which introduced more advanced methods for creating word embeddings. These models marked a departure from the traditional statistical approach, like n-grams, by representing words as dense vectors in a continuous space introduced by \cite{bengio2003neural}. The model still retains an explicit likelihood but uses a neural parameterization:
\begin{equation*}
P(\mathbf{W}=\mathbf{w}|\theta) = P(W_1=w_1|\theta) \prod_{j=2}^T \text{softmax}(f_\theta(w_{[1:j-1]}))
\end{equation*}
where the autoregressive nature remains as in (\ref{eq-ar}), $f_\theta$ is a neural network to model $P(W_j=w_j|W_{[1:j-1]} = w_{[1:j-2]})$ and, for a vector $ \mathbf{x} = [x_1, x_2, \ldots, x_n] $, the softmax function is defined as 
\begin{equation}\label{softmax}
\text{softmax}(\mathbf{x})_i = {e^{x_i}}/\Big( {\sum_{j=1}^n e^{x_j}} \Big),\text{ for } i = 1, 2, \ldots, n. 
\end{equation}
The use of neural networks in this setting is motivated by the universal approximation theorem \citep{cybenko1989approximation}, which implies that neural networks can model an extensive range of functional forms, including complex relationships potentially embedded in language data. This flexibility allows neural networks to approximate nuanced patterns in word co-occurrences and context, expanding the family of structures that can be effectively captured.

Word2Vec \citep{mikolov2013efficient, mikolov2013distributed} exemplifies the early neural approach. It learns word embeddings based on the distributional hypothesis, where words that appear in similar contexts share semantic meaning.  Mathematically, Word2Vec leverages assumption that a word ($w_i$) only depends on the window of size $t$ around it, simplifying (\ref{eq-ae}). The algorithm typically involves training a shallow neural network model by maximizing the conditional probabilities, either using the Continuous Bag of Words (CBOW) 
\begin{equation*}\label{eq-cbow}
    P(W_i = w_i|W_{[i-t+1:i+t-1]\setminus\{i\}} = w_{[i-t+1:i+t-1]\setminus\{i\}})
\end{equation*} or Skip-gram architecture, 
\begin{equation*}\label{eq-skip-grapm}
    P(W_{[i-t+1:i+t-1]\setminus\{i\}} = w_{[i-t+1:i+t-1]\setminus\{i\}}|W_i = w_i)
\end{equation*} to predict the surrounding words given a target word or vice versa.

To train Word2Vec efficiently, noise-contrastive estimation (NCE) \citep{gutmann2012noise, mnih2012fast} is employed to approximate the full softmax (\ref{softmax}) objective by reducing it to a binary classification task. Instead of summing over the entire vocabulary, NCE introduces a set of negative (noise) samples $ w_{\text{noise}} $ drawn from some noise distribution $Q(w)$, and aims to optimize the following objective (Skip-gram case):
\begin{equation*}
\log \sigma(f_\theta(w_i, \mathbf{w}_{[i-t+1:i+t-1] \setminus \{i\}})) + \sum_{j=1}^{k} \mathbb{E}_{w_j \sim Q(w)} \left[ \log \sigma(-f_\theta(w_j, \mathbf{w}_{[i-t+1:i+t-1] \setminus \{i\}})) \right],
\end{equation*}
where $ k $ is the number of negative samples, and $\sigma(x) = 1 / (1 + e^{-x})$. Here, $ f_\theta $ assigns a high score to true word-context pairs and a low score to noise-context pairs, teaching the model to differentiate between real and random associations. This noise-contrastive setup can be viewed as a form of statistical regularization as well as computational practicality. By introducing noise samples, the model learns to distinguish meaningful word-context relationships from random noise, smoothing the estimated distribution and avoiding overfitting to specific word pairs. 

Both Noise-Contrastive Estimation (NCE) in Word2Vec and Generative Adversarial Networks (GANs) \citep{goodfellow2014gan} share a fundamental principle of contrastive learning, where the model learns by distinguishing between real and fake data. In Word2Vec, NCE simplifies the task of predicting the entire vocabulary distribution by reformulating it as a binary classification problem --- determining whether a word-context pair is real or sampled from a noise distribution $ Q(w) $, typically derived from unigram or smoothed unigram probabilities. In GANs, a generator $G(z)$ is trained to produce fake data $\tilde{x} = G(z)$ that fools a discriminator $D(x)$, which aims to distinguish real $x \sim P_\text{data}(x)$ from generated $\tilde{x} \sim P_G(z)$. The GAN objective is:
$$
\min_G \max_D \mathbb{E}_{x \sim P_\text{data}}[\log D(x)] + \mathbb{E}_{z \sim P_z}[\log (1 - D(G(z)))].$$
Unlike Word2Vec, where $Q(w)$ remains static, GANs iteratively train $G$ to improve synthetic data quality, reflecting its dynamic contrastive learning framework.

Since the success of Word2Vec \citep{mikolov2013distributed}, the language model gears toward prediction rather than co-occurence statistics. GloVe \citep{pennington2014glove} bridges classical and neural network methods by directly connecting word co-occurrence statistics with neural embeddings by weighted regression problem.
Unlike usual Bag-of-Words approach, they consider the word co-occurence matrix $\{C_{ij}\}_{i,j = 1}^N$, where $C_{ij}$ represents the number of occurrence of the word $w_i$ in the context of $w_j$ of window size $t$. GloVe approaches the language model as a prediction problem for $C_{ij}$.
The model minimizes a weighted least-squares objective function:
\begin{equation*}\label{eq:glove}
\mathcal{L}(\theta) = \sum_{i=1}^N \sum_{j=1}^N \alpha(C_{ij}) \big(\log C_{ij} - (\mathbf{z}_i^\top \tilde{\mathbf{z}}_j + b_i +\tilde{b}_j)\big )^2,
\end{equation*}
where $\alpha(C_{ij})$ is a pre-defined weighting function, $z_i = f_\theta(w_i) \in \mathbb{R}^p$ is a $p$-dimensional word embedding, $f:\mathcal{W} \to \mathbb{R}^p$ is an embedding function and $b_i$ is the bias term where both $z_i$ and $b_i$ terms are learned by neural networks.

\subsection{Large Language Model (LLMs)}\label{sec:llm}
Recurrent Neural Networks (RNNs) \citep{rumelhart1986learning} and Long Short-Term Memory networks (LSTMs) \citep{hochreiter1997long} were pioneering models in capturing sequential and semantic relationships in data. However, their sequential processing nature posed significant computational bottlenecks, particularly when training on large datasets. The introduction of transformers \citep{vaswani2017attention} marked a paradigm shift in sequence modeling. By leveraging attention mechanisms, transformers replaced the need for sequential processing with a parallelized framework, enabling them to model relationships among all elements in a sequence simultaneously. This innovation, coupled with scaling to larger models and massive datasets, significantly advanced the ability to model both semantic and sequential properties of data efficiently.

In the following sections, we will explore the main building blocks of modern large models: (1) transformer architecture with an emphasis on attention mechanism (2) pre-training scheme and (3) large model as well as massive training dataset. 

\subsubsection{The Transformer}\label{review:transformer}

Unlike previous sequence models \citep{rumelhart1986learning, hochreiter1997long}, which processed input data sequentially, transformers introduced by \cite{vaswani2017attention} solely rely on attention mechanisms allowing to consider relationships among all words in a sequence simultaneously, regardless of their position.


\paragraph{Attention Mechanism}
The attention mechanism in the transformers can be understood as a process of computing a \textbf{weighted average} of word representations in a sequence, where the weights reflect the relevance of each word to the word being processed. For each word in the sequence, the model computes a query vector $Q$, a key vector $K$, and a value vector $V$, all of which are linear transformations of the word's embedding. The attention score between two words is given by the dot product of the query from one word and the key from the other. Specifically, the attention scores for word $w_i$ with respect to any other words $w_j$ in the sequence are computed as:
$$\text{score}(w_i,w_j) = \frac{Q_i K_j^\top}{\sqrt{d_k}}$$
where $d_k$ is the dimension of the key vectors, used to scale the scores and prevent the values from growing too large. This score reflects the similarity or relevance between the two words based on their query and key vectors.

The attention scores are then normalized using the softmax function (\ref{softmax}) to produce a set of weights that sum to 1:

$$\alpha_{ij} = \text{softmax}(\boldsymbol{\alpha}_i)_j = \frac{\exp\left(\text{score}(w_i,w_j)\right)}{\sum_{k=1}^{T} \exp\left(\text{score}(w_i,w_k)\right)}$$
Here, $\alpha_{ij}$ represents the weight or attention that word $w_i$ assigns to word $w_j$. These weights are then used to compute a weighted average of the value vectors $V$ from all the words in the sequence, producing the final output for word $w_i$:
$$\text{Attention}(Q,K,V)_i = z_i = \sum_{j=1}^{T} \alpha_{ij} V_j.$$

This formulation enables transformers to capture sequential dependencies through learned mixing weights, where the weights $\alpha_{ij}$ determine how much each word contributes to the representation of the current word $w_i$. It allows the model to focus on the most relevant words in the sequence when computing a new representation for each word.

From a statistical perspective, the attention score computation, $\exp(\text{score}(w_i,w_j))$, functions as an adaptive kernel method, aligning closely with kernel regression \citep{nadaraya1964estimating, watson1964smooth}. In transformers, attention can be viewed as an adaptive kernel smoother, where the kernel weights are dynamically determined by the trainable query ($Q$), and key ($K$) mechanism.

\paragraph{Multihead Attention} The transformer model achieves superior performance by leveraging multi-head attention, a mechanism that enables parallel computation across different ``heads" \citep{vaswani2017attention}.
\begin{equation*} \text{MultiHead}(Q, K, V) = \text{Concat}(\text{head}_1, \dots, \text{head}_h)W^O, \end{equation*} where $\text{head}_i = \text{Attention}(QW_i^Q, KW_i^K, VW_i^V)$, the projection matrices $W_i^Q \in \mathbb{R}^{d_\text{model} \times d_k}$, $W_i^K \in \mathbb{R}^{d_\text{model} \times d_k}$, $W_i^V \in \mathbb{R}^{d_\text{model} \times d_v}$ and $W^O \in \mathbb{R}^{h d_v \times d_\text{model}}$ matches up back to the model dimension. For example, in GPT-3, total 96 heads are used and each of the head has dimension 128, and the model dimension ($d_\text{model}$) is 12,288. 
It allows the model to focus on different parts of the sequence and the design parallels ensemble methods \citep{breiman1996bagging}, where combining predictions from multiple models (or here, multiple attention heads) trained on diverse views improves robustness and overall performance. Furthermore, the benefits of multi-head attention stem from the diversity of each attention head, much like the role of boosting in ensemble approaches \citep{freund1997decision}, where multiple weak predictors are combined to enhance accuracy.

\paragraph{Encoder Architecture} In the \textbf{encoder}, input tokens are first converted into dense vectors using an embedding layer (\textit{input embeddings}). Since transformers do not inherently capture the order of tokens, \textit{positional encodings}
are added to the embeddings to incorporate information about the token positions within the sequence. These positional encodings are based on sinusoidal functions, ensuring the model can differentiate between the positions of tokens. 

At the heart of the encoder is the \textit{self-attention mechanism}, which allows the model to focus on different parts of the input sequence simultaneously. The model looks at relationships within a single sequence, allowing each word to attend to \textit{all} the other words in the same sequence. This multi-head self-attention allows the model to capture relationships between all tokens in the input sequence, improving its ability to understand context.

Each encoder layer also contains a \textit{feed-forward neural network (FFN)}, which is applied to each token's representation independently. This network consists of two linear transformations with a ReLU activation function in between, expressed as 
$\text{FFN}(\mathbf{x}) = \text{ReLU}(\mathbf{x} \mathbf{W}_1 + \mathbf{b}_1) \mathbf{W}_2 + \mathbf{b}_2$, where $\mathbf{W}_1, \mathbf{W}_2$ are weight matrices, and $\mathbf{b}_1, \mathbf{b}_2$ are bias vectors. All parameters are learnable. The feed-forward layers allow for non-linear transformation of the input, further refining the token representations.

Finally, the encoder outputs a sequence of dense representations of the same length as the input (``embeddings"), capturing rich contextual information for each token. The original transformer model consists of 6 encoder layers, with each layer having a model dimension ($d_\text{model}$) of 512, divided across 8 attention heads.

\paragraph{Decoder Architecture} The \textbf{decoder} in the transformer architecture follows a similar structure to the encoders. The input to the decoder is the target sequence (shifted by one position) combined with positional encodings. To facilitate next word generation, the decoder incorporates two additional mechanisms. First, the \textit{masked self-attention mechanism} prevents the model from attending to future tokens during training, ensuring that each prediction is made based solely on previously generated tokens and the input sequence. This is crucial for autoregressive tasks such as text generation, where the model must generate tokens sequentially. Second, the \textit{encoder-decoder attention mechanism} enables the decoder to focus on relevant parts of the input sequence. In this mechanism, the queries come from the decoder's previous layer, while the keys and values are derived from the encoder's output, allowing the decoder to use contextual information from the input sequence during generation.

After processing through these layers, the decoder outputs a sequence of logits, one for each token position in the target sequence. These logits are then passed through a softmax (\ref{softmax}) layer to generate the  probability distribution over the target vocabulary, enabling text prediction. Like the encoder, the original transformer uses 6 decoder layers, with the same model dimension ($d_\text{model} = 512$) and 8 attention heads.

\subsubsection{Pretraining and Fine-Tuning}
Modern language models typically follow a pretraining and fine-tuning routines, a framework that
aligns well with Bayesian statistical methodologies. This approach was popularized by models
such as the Generative Pretrained Transformer (GPT) \citep{radford2018improvingLU} and has significantly advanced the
performance of language models. In this paradigm, the pre-trained model parameters, denoted as $\theta$, can be viewed as initialization learned from large-scale text corpora. This initial
configuration can be refined by adapting the architecture to specialized datasets.

\subsubsection{The Largeness}

A large number of parameters as well as the massive training datasets are a defining feature of modern ``large'' language model.
For example, GPT-3 has 175 billion parameters and is trained on approximately 300 billion tokens (570 GB of text), while LlaMA 3's largest model has 70 billion parameters and is trained on more than  15 trillion tokens. While traditional statistical learning theory suggests that overly parameterized models are prone to overfitting, recent studies \citep{belkin2019reconciling,nakkiran2019deepdoubledescentbigger,power2022grokking} show that such models, when combined with massive training datasets and appropriate optimization techniques, often exhibit enhanced generalization capabilities. This behavior aligns with the \textbf{double descent} phenomenon \citep{nakkiran2019deepdoubledescentbigger}, where increasing model capacity beyond a critical threshold leads to a second phase of performance improvement, contrary to classical expectations.

\paragraph{Dimension Expansion}
Dimension expansion in LLMs refers to the mapping of discrete input tokens into a high-dimensional continuous embedding space, where the embedding dimension $p $, ($d_\text{model})$  far exceeds both the intrinsic dimension $p^*$ and input sequence length $T$ ($ p \gg \min (p^*, T) $). Unlike traditional statistical language models, such as Latent Dirichlet Allocation (LDA) or Latent Semantic Analysis (LSA), which reduce dimensionality ($ p \ll \min (p^*, T) $), LLMs increase dimensionality to capture fine-grained, context-specific details. In contrast to the traditional numerical data settings where the dimensionality of the data is fixed by intrinsic properties (e.g., features or variables), natural language does not have a clear starting dimensionality ($p^*$). For instance, vocabulary size, while sometimes considered a proxy, is a discrete coding scheme and not a true geometric space. 

High-dimensional embeddings in LLMs enable rich representations by separating semantically or syntactically distinct tokens. The computational complexity of the Transformer's attention mechanism, $O(T^2p)$, allows efficient modeling of relationships between tokens, as opposed to the $O(Tp^2)$ complexity of recurrent neural networks (RNNs) \citep{vaswani2017attention}. For example, in GPT-3, $T = 2,048$ is much smaller than $p = 12,288$, highlighting the reliance on high-dimensional embeddings to model linguistic patterns. While this approach increases model capacity, it also introduces a computational challenge due to the quadratic scaling of cost with $T$. Solutions such as ``context parallelism," introduced by \cite{meta2024llama3}, extend sequence lengths up to $T = 128,000$ while scaling $p$ to 53,248. Other approaches, including linear-time sequence models \citep{gu2024mambalineartimesequencemodeling} and efficient RNN-based methods \citep{feng2024rnnsneeded}, aim to reduce computational overhead while preserving the benefits of dimensionality expansion.

\paragraph{Massive Training Data}
Large Language Models (LLMs) such as GPT \citep{radford2018improvingLU, radford2019gpt2, brown2020gpt3}, BERT \citep{devlin2019bert}, RoBERTa \citep{liu2019robertarobustlyoptimizedbert}, BART \citep{lewis2019bart}, and LLaMA \citep{touvron2023llama} rely heavily on massive training datasets sourced from diverse corpora, including \textit{BookCorpus, Wikipedia, Common Crawl}, and other large-scale textual collections (See Table~\ref{tab:summary-lm} for details). Studies such as \citet{hoffmann2022chinchilla} emphasize the importance of balancing model size $N_\text{param}$ and dataset size $|\mathcal{D}_\text{pretrain}|$, encapsulated by the computational cost formula $C = N_\text{param} \times |\mathcal{D}_\text{pretrain}|$. For instance, Chinchilla demonstrates superior efficiency compared to GPT-3 by reducing the parameter count to 70 billion while increasing the training data to 1.4 trillion tokens.  

\paragraph{Connection Under Double Decscent Paradigm}

The interplay between dimension expansion, and massive training data is best understood under the frameworks of \textbf{double descent} \citep{nakkiran2019deepdoubledescentbigger} and \textbf{grokking} 
 \citep{power2022grokking}. Double descent describes the modern regime where the traditional bias-variance tradeoff cannot fully explain generalization, as generalization error decreases again after the interpolation threshold. 
The grokking phenomenon complements double descent by illustrating how models trained on limited data eventually transition from memorization to generalization. While double descent is primarily driven by increasing model capacity or dataset size, grokking highlights the role of prolonged optimization and regularization in relatively small data regimes. Dimension expansion supports generalization by allowing high-dimensional embeddings to allocate capacity for separating patterns in the data, a mechanism that may contribute to phenomena like grokking. Meanwhile, massive datasets stabilize optimization by providing broad coverage of the underlying distribution, making grokking less necessary in large-scale training. Together, these frameworks explain the surprising generalization capabilities of overparameterized models in the context of LLMs.

\begin{table*}[t]
\centering
\resizebox{\textwidth}{!}{%
\begin{tabular}{@{}ccccccccc@{}}
\toprule
\textbf{Model} & 
\textbf{Developer} &
\textbf{Architecture} & 
\textbf{\begin{tabular}[c]{@{}c@{}}Release \\ Date\end{tabular}} &
\textbf{\begin{tabular}[c]{@{}c@{}}Sequential \\ Input\end{tabular}} &
\textbf{Params} & 
$\mathbf{T}$ &
$\mathbf{p}$ &
\textbf{Citation} \\
\midrule

Word2Vec & 
Google &
Shallow Neural Network & 
2013 &
X & 
- & 
5-10 &
300 &
\cite{mikolov2013efficient} \\

GloVe & 
Stanford &
Shallow Neural Network & 
2014 &
X & 
- & 
5-10 &
300 &
\cite{pennington2014glove} \\

BERT & 
Google &
Transformer Encoder & 
2018-10 &
O & 
110M/340M & 
512 &
1024 &
\cite{devlin2019bert} \\

RoBERTa & 
Meta & 
Transformer Encoder & 
2019-07 &
O & 
355M & 
512 &
1024&
\cite{liu2019robertarobustlyoptimizedbert} \\

GPT-3 & 
OpenAI &
Transformer Decoder & 
2022-03&
O & 
175B & 
2048&
12288 &
\cite{brown2020gpt3} \\

GPT-4 & 
OpenAI &
Transformer Decoder & 
2023-03 &
O & 
NA & 
8192 &
Unknown &
\cite{openai2023gpt4} \\

LLaMA-2 & 
Meta & 
Transformer Decoder & 
2023-07&
O & 
7B/13B/70B & 
4096 &
8192 &
\cite{touvron2023llama} \\

LLaMA-3 & 
Meta & 
Transformer Decoder & 
2024-04 &
O & 
8B/70B & 
128K &
16384 &
\cite{meta2024llama3} \\
\bottomrule
\end{tabular}%
}
\caption{Comprehensive Language Model Comparison. All specifications are sourced from original papers and official releases. Parameter counts (Params) show base/large versions where applicable. Training data sizes are reported in the units used in original publications. The context length (T) refers to maximum input sequence length during pre-training.}
\label{tab:summary-lm}
\end{table*}

\section{Proposed Approach} \label{sec:proposed-approach}

There are a few resources for digitized versions of the Federalist papers, and for our analysis, we use R \verb|syllogi| packages. The data is processed based on \href{https://www.gutenberg.org/ebooks/1404}{ProjectGutenberg} so that each text and metadata are encoded as the list element for each document. There are a total of 86 list documents because there are two different versions of No. 70 offered in \href{https://www.gutenberg.org/ebooks/1404}{ProjectGutenberg}, and we use the first version of No. 70. Our proposed approach is to use the off-the-shelf language model as an embedding generator, a function to transform non-euclidean text data into $p$-dimensional vector representation and use statistical classifiers to make the decision capable of quantifying the uncertainty.

\subsection{Text Preprocessing}
Text preprocessing is a crucial step in natural language processing (NLP) that transforms raw text data into a structured format suitable for statistical and machine learning models. 
Key preprocessing steps include tokenization, lowercasing, removing punctuation, stopword removal, stemming, lemmatization, and converting text to numerical representations. More details on converting text to numerical representations are provided in Sections ~\ref{sec-bow} and ~\ref{sec-embedding}.

Tokenization is the process of breaking down text into smaller units called tokens. Tokens can be words, subwords, or characters, depending on the chosen granularity. Tokenization is the first step in text preprocessing and lays the foundation for subsequent steps. There are several approaches to tokenization. For example, word tokenization splits text into individual words, as in the Bag-of-Words approach (e.g.,``Tokenization is important." becomes [``Tokenization", ``is", ``important", ``."]). The unit to be tokenized can be larger (e.g., sentence) or smaller than a word (e.g., subword or character).

Once tokenization is complete, the input text can be further processed by converting all characters to lowercase and removing punctuation. Stemming and lemmatization can also be applied; these are both text normalization techniques that reduce words to their base or root form. Stemming is an algorithmic process that involves removing suffixes from words to arrive at a base form, often resulting in stems that may not be real words. For example, the words ``running", "runner", and ``runs" might all be reduced to ``run", while ``better" might be reduced to ``bet". This method is generally faster and less computationally intensive but can be crude, producing non-standard word forms. In contrast, lemmatization uses a dictionary and morphological analysis to reduce words to their base or dictionary form, known as the lemma. This process takes into account the context and part of speech of the word, ensuring that the resulting lemma is a valid word. For example, ``running" would be reduced to ``run", and ``better" would be reduced to ``good". Therefore, while stemming is faster and simpler, it is less precise, whereas lemmatization is slower but more accurate and produces meaningful root words. In our work, we perform lemmatization using the \verb|lemmatize_strings| function from the R \verb|textstem| package.

It is common practice to remove stopwords, which may not contribute much to the meaning of the text (e.g.,``and",``the", ``is"), in many NLP applications. However, as \cite{Mosteller1963Inference} pointed out in their analysis, stopwords (referred to as function words in their paper) can reflect the stylistic features of a specific author. For the purpose of authorship attribution, we test three different sets of words: (1) the set of words without stopwords (5,834 words), (2) the set of words with stopwords (5,936 words), and (3) the set of words selected in the previous study by \cite{Mosteller1963Inference} (145 words). Detailed specifications of the words used by Mosteller and Wallace and the stopwords used to process the word count matrix are given in Appendix \ref{app-words}.


\subsection{Bag-of-Word Approach}\label{sec-bow}
\textit{Bag-of-Words (BoW)} represents text as a vector of word counts or frequencies, where the order of words is ignored. This representation serves as the foundation for various document embedding methods based on matrix factorization. Given a corpus of n documents and a vocabulary of size N, we define the BoW matrix as $X \in \mathbb{R}^{n \times N}$, where each row corresponds to a document and each column represents a unique word in the vocabulary. The choice of words included in this matrix significantly influences the resulting embeddings, affecting whether the model captures topical, stylistic, or structural properties of the text.

To examine how different linguistic properties influence document representations, we construct three variations of the term-document matrix. The first variation, referred to as \textit{Type 1}, includes all words from the corpus except for common stop words. This representation preserves both stylistic and topical differences, allowing for a broad semantic analysis of document content. The second variation, \textit{Type 2}, retains contextual words as well as common stop words. By emphasizing content words, this representation highlights thematic differences across documents while reducing stylistic variation. The third variation, \textit{Type 3}, is constructed using Mosteller and Wallace's list of function words, which include determiners, prepositions, conjunctions, and other non-content words. Since function words are largely independent of document topics, this representation captures the structural and stylistic characteristics of the text rather than its meaning. The Type 3 representation is particularly relevant for authorship attribution, as prior studies have shown that function words serve as reliable markers of individual writing styles. Notably, the similarity between the topic distributions of Madison's papers and the disputed papers when using Type 3 input suggests that these documents share common stylistic features, providing further evidence in support of Madison's authorship.

To obtain low-dimensional representations from the BoW matrix $X$, we seek a decomposition of the form $X = SH$ where $S \in \mathbb{R}^{n \times p}$ and $H \in \mathbb{R}^{p \times N}$. There are three different methods that fall into this category: (1) Latent Dirichlet Allocation (LDA), (2) Latent Semantic Analysis (LSA), and (3) Nonnegative Matrix Factorization (NMF), or equivalently pLSA. The first method is based on probabilistic generative modeling assumption, which is reviewed in detail in Section~\ref{sec:review}. LSA is a direct application of Singular Value Decomposition (SVD) on $X$. NMF is a purely numerical method for matrix decomposition.

Instead of trying to analyze the LDA result as is, we want to view this technique as dimensionality reduction onto the document spaces as LSA does. With the lower dimensional representation of documents, given as document-topic distribution, we perform the classification analysis. The same principle is applied to LSA and NMF. For LDA, we tested different numbers of topics ($p = 5, 10, 15, 20, 25$), and we chose the model with 5 topics based on BIC. For LSA and NMF, we use the embedding dimension to be 10.

\subsection{Continuous Embedding Approach}\label{sec-embedding}
The word embedding approach can be viewed as a continuous extension of Bag-of-Words. It converts the text into a dense vector representation, preserving the semantic relationships. Word2Vec \citep{mikolov2013distributed, mikolov2013efficient} or GloVe \citep{pennington2014glove} is a direct extension of the Bag-of-Words approach that does not consider the order of the word. On the other hand, a recent Large Language Model like BERT \citep{devlin2019bert} takes the sequence of the text as an input and outputs the embedding for the sequence. In this study, we consider 7 different models to generate continuous embeddings. For LLMs, we consider BERT \citep{devlin2019bert}, sentence transformer (or sentence BERT) \citep{reimers2019sentencebert}, RoBERTa \citep{liu2019robertarobustlyoptimizedbert}, BART \citep{lewis2019bart}, GPT \citep{brown2020gpt3} and the most recent Llama2, Llama3 \citep{touvron2023llama}.

\subsubsection{Aggregation of Embeddings}\label{sec:aggregation}

Different models generate embeddings based on various token units, requiring an aggregation scheme to obtain document-level embeddings. All the continuous embedding models we consider process chunks of text and generate embeddings for each token that the output embeddings for the document are not matrix anymore, but a tensor. For example, BERT model outputs 768 dimension embeddings of each token, $Z_\text{BERT} \in \mathbb{R}^{n \times N \times 768}$. However, to analyze the text at the document level, it becomes necessary to implement a scheme that aggregates these token-level tensor embeddings into document-level representations. This aggregation can be done in various ways, ensuring that the smaller units (tokens, sentences, or chunks) are effectively combined to capture the overall meaning and structure of the document.

The simplest approach is to average token embeddings to represent a larger unit (sentence or document), when using models like BERT, GPT, BART, or LlaMA. Averaging can provide a rough approximation of sentence meaning, but it is not the most optimal approach. These models generate contextual embeddings, where each token's representation is influenced by the surrounding context. Averaging also implicitly assumes independence among smaller units, and treats them as equally important, which may not hold for structured documents where the sequence or the relationship among them matters. A model like Sentence BERT \citep{reimers2019sentencebert} can be a better alternative for getting sentence-level embeddings, as the model is designed to capture sentence-level information by utilizing the output of a specific token in BERT.

Aggregation can be implemented at two stages: preprocessing (embedding aggregation) or postprocessing (probability aggregation). For preprocessing, document representations can be constructed by averaging token embeddings at the word, sentence, or chunk level. Word2Vec specifically allows for frequency-weighted averaging: given a row-normalized bag-of-words matrix $\tilde{X} \in \mathbb{R}^{85 \times N}$ and word embeddings $Z_\mathcal{W} \in \mathbb{R}^{N \times 300}$, document embeddings are computed as $Z_D = \tilde{X}Z_\mathcal{W} \in \mathbb{R}^{85 \times 300}$. For postprocessing, document-level predictions can be obtained by averaging token-level probabilities or through majority voting of individual predictions across different granularities (word/sentence/chunk). The effectiveness of these aggregation schemes is demonstrated in Figure~\ref{fig:sentence} and Figure~\ref{fig:chunk} in the supplement.

\subsubsection{Tune or Not to Tune}

Fine-tuning refers to the process of adapting a pre-trained model to a specific task by continuing its training on a smaller, task-specific dataset. This allows the model to leverage the broad knowledge learned from a large dataset (used during pre-training) and specialize in the nuances of the target task. 
\cite{yosinski2014transferablefeaturesdeepneural} explored how features learned in pre-trained models are transferable to new tasks, refining them through fine-tuning. \cite{howard2018universal}, with their work on ULMFiT, demonstrated the effectiveness of fine-tuning in text classification, while \cite{zhuang2020comprehensivesurveytransferlearning} provided a comprehensive review of fine-tuning in transfer learning, emphasizing its benefits and challenges. BERT \citep{devlin2019bert} shows pre-training and fine-tuning significantly improve NLP tasks. GPT-3 \citep{brown2020gpt3} further demonstrates that even without fine-tuning, large-scale pre-trained models can perform well on a wide variety of tasks with little task-specific data, which highlights the power of large pre-trained models and the potential to fine-tune for even higher performance.

\begin{figure}
    \begin{subfigure}{.49\textwidth}
        \centering
        \includegraphics[width=.95\linewidth]{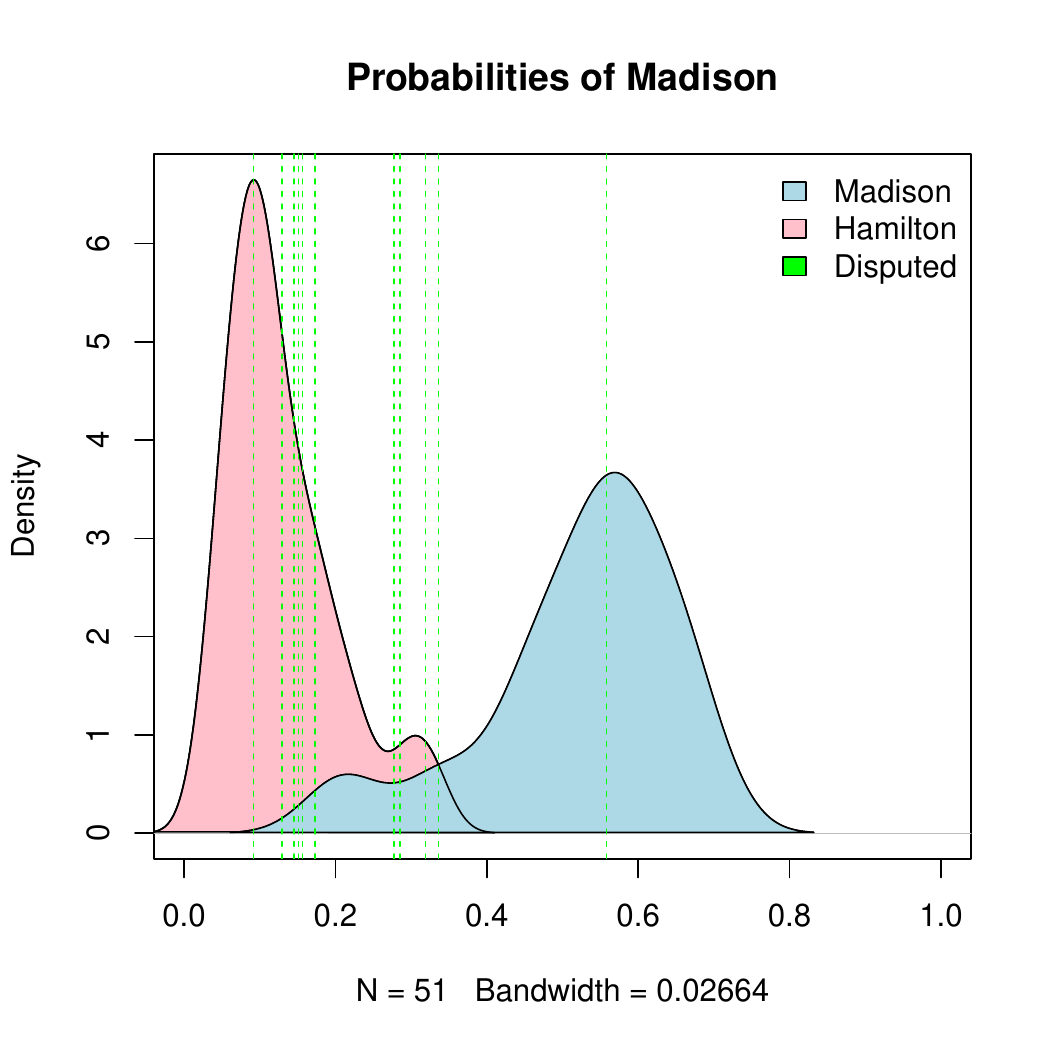}
        \caption{Vanilla embedding (Without fine-tuning)}
    \end{subfigure}
    \begin{subfigure}{.49\textwidth}
        \centering
        \includegraphics[width=.95\linewidth]{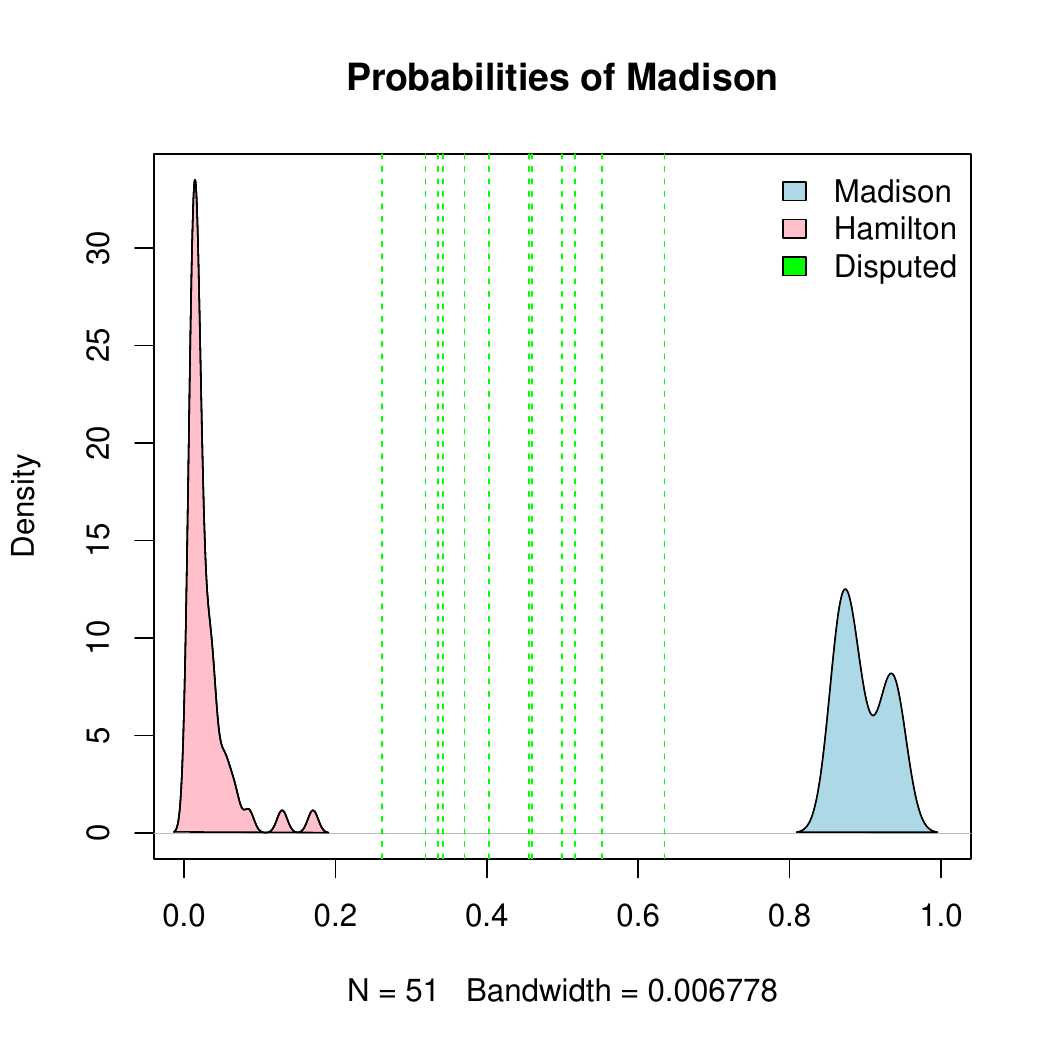}
        \caption{Fine-tuned embedding through classification}
    \end{subfigure}
    \caption{The estimated density for Hamilton and Madison using BART with BERT embeddings is shown. While the fine-tuned embeddings yield perfectly separated density estimates for the training data, the unseen documents (indicated by green vertical lines) do not fall within the estimated density regions but instead lie in the intermediate space. This suggests overfitting during the fine-tuning process.}
    \label{fig:bert-with-finetuning}
\end{figure}

Building on these insights, we apply fine-tuning to the Federalist Papers authorship attribution task. First, we parse the 85 Federalist Papers into 5,738 sentences. Focusing only on the papers authored by Hamilton or Madison, we obtain 4,523 sentences. These are randomly split into a training set of 3,618 sentences and a test set of 905 sentences. Using this dataset, we update the parameters of basic models with a classification task. Afterward, we generate embeddings for each document, as before. The fine-tuning leads to overfitting to the training data (Figure~\ref{fig:bert-with-finetuning}). That is, the predicted probabilities from BART, using the fine-tuned embeddings, provide a better separation of the papers by known authorship; the fine-tuned embeddings are not as informative as in unseen disputed papers.

\subsection{Classification Methods}\label{sec-classifier}
In the previous section (Section~\ref{sec:proposed-approach}), we described the procedure to encode the text data into a numeric vector (``embedding"). In this section, we will elaborate on the model we use to determine the authorship based on these embeddings. We aim to focus on the effect of different types of embeddings in the authorship prediction task, so we limit our choice of classifiers to rather classical ones: the least absolute shrinkage and selection operator (LASSO) \citep{tibshirani1996lasso} and BART \citep{Chipman2010bart}. 
Note the BART \citep{Chipman2010bart} here refers to the Bayesian Additive Regression Tree, which has the same acronym as LLM BART (Bidirectional and Auto-Regressive Transformers) \citep{lewis2019bart}. Most of the context they will be distinguished without any confusion as LLM will be used to extract the embeddings while the classifier BART will be used for prediction.


\paragraph{LASSO} 
The Lasso \citep{tibshirani1996lasso} is a regression technique that imposes an $\ell_1$-norm penalty on the model parameters, enabling both regularization and feature selection. In logistic regression, Lasso introduces sparsity by shrinking some $\beta$ coefficients to exactly zero. The objective function becomes
$$\hat\beta = \underset{\beta}{\text{argmin}} \left( - \sum_{i=1}^n \Big( y_i \log(p_i) + (1 - y_i) \log(1 - p_i) \Big) + \lambda \|\beta\|_1 \right),$$
where $\lambda$ controls the strength of the regularization. This approach is particularly valuable in high-dimensional settings, as it reduces model complexity and enhances generalization.

In the Bayesian framework, the Bayesian Lasso uses a Laplace (double-exponential) prior, $p(\beta_j | \lambda) = (\lambda/2) \exp(-\lambda |\beta_j|)$,
where $\lambda$ corresponds to the classical Lasso's regularization parameter, encouraging sparsity by shrinking small coefficients toward zero \citep{park2008bayesian}. The Spike-and-Slab Lasso (SSLASSO) \citep{rovckova2018spike} extends this idea by using a hierarchical prior,
$$p(\beta_j | \gamma, \lambda_0, \lambda_1) = (1-\gamma_j) \cdot \text{Laplace}(\beta_j|\lambda_0) + \gamma_j \cdot \text{Laplace}(\beta_j|\lambda_1),$$
where \(\lambda_0\) (spike) encourages coefficients to be exactly zero, and \(\lambda_1\) (slab) allows non-zero coefficients with controlled shrinkage. The spike-and-slab structure enables adaptive sparsity, accounting for dependencies among covariates and improving over standard Lasso.

In the authorship attribution context, without preselection of words, the LASSO model successfully identifies the set of words that were found to be significant in \cite{Mosteller1963Inference}. For example, when training the binary lasso model using the 65 papers with known authorship, `whilst' turns out to be the the word with the highest coefficient in absolute value (.57), among 10 words selected by the model. We used \verb|gamlr| function from \verb|gamlr| package for the lasso with binary responses. 

\paragraph{Bayesian Additive Regression Tree (BART)} 
BART is a non-parametric regression technique introduced by \cite{Chipman2010bart}. It is an ensemble method that combines Bayesian inference with decision trees, providing a tool for predictive modeling. Unlike traditional regression methods, BART does not assume a specific functional form for the relationship between the predictors and the response variable, making it adaptable to various data structures.

A binary classification with BART \citep{Chipman2010bart}, the probability of the binary outcome $Y \in \{0, 1\}$, is modeled using a latent variable approach. The probability is linked to the predictor variables through the \textit{probit} function, which uses the cumulative distribution function (CDF) of the standard normal distribution, denoted as $ \Phi(\cdot)$. The model is expressed as:
$$P(y_i = 1 \mid \mathbf{x}) = \Phi\left( \sum_{j=1}^{m} g(\mathbf{x},T_j,M_j) \right)$$
where $g(\cdot, T_j, M_j)$ represents the $j$-th additive trees, $T_j$ represents the tree structure (i.e., how the predictor space is partitioned), and $ \mathcal{M}_j$ denotes the set of terminal node parameters (i.e., the predicted values in the leaves of the tree), and $m$ is the number of trees in the ensemble, where $g(\mathbf{x}; T_j, M_j)$. The use of the probit function $ \Phi \left( \cdot \right)$ implies a latent variable formulation where the binary outcome depends on an underlying continuous latent variable that is normally distributed.

We used \verb|gbart| function from \verb|BART| package for the BART with binary responses.

\subsection{Determining the Classification Threshold}

For simplicity, we formulate the problem as a binary classification task, where the goal is to determine whether the author of the disputed papers is Hamilton or Madison. Both models compute the predicted probability of the given paper being authored by Madison, and the final binary prediction is dependent on the choice of the threshold. 

A natural way to classify documents is to use the kernel-smoothed densities of the estimated probabilities of Madison for the Madison and Hamilton labeled samples. If the estimated densities have a common support and the densities are close to unimodal (i.e. the densities overlap on $[0,1]$ just like on Figure \ref{fig:uq-prediction} (b)), the intersection point between the two densities could be a suitable classification threshold. This would mean that the estimated density ratio for a predictive Madison probability of a disputed paper would indicate evidence for (if greater than one) or against (if smaller than one) Madison. However, when the densities do not overlap (just like on Figure \ref{fig:uq-prediction} (a)), the denominator of the density ratio approaches zero for one of the classes, making the classification decision ill-posed. In such cases, an alternative method is needed to determine authorship.

A straightforward alternative is to use a simple thresholding approach, where we classify a document as Madison's if the mode'’s predicted probability exceeds a fixed threshold, such as $t = 0.5$. However, this may not be optimal due to the class imbalance in the dataset. Out of 85 papers, 51 were written by Hamilton, 14 by Madison, 3 by Jay, and 12 remain disputed, while the remaining 3 were jointly authored by Hamilton and Madison. Given that Hamilton's papers significantly outnumber Madison's, a threshold of 0.5 may systematically favor Hamilton, leading to biased classifications. 

To address both the instability of the density ratio in non-overlapping regions and the suboptimality of a naive threshold due to class imbalance, we employ thresholding strategies based on the Receiver Operating Characteristic (ROC) curve and the $F_1$ score. These methods do not directly rely on density estimation but instead optimize the classification threshold using performance metrics.

In binary classification, given predicted and true labels, each outcome falls into one of four categories: True Positive (TP), False Positive (FP), True Negative (TN), and False Negative (FN). The ROC curve plots recall $\Big(\frac{TP}{TP+FN}\Big)$ against specificity $\Big(\frac{TN}{TN+FP}\Big)$. We select the threshold that maximizes Youden's J statistic \citep{Youden1950Index}, defined as the sum of recall and specificity. This method balances sensitivity and specificity, making it well-suited for cases where false positives and false negatives have similar costs.

Alternatively, we consider the $F_1$ score, which balances precision $\Big(\frac{TP}{TP+FP}\Big)$ and recall $\Big(\frac{TP}{TP+FN}\Big)$ by computing their harmonic mean. Unlike ROC-based selection, the $F_1$ score is particularly effective in imbalanced datasets, where overall accuracy can be misleading due to class dominance. By emphasizing the minority (positive) class, the $F_1$ score ensures that both false positives and false negatives are controlled.
  
For each model, we determine the optimal threshold and compute the corresponding classification error. In most cases, the selected thresholds are around 0.3, which is close to the empirical class ratio, $\frac{14}{51} \approx 0.27$. The specific thresholds obtained using the ROC and $F_1$ criteria for each embedding and classifier are reported in Table~\ref{tab:threshold} in the supplement.  

\section{Analysis of Results} \label{sec-res}


In this section, we will present and analyze the authorship attribution task focusing on the effect of different embeddings. Our framework generates different types of embeddings using various techniques introduced in Section~\ref{sec:review}, then trains the classifiers introduced in Section~\ref{sec-classifier}. We analyze the authorship attribution from various perspectives in terms of their topics, $\ell_2$ loss, classification accuracy, and variable selections.

\subsection{Topics}\label{sec:topics}
It is natural to suspect that different authors may have specialized in particular topics, potentially dividing their work according to areas of expertise. In Section~\ref{sec:small-lm}, we discussed Latent Dirichlet Allocation (LDA) \citep{blei2003latent} as a widely used probabilistic model for discovering latent topic structures within a collection of texts. We apply LDA with different term-document matrix constructions (\textit{Type 1, Type 2, Type 3}) introduced in Section~\ref{sec-bow} to examine whether the inferred topic distributions align with known authorship patterns and whether they provide additional insight into the stylistic and thematic characteristics of the disputed papers.

\begin{figure}
    \centering
    \includegraphics[width=\textwidth]{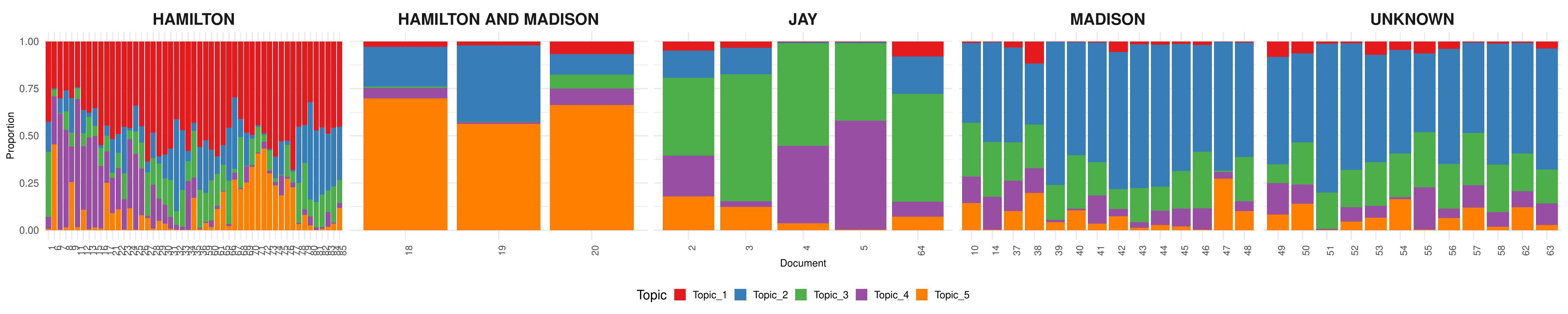}
    \caption{Document-Topic distribution by LDA trained on 145 selected words (Type 3). The similarity in topic distributions for Madison authored papers and the disputed papers implies the shared stylometry among them.}
    \label{fig:doc-topic-distribution-tdm3}
\end{figure}

We first consider a term-document matrix constructed using Mosteller and Wallace's list of function words (Type 3 input), as shown in Figure~\ref{fig:doc-topic-distribution-tdm3}. Unlike the other two representations (Type 1 or Type 2), this input excludes contextual words that convey meaning, focusing instead on the structural and stylistic aspects of the text. Consequently, the extracted topics do not correspond to semantic themes but instead capture latent stylistic structures. Notably, the topic distributions of the papers authored by Madison and the disputed papers appear surprisingly similar, which is consistent with prior stylometric analyses supporting Madison's authorship of the disputed papers \citep{Mosteller1963Inference}. The effect of dimensionality reduction is also apparent, as the term-document matrix is constructed from a vocabulary of over 145 words, yet the inferred topic space is compressed into a much lower-dimensional representation ($p = 5$).

A more interpretable document-topic distribution emerges when retaining contextual words in the term-document matrices (Type 1 and Type 2 inputs), which we present in the supplement (Figures~\ref{fig:doc-topic-distribution-tdm1} and~\ref{fig:doc-topic-distribution-tdm2}). In these representations, certain topics align well with known themes in the Federalist Papers. For example, Papers No. 2 to No. 5, authored by Jay and titled `Concerning Dangers from Foreign Force and Influence', primarily discuss foreign policy, a theme that is reflected in the document-topic distributions. In Type 1, Topic 3 (green) appears to capture foreign policy, while in Type 2, Topic 4 (purple) represents the same theme. Interestingly, Paper No. 64 (The Powers of the Senate), which differs in content from Jay's other papers, exhibits a distinct topic distribution in both representations.

For papers authored by Hamilton and Madison, the topic distinctions are less pronounced. This is particularly evident for the disputed and jointly authored papers, where topic assignments remain ambiguous. In both Type 1 and Type 2 inputs, the disputed papers are primarily dominated by Topic 1, while the jointly authored papers are mostly assigned to Topic 4. The lack of a clear distinction suggests that topic-based clustering alone may not provide conclusive evidence for authorship attribution. 

Overall, the results highlight the contrast between content-based and style-based document representations. The similarity in distributions for Madison and the disputed papers (Type 3) aligns with the hypothesis that Madison authored the disputed papers, as function words are stable markers of authorship. In contrast, the contextual-word representations (Types 1 and 2) reveal meaningful thematic differences but still struggle to cleanly separate the writings of Hamilton and Madison. The complete document-topic distributions for these alternative representations are provided in the supplement.

\subsection{Word Screening and Variable Selection}\label{sec:word-screening}

The findings from the topic modeling analysis suggest that different input representations capture distinct linguistic characteristics. Function-word-based representations (Type 3) highlight stylistic elements, whereas contextual-word-based representations (Type 1 and Type 2) emphasize semantic themes. To further investigate how input selection influences authorship attribution, we perform word screening to identify the most discriminative words.

\begin{figure}
    \centering
    \begin{subfigure}{.32\textwidth}
    \includegraphics[width=\linewidth]{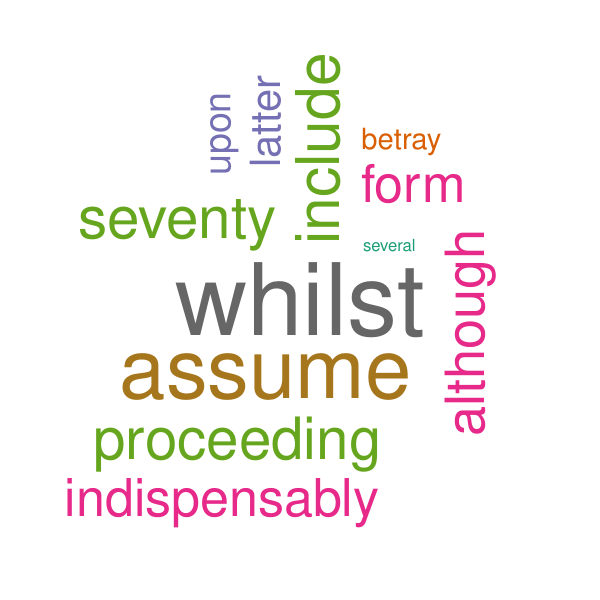}
        \caption{Type 1}
    \end{subfigure}
    \begin{subfigure}{.32\textwidth}
        \includegraphics[width=\linewidth]{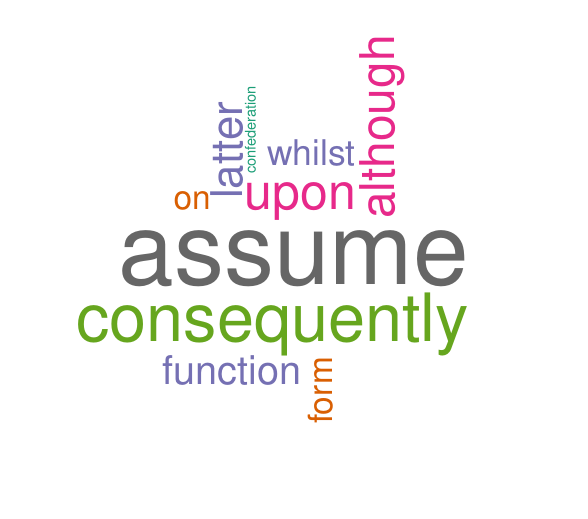}
        \caption{Type 2}
    \end{subfigure}
   \begin{subfigure}{.32\textwidth}
        \includegraphics[width=\linewidth]{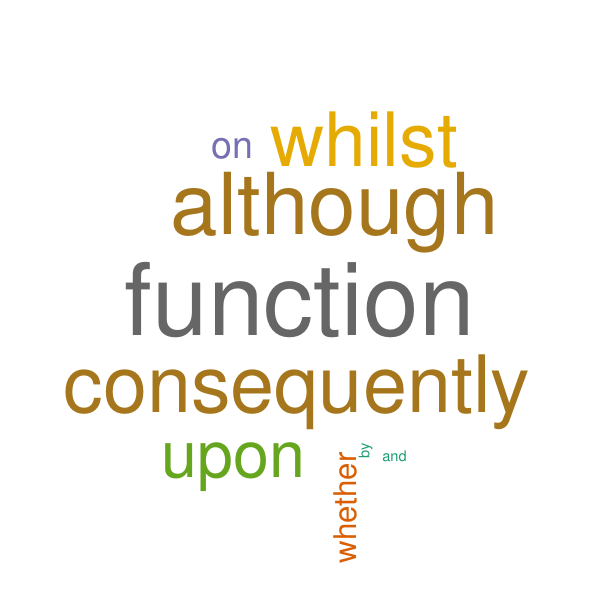}
        \caption{Type 3}
    \end{subfigure}
    \caption{Word cloud representation of significant words found by LASSO with different sets of word count matrices. Type 1 includes contextual words only, Type 2 includes both contextual words and stopwords, and Type 3 includes a curated set of words by \cite{Mosteller1963Inference}. LASSO successfully recovers some of the words reported in the original study (Table~\ref{tab:function-words}). The most discriminative words such as `whilst' or `upon' are consistently recovered in all three types of inputs. }
    \label{fig:words-clouds-lasso}
\end{figure}

Mosteller and Wallace \citep{Mosteller1963Inference} demonstrated that function words serve as stable indicators of authorship. The key idea is to test for the heterogeneity of word usage across the documents. The usage of a word like `war' varies across the documents by their topic, and it is less likely to be attributed to the characteristics of certain authors. They ended up with 30 words whose usage is homogeneous across the document given an author (Table~\ref{tab:function-words}). Using LASSO for feature selection, we analyze whether different word sets yield meaningful patterns in authorship prediction. Figure~\ref{fig:words-clouds-lasso} presents the words selected by LASSO across different input types. The function-word-based representation recovers many of the same words used in the original study, reaffirming the importance of stylistic markers in authorship attribution. Table~\ref{tab:bow-approach-diff-inputs-l2} shows that restricting the input to curated function words (Type 3) significantly improves $\ell_2$ loss. Notably, more complex methods like Word2Vec do not necessarily outperform simpler approaches; in fact, they often underperform compared to the basic Bag-of-Words input. 

Overall, the careful selection of words within each method significantly impacts the results. This raises the natural question: how can we create a well-curated list of words? Inspired by the approach in \cite{Kipnis2022Higher}, where HC statistics are viewed as distance measures and the corresponding HC threshold is used to identify the most discriminative words, we apply multiple-testing methods such as Bonferroni and Benjamini-Hochberg procedures to refine our word selection. The results are presented in Figure~\ref{fig:words-clouds-mt} and detailed in the supplement.

\begin{table}[]
\resizebox{\textwidth}{!}{%
\begin{tabular}{@{}ccccccccccc@{}}
\toprule
       & \multicolumn{5}{c}{LASSO}                    & \multicolumn{5}{c}{BART}                              \\ \midrule
       & BoW    & LDA    & LSA    & NMF    & Word2Vec & BoW    & LDA    & LSA             & NMF    & Word2Vec \\ \midrule
Type 1 & 0.1270 & 0.1745 & 0.1756 & 0.1626 & 0.1709   & 0.0935 & 0.1464 & 0.0962          & 0.1385 & 0.1414   \\
Type 2 & 0.1134 & 0.1101 & 0.0222 & 0.0759 & 0.1668   & 0.1101 & 0.0735 & \textbf{0.0579} & 0.0892 & 0.1252   \\
Type 3 &
  \textbf{0.0486} &
  \textbf{0.0022} &
  \textbf{0.0093} &
  \textbf{0.0094} &
  \textbf{0.0749} &
  \textbf{0.0475} &
  \textbf{0.0106} &
  0.0664 &
  \textbf{0.0463} &
  \textbf{0.0884} \\ \bottomrule
\end{tabular}%
}
\caption{$\ell_2$ loss for Leave-One-Out Cross-Validation (LOOCV) on 65 training datasets. The bold number indicates the lowest (best) value for a given method. Note that the error varies by the choice of words for all four non-sequential methods. Among the methods and input types compared, LDA demonstrates the best performance, even achieving an $\ell_2$ loss close to zero ($0.0022$) when well-curated list of words (Type 3) are used as inputs. }
\label{tab:bow-approach-diff-inputs-l2}
\end{table}

\subsection{Authorship Prediction and Model Comparison}

Building on the word screening analysis, we now evaluate how different embedding methods impact authorship attribution in a binary classification setting. We frame the problem as a binary task, where 1 indicates Madison as the author and 0 indicates Hamilton. The training dataset consists of 65 papers with known authorship, while the 12 disputed papers form the test set. We exclude five papers authored by John Jay. For the three jointly authored papers (No. 18, 19, 20), a detailed analysis is provided in Section~\ref{sec:discussion}.

To assess classifier performance, we conduct leave-one-out cross-validation (LOOCV) on the training data and compute the $\ell_2$ loss for each embedding method. The results are summarized in Table~\ref{tab:l2_loss_for_all}. Among Bag-of-Words (BoW) embeddings, LDA achieves the lowest $\ell_2$ loss, while LSA and NMF also perform well with the LASSO classifier. Among the Bag-of-Words (BoW) embeddings, LSA produced the best results, while LDA and NMF also performed well when using BART as the classifier. For continuous embedding techniques, Word2Vec performed the best with the BART classifier, and RoBERTa with the LASSO classifier. However, even the best continuous embeddings (RoBERTa with LASSO) did not perform as well as the least effective BoW embeddings (BoW with LASSO), highlighting the effectiveness of explicit word frequency representations in this context.

\begin{table}
\resizebox{\textwidth}{!}{%
\begin{tabular}{@{}cccccccccccc@{}}
\toprule
 & \multicolumn{4}{c}{BoW Embeddings} & \multicolumn{7}{c}{Continuous Embeddings}                 \\ \midrule
 & BoW     & LDA    & SVD    & NMF    & Word2Vec & BERT & RoBERTa & BART & GPT4 & Llama2 & Llama3 \\ \midrule
LASSO & 0.1134 & 0.1101 & \textbf{0.0222} & 0.0759 & 0.1668 & 0.1281          & \textbf{0.1198} & 0.1613 & 0.1735 & 0.1778 & 0.1742 \\
BART  & 0.1101 & 0.0735 & \textbf{0.0579} & 0.0892 & \textbf{0.1252} & 0.1349 & 0.1441          & 0.1522 & 0.1540 & 0.1400 & 0.1559 \\ \bottomrule
\end{tabular}%
}
\caption{Leave-one-out cross-validation $\ell_2$ loss for 65 papers with known authorship. The lowest (best) $\ell_2$ value is bolded. For the Bag-of-Words type of embeddings, the results out of the Type 2 term document matrix are included. Among LLMs, the encoder-based models (BERT, RoBERTa) perform better than the decoder-based models (GPT, Llama).}
\label{tab:l2_loss_for_all}
\end{table}

\begin{figure}
    \centering
    \begin{subfigure}{.32\textwidth}
    \centering
    \includegraphics[width=.95\linewidth]{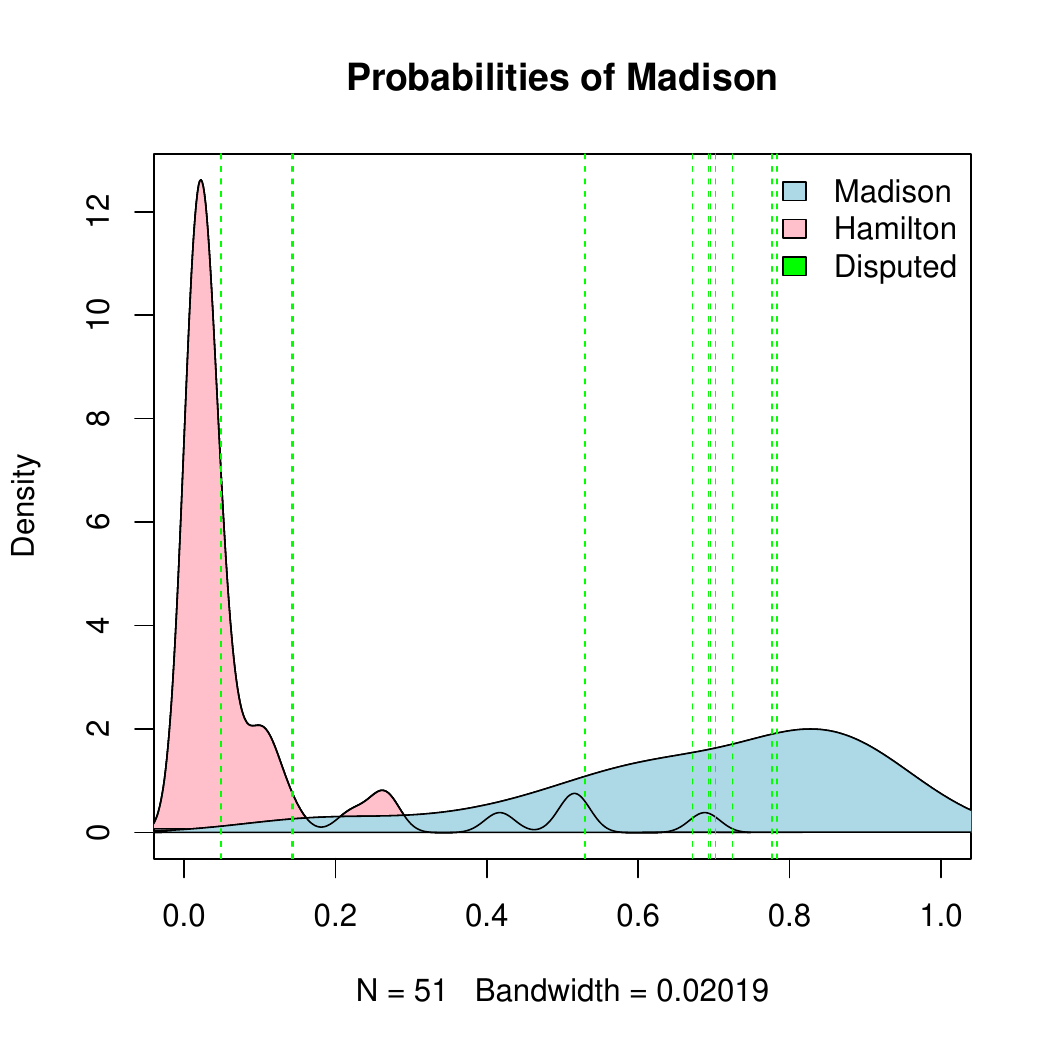}
    \caption{Bag-of-Words approach}
    \end{subfigure}
    \begin{subfigure}{.32\textwidth}
    \centering
    \includegraphics[width=.95\linewidth]{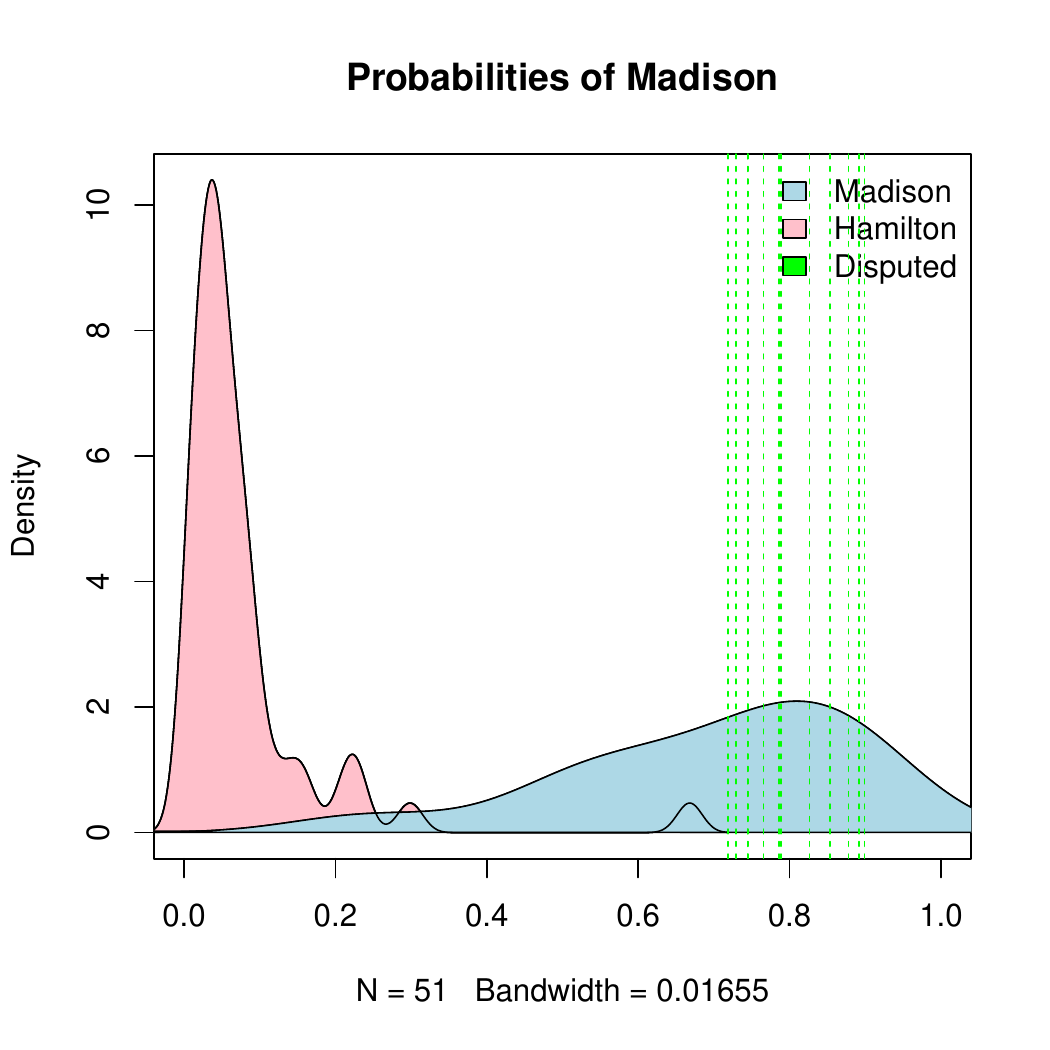}
    \caption{Doc-Topic from LDA}
    \end{subfigure}
    \begin{subfigure}{.32\textwidth}
    \centering
    \includegraphics[width=.95\linewidth]{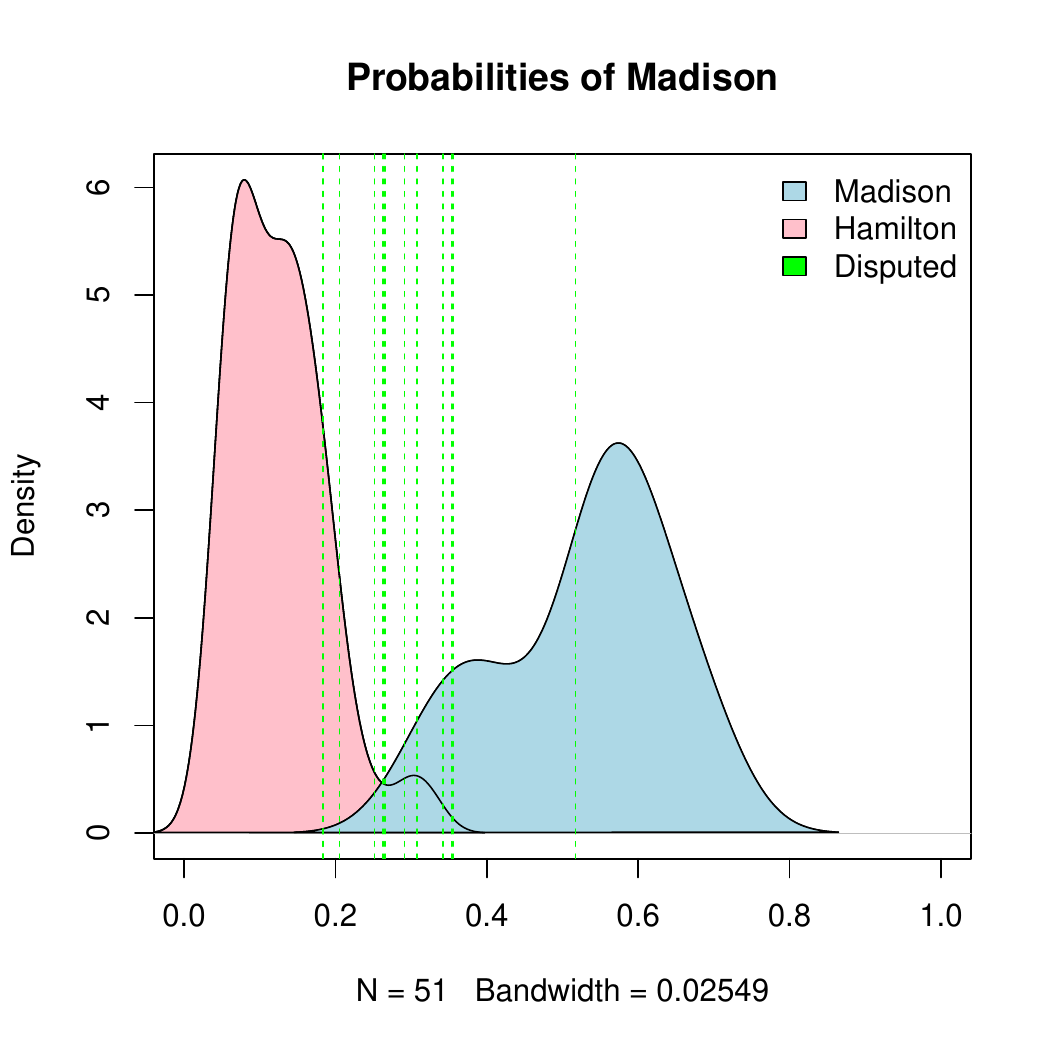}
    \caption{Llama2}
    \end{subfigure}
    \caption{The estimated densities from BART. The red density is the kernel density estimate of predicted probabilities of BART for papers authored by Hamilton, and the blue density is the kernel density estimate of the ones by Madison. The predicted probabilities of disputed papers are denoted as green vertical lines. For BoW and LDA, the results are based on Type 2 input.  For continuous embeddings (c), the predicted probabilities for the disputed papers are not extreme in their range compared with the one using BoW embeddings (b), and tend to be concentrated in the overlap of two probability distributions. }
    \label{fig:predicted-prob-density}
\end{figure}

A key distinction emerges between BoW and continuous embeddings in terms of the estimated probability distributions. As illustrated in Figure~\ref{fig:predicted-prob-density}, BoW-based embeddings yield a more polarized probability distribution, effectively separating Hamilton and Madison's papers. In contrast, continuous embeddings tend to concentrate the predicted probabilities within the overlap region, leading to greater uncertainty in authorship attribution.

Despite their success in various NLP tasks, Large Language Models (LLMs) struggle with authorship attribution in this context. Several factors contribute to this underperformance. First, LLMs primarily capture semantic meaning rather than stylistic patterns. Authorship attribution relies heavily on function words and subtle stylistic markers, which LLMs deprioritize in favor of content-related word relationships. This is particularly evident in their inability to distinguish Madison's function-word usage from Hamilton's, a key differentiator identified in previous stylometric studies \citep{Mosteller1963Inference}.

Second, continuous embeddings generated by LLMs tend to distribute documents in a high-dimensional latent space, where stylistic differences become less pronounced compared to discrete frequency-based representations. In contrast, BoW-based models maintain sharp stylistic distinctions by preserving explicit word frequency distributions.

Finally, LLMs are trained on extensive and diverse corpora that introduce broad linguistic generalization, often diluting author-specific writing patterns. Even after fine-tuning the architecture LLMs on a Federalist Papers-specific dataset, it may not necessarily improve the performance (see Section~\ref{sec:fine-tuning} in the supplement). Table~\ref{tab:l2_loss_for_all} and an additional table in the supplement (Table~\ref{tab:predicted_prob_bart})) further illustrate these findings, showing that LLMs yield higher $\ell_2$ loss and more ambiguous predicted probabilities compared to simpler models.

\subsection{Evidence for Joint Authorship}\label{sec:discussion}

In previous sections, we mainly focused on the binary classification task as much other literature does, both for simplicity and interpretability. However, there has been a longstanding debate regarding the feasibility of joint authorship instead of individual authorship \citep{authorship1944adair, Mosteller1963Inference}. Although not admitted by \cite{authorship1944adair}, No. 18, 19, 20 are mostly widely recognized jointly authored papers by Hamilton and Madison. In \cite{Mosteller1963Inference}, No. 18 and No.19 are shown to be mainly authored by Madison, although being jointly authored. The amount of contribution by Hamilton and Madison for No.20 remains unclear. \cite{authorship1944adair} mentioned that according to Madison's note on No.20, he borrowed much from Sir William Temple or Felice, and thus less support toward Madison might not necessarily mean more contribution by Hamilton for this exact case. 

In the Bag-of-Words approach with different types of input (Table~\ref{tab:predicted-prob-joint-tdms}), the support for Madison in Federalist No. 20 is relatively weak, particularly with Type 3 input, where all methods show the weakest support. The predicted probabilities in Table~\ref{tab:predicted-prob-joint-papers} are rather inconsistent as the papers with strongest evidence and weakest evidence toward Madison vary a lot by method. However, for continuous embeddings, as we have observed earlier in Figure~\ref{fig:predicted-prob-density}, the range of the support is small compared with the BoW type of methods. 

\begin{table}[]
\resizebox{\textwidth}{!}{%
\begin{tabular}{@{}cccccccccccc@{}}
\toprule
 &
  \multicolumn{4}{c}{BoW Embeddings} &
  \multicolumn{7}{c}{Continuous Embeddings} \\ \midrule
 &
  BoW &
  LDA &
  SVD &
  NMF &
  Word2Vec &
  BERT &
  RoBERTa &
  BART &
  GPT4 &
  Llama2 &
  Llama3 \\ \midrule
No.18 &
  0.6375 &
  \textbf{0.2489} &
  \textbf{0.6272} &
  \textbf{0.6360} &
  0.3565 &
  \textbf{0.4844} &
  \textbf{0.3856} &
  {\ul \textit{0.1702}} &
  0.2762 &
  0.2129 &
  \textbf{0.3325} \\
No.19 &
  \textbf{0.7328} &
  {\ul \textit{0.1156}} &
  {\ul \textit{0.5497}} &
  0.5704 &
  \textbf{0.4185} &
  {\ul \textit{0.3928}} &
  {\ul \textit{0.1336}} &
  \textbf{0.4846} &
  {\ul \textit{0.2377}} &
  {\ul \textit{0.1898}} &
  0.2777 \\
No.20 &
  {\ul \textit{0.5228}} &
  0.1549 &
  0.5544 &
  {\ul \textit{0.4602}} &
  {\ul \textit{0.3334}} &
  0.4748 &
  0.1958 &
  0.2121 &
  \textbf{0.2903} &
  \textbf{0.2473} &
  {\ul \textit{0.2535}} \\ \bottomrule
\end{tabular}%
}
\caption{Predicted probabilities from BART classifier for jointly authored papers (No. 18, 19, 20). For BoW, the results based on Type 2 inputs are reported. The highest support toward Madison authorship is bolded and the lowest support is underlined. }
\label{tab:predicted-prob-joint-papers}
\end{table}

\section{Suggestions for Practitioners: Insights from LLMs and Case Studies}\label{sec:advice}
After reviewing the performance of Large Language Models (LLMs) without fine-tuning and applying them to a case study like the Federalist Papers, several key suggestions can be drawn for practitioners:
\begin{enumerate}
    \item \textbf{Understand the Limits of General-Purpose Embeddings}: Large Langage Model is not the panacea. It requires delicate fine-tuning for good performance \citep{fabien2020bertaa}, and vanilla embedding's performance is very much task-dependent. Practitioners should be aware that while LLMs like GPT, BERT, and RoBERTa may not always perform optimally for specialized tasks like authorship attribution without fine-tuning, although they are capable of producing high-quality general-purpose embeddings. In cases like the Federalist Papers, where subtle stylistic or linguistic features are critical, relying solely on general embeddings may overlook important nuances.

    \item  \textbf{Choose the Right LLM for Embedding}: It is crucial to carefully consider the choice of Large Language Models (LLMs) based on the specific task at hand. If the goal is to obtain general-purpose embeddings that capture rich contextual information, encoder models like BERT or RoBERTa are generally more effective. These models are specifically designed for tasks that require deep understanding of input text, as they process the entire sequence bidirectionally and generate embeddings that reflect both the preceding and following context. In contrast, autoregressive models like GPT or LlaMA are better suited for tasks that involve text generation rather than producing general-purpose embeddings.

    \item \textbf{Task-Specific Fine-Tuning for LLMs}: General-purpose LLMs, though powerful, may miss task-specific patterns unless fine-tuned. Fine-tune LLMs for highly specialized tasks like authorship attribution, and explore traditional statistical models for complementary insights \citep{fabien2020bertaa}. See the experiment results in Appendix~\ref{sec:fine-tuning}.

    \item \textbf{Feed the Quality Input}: In our experiments, well-curated set of words plays a key role in boosting classification performances. The quality of input text plays a crucial role in obtaining meaningful embeddings from any model. This cannot be overly emphasized for Large Language Models, because they do not rely on modeling assumptions but learn from the data. 
    
    \item \textbf{Leverage Traditional Methods for Fine-Grained Analysis}: In tasks like authorship attribution, where differences in writing style may be subtle and involve fine-grained linguistic features, traditional statistical models often outperform general-purpose LLMs. Models like LDA, which assume probabilistic structures, can capture topic-based distinctions that LLM embeddings may miss without fine-tuning. Practitioners should not discard traditional methods like LDA or PCA. Combining these methods with LLM embeddings can lead to a more comprehensive analysis.

    \item \textbf{Consider Dimensionality Reduction Techniques}: Increasing the dimensionality of embeddings through LLMs does not always lead to better performance for tasks like authorship attribution. In some cases, reducing dimensionality (e.g., with PCA) can highlight stylistic features more effectively by removing noise and focusing on the core attributes relevant to the task. 
\end{enumerate}

We hope these observations are insightful and that the practitioners can strike a balance between leveraging the power of LLMs and using traditional techniques, ensuring they get the most out of both approaches for tasks like authorship attribution and other specialized text analyses.

\section{Conclusion}\label{sec:conclusion}

Our study demonstrates that embeddings from larger models or neural network-based models do not necessarily improve authorship attribution results. Instead, the predicted probabilities generated by BART \citep{Chipman2010bart}, using LDA embeddings, provide the best classification performance, yielding the lowest $\ell_2$ loss and classification error across different thresholds. We also provide guidelines on how to combine Large Language Models (LLMs) with statistical models in practical applications.

In contrast to many language classification tasks where semantic encodings play a critical role, the primary discriminators in authorship attribution are \textbf{function words}. For the Federalist Papers, words such as `upon' vs. 'on' and `while' vs. `whilst' were found to be significant markers for distinguishing one author from another. This set of words was initially investigated by Mosteller and Wallace (1963) and has been frequently used in successful authorship studies \citep{Tweedie1996Neural, Robert1998Seperating}. However, in contextual embedding models, these function words are encoded close to each other in the embedding space, as measured by cosine similarity. For example, in the Word2Vec model, the cosine similarity between `upon' and `on' is 0.51, and between `while' and `whilst' is 0.58, close to the similarity between `king' and `queen' (0.65). This proximity reduces the discriminative power of these function words in contextual models.

The scaling laws proposed by \cite{kaplan2020scaling} show that larger neural language models improve predictably as model size, dataset size, and compute power increase. These laws have driven the development of models like GPT-3 and GPT-4, with significant investments in scaling up model parameters and datasets to improve performance on NLP tasks. However, statistical language models (SLMs) can still outperform neural LLMs in specific scenarios. When data is scarce or domain-specific, SLMs may perform better, especially when neural models are prone to overfitting due to insufficient fine-tuning data. Additionally, statistical models offer greater transparency and interpretability, which is critical in areas like legal or healthcare applications where explainability is crucial. While scaling laws favor the development of larger neural models, SLMs remain advantageous in cases requiring efficiency, interpretability, or when dealing with small, domain-specific datasets.


We hope this study improves the understanding of modern language model in statistical perspective and sheds light on the statistical language modeling still can have an edge on analysis with more interpretability with theoretical grounds. 

\section*{Acknowledgement}
This paper includes text and revisions generated with the assistance of generative AI tools (e.g., ChatGPT). Final content was reviewed and edited by the authors.

\newpage
\bibliographystyle{chicago}
\bibliography{main.bib}

\begin{thebibliography}{}

\bibitem[\protect\citeauthoryear{Adair}{Adair}{1944}]{authorship1944adair}
Adair, D. (1944).
\newblock The authorship of the disputed federalist papers.
\newblock {\em The William and Mary Quarterly\/}~{\em 1\/}(2), 98--122.

\bibitem[\protect\citeauthoryear{Arnold and Press}{Arnold and
  Press}{1989}]{arnold1989compatible}
Arnold, B.~C. and S.~J. Press (1989).
\newblock Compatible conditional distributions.
\newblock {\em Journal of the American Statistical Association\/}~{\em
  84\/}(405), 152--156.

\bibitem[\protect\citeauthoryear{Belkin, Hsu, Ma, and Mandal}{Belkin
  et~al.}{2019}]{belkin2019reconciling}
Belkin, M., D.~Hsu, S.~Ma, and S.~Mandal (2019, July).
\newblock Reconciling modern machine-learning practice and the classical
  bias–variance trade-off.
\newblock {\em Proceedings of the National Academy of Sciences\/}~{\em
  116\/}(32), 15849–15854.

\bibitem[\protect\citeauthoryear{Bengio, Ducharme, Vincent, and Janvin}{Bengio
  et~al.}{2003}]{bengio2003neural}
Bengio, Y., R.~Ducharme, P.~Vincent, and C.~Janvin (2003, mar).
\newblock A neural probabilistic language model.
\newblock {\em J. Mach. Learn. Res.\/}~{\em 3\/}(null), 1137–1155.

\bibitem[\protect\citeauthoryear{Besag}{Besag}{1974}]{besag1974spatial}
Besag, J. (1974).
\newblock Spatial interaction and the statistical analysis of lattice systems.
\newblock {\em Journal of the Royal Statistical Society: Series B
  (Methodological)\/}~{\em 36\/}(2), 192--225.

\bibitem[\protect\citeauthoryear{Blei, Ng, and Jordan}{Blei
  et~al.}{2003}]{blei2003latent}
Blei, D.~M., A.~Y. Ng, and M.~I. Jordan (2003, mar).
\newblock Latent dirichlet allocation.
\newblock {\em J. Mach. Learn. Res.\/}~{\em 3\/}(null), 993–1022.

\bibitem[\protect\citeauthoryear{Bommasani, Hudson, Adeli, Altman, Arora, von
  Arx, Bernstein, Bohg, Bosselut, Brunskill, Brynjolfsson, Buch, Card,
  Castellon, Chatterji, Chen, Creel, Davis, Demszky, Donahue, Doumbouya,
  Durmus, Ermon, Etchemendy, Ethayarajh, Fei-Fei, Finn, Gale, Gillespie, Goel,
  Goodman, Grossman, Guha, Hashimoto, Henderson, Hewitt, Ho, Hong, Hsu, Huang,
  Icard, Jain, Jurafsky, Kalluri, Karamcheti, Keeling, Khani, Khattab, Koh,
  Krass, Krishna, Kuditipudi, Kumar, Ladhak, Lee, Lee, Leskovec, Levent, Li,
  Li, Ma, Malik, Manning, Mirchandani, Mitchell, Munyikwa, Nair, Narayan,
  Narayanan, Newman, Nie, Niebles, Nilforoshan, Nyarko, Ogut, Orr,
  Papadimitriou, Park, Piech, Portelance, Potts, Raghunathan, Reich, Ren, Rong,
  Roohani, Ruiz, Ryan, Ré, Sadigh, Sagawa, Santhanam, Shih, Srinivasan,
  Tamkin, Taori, Thomas, Tramèr, Wang, Wang, Wu, Wu, Wu, Xie, Yasunaga, You,
  Zaharia, Zhang, Zhang, Zhang, Zhang, Zheng, Zhou, and Liang}{Bommasani
  et~al.}{2022}]{bommasani2022opportunities}
Bommasani, R., D.~A. Hudson, E.~Adeli, R.~Altman, S.~Arora, S.~von Arx, M.~S.
  Bernstein, J.~Bohg, A.~Bosselut, E.~Brunskill, E.~Brynjolfsson, S.~Buch,
  D.~Card, R.~Castellon, N.~Chatterji, A.~Chen, K.~Creel, J.~Q. Davis,
  D.~Demszky, C.~Donahue, M.~Doumbouya, E.~Durmus, S.~Ermon, J.~Etchemendy,
  K.~Ethayarajh, L.~Fei-Fei, C.~Finn, T.~Gale, L.~Gillespie, K.~Goel,
  N.~Goodman, S.~Grossman, N.~Guha, T.~Hashimoto, P.~Henderson, J.~Hewitt,
  D.~E. Ho, J.~Hong, K.~Hsu, J.~Huang, T.~Icard, S.~Jain, D.~Jurafsky,
  P.~Kalluri, S.~Karamcheti, G.~Keeling, F.~Khani, O.~Khattab, P.~W. Koh,
  M.~Krass, R.~Krishna, R.~Kuditipudi, A.~Kumar, F.~Ladhak, M.~Lee, T.~Lee,
  J.~Leskovec, I.~Levent, X.~L. Li, X.~Li, T.~Ma, A.~Malik, C.~D. Manning,
  S.~Mirchandani, E.~Mitchell, Z.~Munyikwa, S.~Nair, A.~Narayan, D.~Narayanan,
  B.~Newman, A.~Nie, J.~C. Niebles, H.~Nilforoshan, J.~Nyarko, G.~Ogut, L.~Orr,
  I.~Papadimitriou, J.~S. Park, C.~Piech, E.~Portelance, C.~Potts,
  A.~Raghunathan, R.~Reich, H.~Ren, F.~Rong, Y.~Roohani, C.~Ruiz, J.~Ryan,
  C.~Ré, D.~Sadigh, S.~Sagawa, K.~Santhanam, A.~Shih, K.~Srinivasan,
  A.~Tamkin, R.~Taori, A.~W. Thomas, F.~Tramèr, R.~E. Wang, W.~Wang, B.~Wu,
  J.~Wu, Y.~Wu, S.~M. Xie, M.~Yasunaga, J.~You, M.~Zaharia, M.~Zhang, T.~Zhang,
  X.~Zhang, Y.~Zhang, L.~Zheng, K.~Zhou, and P.~Liang (2022).
\newblock On the opportunities and risks of foundation models.

\bibitem[\protect\citeauthoryear{Bosch and Smith}{Bosch and
  Smith}{1998}]{Robert1998Seperating}
Bosch, R.~A. and J.~A. Smith (1998).
\newblock Separating hyperplanes and the authorship of the disputed federalist
  papers.
\newblock {\em The American Mathematical Monthly\/}~{\em 105\/}(7), 601--608.

\bibitem[\protect\citeauthoryear{Breiman}{Breiman}{1996}]{breiman1996bagging}
Breiman, L. (1996).
\newblock Bagging predictors.
\newblock {\em Machine learning\/}~{\em 24}, 123--140.

\bibitem[\protect\citeauthoryear{Brown, Mann, Ryder, Subbiah, Kaplan, Dhariwal,
  Neelakantan, Shyam, Sastry, Askell, Agarwal, Herbert-Voss, Krueger, Henighan,
  Child, Ramesh, Ziegler, Wu, Winter, Hesse, Chen, Sigler, Litwin, Gray, Chess,
  Clark, Berner, McCandlish, Radford, Sutskever, and Amodei}{Brown
  et~al.}{2020}]{brown2020gpt3}
Brown, T.~B., B.~Mann, N.~Ryder, M.~Subbiah, J.~Kaplan, P.~Dhariwal,
  A.~Neelakantan, P.~Shyam, G.~Sastry, A.~Askell, S.~Agarwal, A.~Herbert-Voss,
  G.~Krueger, T.~Henighan, R.~Child, A.~Ramesh, D.~M. Ziegler, J.~Wu,
  C.~Winter, C.~Hesse, M.~Chen, E.~Sigler, M.~Litwin, S.~Gray, B.~Chess,
  J.~Clark, C.~Berner, S.~McCandlish, A.~Radford, I.~Sutskever, and D.~Amodei
  (2020).
\newblock Language models are few-shot learners.

\bibitem[\protect\citeauthoryear{Chipman, George, and McCulloch}{Chipman
  et~al.}{2010}]{Chipman2010bart}
Chipman, H.~A., E.~I. George, and R.~E. McCulloch (2010, March).
\newblock Bart: Bayesian additive regression trees.
\newblock {\em The Annals of Applied Statistics\/}~{\em 4\/}(1).

\bibitem[\protect\citeauthoryear{Collins, Kaufer, Vlachos, Butler, and
  Ishizaki}{Collins et~al.}{2004}]{Collins2004Detecting}
Collins, J., D.~Kaufer, P.~Vlachos, B.~Butler, and S.~Ishizaki (2004).
\newblock Detecting collaborations in text: Comparing the authors' rhetorical
  language choices in the federalist papers.
\newblock {\em Computers and the Humanities\/}~{\em 38\/}(1), 15--36.

\bibitem[\protect\citeauthoryear{Cybenko}{Cybenko}{1989}]{cybenko1989approximation}
Cybenko, G. (1989).
\newblock Approximation by superpositions of a sigmoidal function.
\newblock {\em Mathematics of control, signals and systems\/}~{\em 2\/}(4),
  303--314.

\bibitem[\protect\citeauthoryear{Deerwester, Dumais, Furnas, Landauer, and
  Harshman}{Deerwester et~al.}{1990}]{deerwester1990indexing}
Deerwester, S., S.~T. Dumais, G.~W. Furnas, T.~K. Landauer, and R.~Harshman
  (1990).
\newblock Indexing by latent semantic analysis.
\newblock {\em Journal of the American society for information science\/}~{\em
  41\/}(6), 391--407.

\bibitem[\protect\citeauthoryear{Devlin, Chang, Lee, and Toutanova}{Devlin
  et~al.}{2019}]{devlin2019bert}
Devlin, J., M.-W. Chang, K.~Lee, and K.~Toutanova (2019).
\newblock Bert: Pre-training of deep bidirectional transformers for language
  understanding.

\bibitem[\protect\citeauthoryear{Diederich, Kindermann, Leopold, and
  Paass}{Diederich et~al.}{2003}]{Diederich2003SVM}
Diederich, J., J.~Kindermann, E.~Leopold, and G.~Paass (2003, may).
\newblock Authorship attribution with support vector machines.
\newblock {\em Applied Intelligence\/}~{\em 19\/}(1–2), 109–123.

\bibitem[\protect\citeauthoryear{Ding, Li, and Peng}{Ding
  et~al.}{2006}]{ding2006nonnegative}
Ding, C., T.~Li, and W.~Peng (2006).
\newblock Nonnegative matrix factorization and probabilistic latent semantic
  indexing: Equivalence chi-square statistic, and a hybrid method.
\newblock In {\em AAAI}, Volume~42, pp.\  137--43.

\bibitem[\protect\citeauthoryear{Donoho and Jin}{Donoho and
  Jin}{2004}]{Donoho2004Higher}
Donoho, D. and J.~Jin (2004).
\newblock {Higher criticism for detecting sparse heterogeneous mixtures}.
\newblock {\em The Annals of Statistics\/}~{\em 32\/}(3), 962 -- 994.

\bibitem[\protect\citeauthoryear{Fabien, Villatoro-Tello, Motlicek, and
  Parida}{Fabien et~al.}{2020}]{fabien2020bertaa}
Fabien, M., E.~Villatoro-Tello, P.~Motlicek, and S.~Parida (2020, December).
\newblock {B}ert{AA} : {BERT} fine-tuning for authorship attribution.
\newblock In P.~Bhattacharyya, D.~M. Sharma, and R.~Sangal (Eds.), {\em
  Proceedings of the 17th International Conference on Natural Language
  Processing (ICON)}, Indian Institute of Technology Patna, Patna, India, pp.\
  127--137. NLP Association of India (NLPAI).

\bibitem[\protect\citeauthoryear{Feng, Tung, Ahmed, Bengio, and
  Hajimirsadegh}{Feng et~al.}{2024}]{feng2024rnnsneeded}
Feng, L., F.~Tung, M.~O. Ahmed, Y.~Bengio, and H.~Hajimirsadegh (2024).
\newblock Were rnns all we needed?

\bibitem[\protect\citeauthoryear{Freund and Schapire}{Freund and
  Schapire}{1997}]{freund1997decision}
Freund, Y. and R.~E. Schapire (1997).
\newblock A decision-theoretic generalization of on-line learning and an
  application to boosting.
\newblock {\em Journal of computer and system sciences\/}~{\em 55\/}(1),
  119--139.

\bibitem[\protect\citeauthoryear{Goodfellow, Pouget-Abadie, Mirza, Xu,
  Warde-Farley, Ozair, Courville, and Bengio}{Goodfellow
  et~al.}{2014}]{goodfellow2014gan}
Goodfellow, I.~J., J.~Pouget-Abadie, M.~Mirza, B.~Xu, D.~Warde-Farley,
  S.~Ozair, A.~Courville, and Y.~Bengio (2014).
\newblock Generative adversarial networks.

\bibitem[\protect\citeauthoryear{Gu and Dao}{Gu and
  Dao}{2024}]{gu2024mambalineartimesequencemodeling}
Gu, A. and T.~Dao (2024).
\newblock Mamba: Linear-time sequence modeling with selective state spaces.

\bibitem[\protect\citeauthoryear{Gutmann and Hyv{\"a}rinen}{Gutmann and
  Hyv{\"a}rinen}{2012}]{gutmann2012noise}
Gutmann, M.~U. and A.~Hyv{\"a}rinen (2012).
\newblock Noise-contrastive estimation of unnormalized statistical models, with
  applications to natural image statistics.
\newblock {\em Journal of machine learning research\/}~{\em 13\/}(2).

\bibitem[\protect\citeauthoryear{Hochreiter}{Hochreiter}{1997}]{hochreiter1997long}
Hochreiter, S. (1997).
\newblock Long short-term memory.
\newblock {\em Neural Computation MIT-Press\/}.

\bibitem[\protect\citeauthoryear{Hoffmann, Borgeaud, Mensch, Buchatskaya, Cai,
  Rutherford, de~Las~Casas, Hendricks, Welbl, Clark, Hennigan, Noland,
  Millican, van~den Driessche, Damoc, Guy, Osindero, Simonyan, Elsen, Rae,
  Vinyals, and Sifre}{Hoffmann et~al.}{2022}]{hoffmann2022chinchilla}
Hoffmann, J., S.~Borgeaud, A.~Mensch, E.~Buchatskaya, T.~Cai, E.~Rutherford,
  D.~de~Las~Casas, L.~A. Hendricks, J.~Welbl, A.~Clark, T.~Hennigan, E.~Noland,
  K.~Millican, G.~van~den Driessche, B.~Damoc, A.~Guy, S.~Osindero,
  K.~Simonyan, E.~Elsen, J.~W. Rae, O.~Vinyals, and L.~Sifre (2022).
\newblock Training compute-optimal large language models.

\bibitem[\protect\citeauthoryear{Hofmann}{Hofmann}{1999}]{hofmann1999learning}
Hofmann, T. (1999).
\newblock Learning the similarity of documents: An information-geometric
  approach to document retrieval and categorization.
\newblock {\em Advances in neural information processing systems\/}~{\em 12}.

\bibitem[\protect\citeauthoryear{Holmes and Forsyth}{Holmes and
  Forsyth}{1995}]{holmes1995federalist}
Holmes, D.~I. and R.~S. Forsyth (1995).
\newblock The federalist revisited: New directions in authorship attribution.
\newblock {\em Literary and Linguistic computing\/}~{\em 10\/}(2), 111--127.

\bibitem[\protect\citeauthoryear{Howard and Ruder}{Howard and
  Ruder}{2018}]{howard2018universal}
Howard, J. and S.~Ruder (2018).
\newblock Universal language model fine-tuning for text classification.

\bibitem[\protect\citeauthoryear{Kaplan, McCandlish, Henighan, Brown, Chess,
  Child, Gray, Radford, Wu, and Amodei}{Kaplan
  et~al.}{2020}]{kaplan2020scaling}
Kaplan, J., S.~McCandlish, T.~Henighan, T.~B. Brown, B.~Chess, R.~Child,
  S.~Gray, A.~Radford, J.~Wu, and D.~Amodei (2020).
\newblock Scaling laws for neural language models.

\bibitem[\protect\citeauthoryear{Kim, O'Hagan, and Rockova}{Kim
  et~al.}{2024}]{kim2024adaptive}
Kim, J., S.~O'Hagan, and V.~Rockova (2024).
\newblock Adaptive uncertainty quantification for generative ai.

\bibitem[\protect\citeauthoryear{Kipnis}{Kipnis}{2022}]{Kipnis2022Higher}
Kipnis, A. (2022).
\newblock {Higher criticism for discriminating word-frequency tables and
  authorship attribution}.
\newblock {\em The Annals of Applied Statistics\/}~{\em 16\/}(2), 1236 -- 1252.

\bibitem[\protect\citeauthoryear{Lee and Seung}{Lee and
  Seung}{1999}]{lee1999learning}
Lee, D.~D. and H.~S. Seung (1999).
\newblock Learning the parts of objects by non-negative matrix factorization.
\newblock {\em nature\/}~{\em 401\/}(6755), 788--791.

\bibitem[\protect\citeauthoryear{Lewis, Liu, Goyal, Ghazvininejad, Mohamed,
  Levy, Stoyanov, and Zettlemoyer}{Lewis et~al.}{2019}]{lewis2019bart}
Lewis, M., Y.~Liu, N.~Goyal, M.~Ghazvininejad, A.~Mohamed, O.~Levy,
  V.~Stoyanov, and L.~Zettlemoyer (2019).
\newblock Bart: Denoising sequence-to-sequence pre-training for natural
  language generation, translation, and comprehension.

\bibitem[\protect\citeauthoryear{Liu, Ott, Goyal, Du, Joshi, Chen, Levy, Lewis,
  Zettlemoyer, and Stoyanov}{Liu
  et~al.}{2019}]{liu2019robertarobustlyoptimizedbert}
Liu, Y., M.~Ott, N.~Goyal, J.~Du, M.~Joshi, D.~Chen, O.~Levy, M.~Lewis,
  L.~Zettlemoyer, and V.~Stoyanov (2019).
\newblock Roberta: A robustly optimized bert pretraining approach.

\bibitem[\protect\citeauthoryear{Lu, Li, Cai, Yi, Liu, Zhang, Lane, and Xu}{Lu
  et~al.}{2024}]{lu2024smalllanguagemodelssurvey}
Lu, Z., X.~Li, D.~Cai, R.~Yi, F.~Liu, X.~Zhang, N.~D. Lane, and M.~Xu (2024).
\newblock Small language models: Survey, measurements, and insights.

\bibitem[\protect\citeauthoryear{Meta}{Meta}{2024}]{meta2024llama3}
Meta (2024).
\newblock Introducing llama 3.
\newblock \url{https://ai.meta.com/llama/}.

\bibitem[\protect\citeauthoryear{Mikolov}{Mikolov}{2013}]{mikolov2013efficient}
Mikolov, T. (2013).
\newblock Efficient estimation of word representations in vector space.
\newblock {\em arXiv preprint arXiv:1301.3781\/}.

\bibitem[\protect\citeauthoryear{Mikolov, Sutskever, Chen, Corrado, and
  Dean}{Mikolov et~al.}{2013}]{mikolov2013distributed}
Mikolov, T., I.~Sutskever, K.~Chen, G.~Corrado, and J.~Dean (2013).
\newblock Distributed representations of words and phrases and their
  compositionality.

\bibitem[\protect\citeauthoryear{Mnih and Teh}{Mnih and
  Teh}{2012}]{mnih2012fast}
Mnih, A. and Y.~W. Teh (2012).
\newblock A fast and simple algorithm for training neural probabilistic
  language models.
\newblock {\em arXiv preprint arXiv:1206.6426\/}.

\bibitem[\protect\citeauthoryear{Mosteller}{Mosteller}{1987}]{mosteller1987statistical}
Mosteller, F. (1987).
\newblock A statistical study of the writing styles of the authors of "the
  federalist" papers.
\newblock {\em Proceedings of the American Philosophical Society\/}~{\em
  131\/}(2), 132--140.

\bibitem[\protect\citeauthoryear{Mosteller and Wallace}{Mosteller and
  Wallace}{1963}]{Mosteller1963Inference}
Mosteller, F. and D.~L. Wallace (1963).
\newblock Inference in an authorship problem.
\newblock {\em Journal of the American Statistical Association\/}~{\em
  58\/}(302), 275--309.

\bibitem[\protect\citeauthoryear{Nadaraya}{Nadaraya}{1964}]{nadaraya1964estimating}
Nadaraya, E.~A. (1964).
\newblock On estimating regression.
\newblock {\em Theory of Probability \& Its Applications\/}~{\em 9\/}(1),
  141--142.

\bibitem[\protect\citeauthoryear{Nakkiran, Kaplun, Bansal, Yang, Barak, and
  Sutskever}{Nakkiran et~al.}{2019}]{nakkiran2019deepdoubledescentbigger}
Nakkiran, P., G.~Kaplun, Y.~Bansal, T.~Yang, B.~Barak, and I.~Sutskever (2019).
\newblock Deep double descent: Where bigger models and more data hurt.

\bibitem[\protect\citeauthoryear{OpenAI}{OpenAI}{2023}]{openai2023gpt4}
OpenAI (2023).
\newblock Gpt-4 technical report.
\newblock \url{https://openai.com/research/gpt-4}.

\bibitem[\protect\citeauthoryear{Park and Casella}{Park and
  Casella}{2008}]{park2008bayesian}
Park, T. and G.~Casella (2008).
\newblock The bayesian lasso.
\newblock {\em Journal of the american statistical association\/}~{\em
  103\/}(482), 681--686.

\bibitem[\protect\citeauthoryear{Pennington, Socher, and Manning}{Pennington
  et~al.}{2014}]{pennington2014glove}
Pennington, J., R.~Socher, and C.~D. Manning (2014).
\newblock Glove: Global vectors for word representation.
\newblock In {\em Empirical Methods in Natural Language Processing (EMNLP)},
  pp.\  1532--1543.

\bibitem[\protect\citeauthoryear{Popescu and Dinu}{Popescu and
  Dinu}{2007}]{popescu2007kernel}
Popescu, M. and L.~P. Dinu (2007).
\newblock Kernel methods and string kernels for authorship identification: The
  federalist papers case.
\newblock In {\em Recent Advances in Natural Language Processing}, pp.\  484.

\bibitem[\protect\citeauthoryear{Power, Burda, Edwards, Babuschkin, and
  Misra}{Power et~al.}{2022}]{power2022grokking}
Power, A., Y.~Burda, H.~Edwards, I.~Babuschkin, and V.~Misra (2022).
\newblock Grokking: Generalization beyond overfitting on small algorithmic
  datasets.

\bibitem[\protect\citeauthoryear{Rabiner}{Rabiner}{1989}]{rabiner1989tutorial}
Rabiner, L.~R. (1989).
\newblock A tutorial on hidden markov models and selected applications in
  speech recognition.
\newblock {\em Proceedings of the IEEE\/}~{\em 77\/}(2), 257--286.

\bibitem[\protect\citeauthoryear{Radford}{Radford}{2018}]{radford2018improvingLU}
Radford, A. (2018).
\newblock Improving language understanding by generative pre-training.

\bibitem[\protect\citeauthoryear{Radford, Wu, Child, Luan, Amodei, Sutskever,
  et~al.}{Radford et~al.}{2019}]{radford2019gpt2}
Radford, A., J.~Wu, R.~Child, D.~Luan, D.~Amodei, I.~Sutskever, et~al. (2019).
\newblock Language models are unsupervised multitask learners.
\newblock {\em OpenAI blog\/}~{\em 1\/}(8), 9.

\bibitem[\protect\citeauthoryear{Reimers and Gurevych}{Reimers and
  Gurevych}{2019}]{reimers2019sentencebert}
Reimers, N. and I.~Gurevych (2019).
\newblock Sentence-bert: Sentence embeddings using siamese bert-networks.

\bibitem[\protect\citeauthoryear{Ro{\v{c}}kov{\'a} and
  George}{Ro{\v{c}}kov{\'a} and George}{2018}]{rovckova2018spike}
Ro{\v{c}}kov{\'a}, V. and E.~I. George (2018).
\newblock The spike-and-slab lasso.
\newblock {\em Journal of the American Statistical Association\/}~{\em
  113\/}(521), 431--444.

\bibitem[\protect\citeauthoryear{Rumelhart, Hinton, and Williams}{Rumelhart
  et~al.}{1986}]{rumelhart1986learning}
Rumelhart, D.~E., G.~E. Hinton, and R.~J. Williams (1986).
\newblock Learning internal representations by error propagation, parallel
  distributed processing, explorations in the microstructure of cognition, ed.
  de rumelhart and j. mcclelland. vol. 1. 1986.
\newblock {\em Biometrika\/}~{\em 71\/}(599-607), 6.

\bibitem[\protect\citeauthoryear{Sari, Stevenson, and Vlachos}{Sari
  et~al.}{2018}]{sari2018topic}
Sari, Y., M.~Stevenson, and A.~Vlachos (2018, August).
\newblock Topic or style? exploring the most useful features for authorship
  attribution.
\newblock In E.~M. Bender, L.~Derczynski, and P.~Isabelle (Eds.), {\em
  Proceedings of the 27th International Conference on Computational
  Linguistics}, Santa Fe, New Mexico, USA, pp.\  343--353. Association for
  Computational Linguistics.

\bibitem[\protect\citeauthoryear{Seroussi, Zukerman, and Bohnert}{Seroussi
  et~al.}{2014}]{Seroussi2014Authorship}
Seroussi, Y., I.~Zukerman, and F.~Bohnert (2014, 06).
\newblock {Authorship Attribution with Topic Models}.
\newblock {\em Computational Linguistics\/}~{\em 40\/}(2), 269--310.

\bibitem[\protect\citeauthoryear{Tibshirani}{Tibshirani}{1996}]{tibshirani1996lasso}
Tibshirani, R. (1996).
\newblock Regression shrinkage and selection via the lasso.
\newblock {\em Journal of the Royal Statistical Society. Series B
  (Methodological)\/}~{\em 58\/}(1), 267--288.

\bibitem[\protect\citeauthoryear{Touvron, Lavril, Izacard, Martinet, Lachaux,
  Lacroix, Rozière, Goyal, Hambro, Azhar, Rodriguez, Joulin, Grave, and
  Lample}{Touvron et~al.}{2023}]{touvron2023llama}
Touvron, H., T.~Lavril, G.~Izacard, X.~Martinet, M.-A. Lachaux, T.~Lacroix,
  B.~Rozière, N.~Goyal, E.~Hambro, F.~Azhar, A.~Rodriguez, A.~Joulin,
  E.~Grave, and G.~Lample (2023).
\newblock Llama: Open and efficient foundation language models.

\bibitem[\protect\citeauthoryear{Tweedie, Singh, and Holmes}{Tweedie
  et~al.}{1996}]{Tweedie1996Neural}
Tweedie, F.~J., S.~Singh, and D.~I. Holmes (1996).
\newblock Neural network applications in stylometry: The "federalist papers".
\newblock {\em Computers and the Humanities\/}~{\em 30\/}(1), 1--10.

\bibitem[\protect\citeauthoryear{Vaswani, Shazeer, Parmar, Uszkoreit, Jones,
  Gomez, Kaiser, and Polosukhin}{Vaswani et~al.}{2017}]{vaswani2017attention}
Vaswani, A., N.~Shazeer, N.~Parmar, J.~Uszkoreit, L.~Jones, A.~N. Gomez,
  L.~Kaiser, and I.~Polosukhin (2017).
\newblock Attention is all you need.

\bibitem[\protect\citeauthoryear{Watson}{Watson}{1964}]{watson1964smooth}
Watson, G.~S. (1964).
\newblock Smooth regression analysis.
\newblock {\em Sankhy{\=a}: The Indian Journal of Statistics, Series A\/},
  359--372.

\bibitem[\protect\citeauthoryear{Williams}{Williams}{1975}]{Williams1975Mendenhall}
Williams, C.~B. (1975).
\newblock Mendenhall's studies of word-length distribution in the works of
  shakespeare and bacon.
\newblock {\em Biometrika\/}~{\em 62\/}(1), 207--212.

\bibitem[\protect\citeauthoryear{Yosinski, Clune, Bengio, and Lipson}{Yosinski
  et~al.}{2014}]{yosinski2014transferablefeaturesdeepneural}
Yosinski, J., J.~Clune, Y.~Bengio, and H.~Lipson (2014).
\newblock How transferable are features in deep neural networks?

\bibitem[\protect\citeauthoryear{Youden}{Youden}{1950}]{Youden1950Index}
Youden, W.~J. (1950).
\newblock Index for rating diagnostic tests.
\newblock {\em Cancer\/}~{\em 3\/}(1), 32--35.

\bibitem[\protect\citeauthoryear{Yule}{Yule}{1939}]{yule1939sentence}
Yule, G.~U. (1939, 01).
\newblock {On Setence - Length as a Statistical Characteristic of Style in
  Prose: With Application to Two Cases of Disputed Authorship}.
\newblock {\em Biometrika\/}~{\em 30\/}(3-4), 363--390.

\bibitem[\protect\citeauthoryear{Zhao, Zhou, Li, Tang, Wang, Hou, Min, Zhang,
  Zhang, Dong, Du, Yang, Chen, Chen, Jiang, Ren, Li, Tang, Liu, Liu, Nie, and
  Wen}{Zhao et~al.}{2023}]{zhao2023survey}
Zhao, W.~X., K.~Zhou, J.~Li, T.~Tang, X.~Wang, Y.~Hou, Y.~Min, B.~Zhang,
  J.~Zhang, Z.~Dong, Y.~Du, C.~Yang, Y.~Chen, Z.~Chen, J.~Jiang, R.~Ren, Y.~Li,
  X.~Tang, Z.~Liu, P.~Liu, J.-Y. Nie, and J.-R. Wen (2023).
\newblock A survey of large language models.

\bibitem[\protect\citeauthoryear{Zhuang, Qi, Duan, Xi, Zhu, Zhu, Xiong, and
  He}{Zhuang et~al.}{2020}]{zhuang2020comprehensivesurveytransferlearning}
Zhuang, F., Z.~Qi, K.~Duan, D.~Xi, Y.~Zhu, H.~Zhu, H.~Xiong, and Q.~He (2020).
\newblock A comprehensive survey on transfer learning.

\end{thebibliography}

\newpage
\appendix
\begin{center}
{\LARGE\bf Supplementary Material:}
\end{center}

\section{Detailed Review of \cite{Mosteller1963Inference}}\label{sec:detailed-MW}


The key component for determining the authorship in the paper \citep{Mosteller1963Inference} is the posterior log-odds. For each word count $x_w \in \mathcal{W}$,

\begin{equation*}\label{eq:posterior-log-odds}
\text{Posterior Odds(Hamilton to Madison)} = \frac{P(H|x_w)}{P(M|x_w)} = \underbrace{\frac{p_H}{p_M}}_{\text{initial odds}}\underbrace{\frac{\ell_H(x_w)}{\ell_M(x_w)}}_{\text{likelihood ratio}}
\end{equation*}
where $p_H$, $p_M$ denote the prior probability of observing a word $w \in \mathcal{W}$ for $x_w$ times per thousand words and $\ell_H$, $\ell_M$ denote the likelihood of $x_w$ in Hamilton's and Madison's document respectively. 

The difficulties in computation come from two components (1) the choice of initial odds and (2) the estimation of the unknown parameters in likelihood. The first problem is directly interpreted as how to choose the prior, but with more and better choices of discriminators (words), the likelihood term would overwhelm the influence of the initial odds, in either direction. As such, most of the work in \cite{Mosteller1963Inference} focused on the likelihood estimation part. 
Likelihood estimation is not easy for many reasons; there exists uncertainty in distribution, and even though we know the true distribution, parameter estimation is another problem. Lastly, the choice of words to use matters. To deal with these issues, the paper tested two distributions that are widely used to model the count of rare events: Poisson and negative binomial distribution. For parameter estimation of each distribution, appropriate models for the prior are introduced. 
In this review, we will only elaborate on the likelihood modeling for the negative binomial case. 

\subsubsection{Modeling the Likelihood}

The occurrence of a word per thousands of words are almost always a rare event that the Poisson distribution is a natural choice for model its distribution. However, for some words, like `\textit{may}' or `\textit{his}', the empirical distribution does not fit well by Poisson distribution \citep{mosteller1987statistical}. When modeling the counts with Poisson distribution, the main constraint posed is the mean and the variance should be equal. A negative binomial distribution can be an alternative to handle the overdispersed case for count data. In the Equation-\ref{eq:negative-binomial}, the mean is $\mu$, and the variance is $\mu(1+\mu/\kappa)$. Note that as $\kappa \to \infty$, the distribution converges to the Poisson distribution with rate $\mu$. We will call $\mu/\kappa$ as the measure of non-Poissonness. A negative binomial modeling can also be viewed as the mixture of Poisson and Gamma distribution. To be more precise, an author uses words following Poisson frequencies, but the rate changes from one document to another. If the rate follows the gamma distribution, then the resulting word distribution follows a negative binomial. In other words, $x|\mu \sim pois(\mu), \theta \sim \Gamma(\kappa, \kappa/\lambda)$, then $x \sim NB\Big( \kappa, \frac{1}{1+\mu/\kappa} \Big)$. Thus, the non-Poissonness parameter $\mu/\kappa$ can quantify how much word usage would change by documents and could further be used to determine the context-dependent words. 


\begin{equation}\label{eq:negative-binomial}
X \sim \text{NB}\Big(\kappa, \frac{1}{1+\mu/\kappa}\Big), \quad \pi(x) = \frac{\Gamma(x+\kappa)}{x!\Gamma(\kappa)}\Big(\frac{\mu/\kappa}{1+\mu/\kappa}\Big)^x \Big(\frac{1}{1+\mu/\kappa}\Big)^\kappa
\end{equation}


As noted earlier, even with the right distributional assumption (in this case, the word count follows a negative binomial distribution), the estimation of the parameter is not an easy problem. To circumvent this issue, an appropriate prior structure is introduced. For this setup, four parameters need to be estimated: the mean rate $\mu$, $\sigma = \mu_H + \mu_M$, $\tau = \frac{\mu_H}{\mu_H + \mu_M}$, the non-Poissonness parameter $\delta = \mu/\kappa$, $\lambda = \log(1+\delta)$ and $\xi =  \lambda_M + \lambda_H$,  $\eta = \lambda_H/\xi$. The relationship between these parameters are modeled through 5 underlying constants $\beta = (\beta_1, \beta_2, \beta_3, \beta_4, \beta_5)$, and we assume given $\beta$, 

\begin{enumerate}[label=(A\arabic*)]
    \item $(\sigma, \tau, \xi, \eta)$ are independent across words
    \item $(\sigma, \tau), \xi, \eta$ are independent of each other for each word
    \item $\sigma$ has a distribution that can be adequately approximated by a constant density
    \item $ \tau | \sigma \sim \text{Beta}(\beta_1 +\beta_2 \sigma,\beta_1 + \beta_2 \sigma)$ 
    \item $\eta \sim \text{Beta}(\beta_3, \beta_3)$
    \item $\xi \sim \Gamma(\beta_5,\frac{\beta_5}{\beta_4})$ with density $f(\xi|\beta_4, \beta_5) =\frac{(\beta_5/\beta_4)^{\beta_5}}{\Gamma(\beta_4)} \xi^{\beta_5-1} e^{-\frac{\beta_5}{\beta_4}\xi}$
\end{enumerate}

With this assumption, the full posterior of our interest, given the count for a word as $x_H, x_M$, is

\begin{align}\label{eq:full-posterior-nb}
p(\sigma, \tau, \xi, \eta| x_H, x_M) &= C(X) p(\sigma, \tau, \xi, \eta) p(x_H, x_M| \sigma, \tau, \xi, \eta)\\
& = C(X) p(\sigma, \tau)p(\xi)p(\eta)p(x_H, x_M| \sigma, \tau, \xi , \eta)
\end{align}

To avoid the infinite loop of putting prior over prior, the authors of the original work constrain the set of possible options for $\mathbf{\beta}$, and chose $(\beta_1, \beta_2, \beta_3, \beta_4, \beta_5) = (10, 0, 12, 0.83, 1.2)$ as their underlying constants for the final computation. An approximate method is applied to compute the posterior mode for each parameter, which in turn gives the estimated rates($\hat\mu_H, \hat\mu_M)$ for each author.

\subsubsection{Higher Criticism for Authorship Attribution}\label{sec:hc-authorship}
The Higher Criticism (HC) statistic is designed to assess the overall deviation of a set of p-values from their expected distribution under the null hypothesis \cite{Donoho2004Higher}. Assume you have a large set of test statistics $T_1, T_2, ..., T_n$ corresponding to hypotheses $ H_1, H_2, ..., H_n $. Under the null hypothesis $H_i$ (no signal), each test statistic $T_i$ is transformed into a p-value $p_i $. Given the sorted the p-values $p_{(1)} \leq p_{(2)} \leq ... \leq p_{(n)}$, the HC statistic is defined as: 
$$HC_n = \max_{1 \leq i \leq n} \left| \frac{i/n - p_{(i)}}{\sqrt{(i/n)(1 - i/n)}} \right|$$ This formula compares the empirical distribution of the p-values to the uniform distribution, normalized by the standard error. A high value of \( HC_n \) indicates that there are more small p-values than expected under the null hypothesis, suggesting the presence of a signal.

To solve the authorship attribution task in multiple testing perspectives, Kipnis\cite{Kipnis2022Higher} redefine the authorship prediction task as detecting the difference between two large word frequency tables. For the number of occurrences $x_w$ of each word $w \in \mathcal{W}$, perform the binomial test to see if there is a difference in frequency among the documents written by different authors. Consider the simplest case where there are only two documents. For a word $w \in \mathcal{W}$, $x_w$ denotes the number of occurrences of the word in a document $D_1$ as $N(w|D_1)$.

\begin{align*}
    H_0 &: N(x_w|D_1) \sim Bin(m,q)\\
    H_1 &: \text{otherwise}
\end{align*} where $m = N(x_w|D_1) + N(x_w|D_2)$, $q = \frac{\sum_{w' \in \mathcal{W}}N(x_{w'}|D_1)}{\sum_{w' \in \mathcal{W}, w' \neq w} N(x_{w'}|D_1) + N(x_{w'}|D_2)}$. The p-value under the null hypothesis is given as 
\begin{equation}\label{eq:binomial-allocation}
    \pi(x_w|D_1, D_2) = \mathbb{P}(|Bin(m,q) - mq| \geq |N(x_w|D_1)-mq|).
\end{equation}
By applying the test to all $w \in \mathcal{W}$, we have $n = |\mathcal{W}|$ number of p-values where we can apply HC framework. 

Their main insight is to view the HC statistics comparing two word-frequency tables as a distance. Since the HC statistics being large implies the distribution of p-values deviates a lot from a uniform distribution under the null, which further implies there is a discrepancy in the usage of words between the two documents. 

\begin{equation}\label{eq:HC-dist}
d_{HC}(D_1, D_2) = HC^\dagger = \underset{1 \leq i \leq \gamma_0 N, 1/N \leq \pi_{(i)}}{\text{max}}\sqrt{N} \frac{i/N - \pi_{(i)}}{\sqrt{\frac{i}{N} (1-\frac{i}{N})}}
\end{equation}
With this distance, the decision rule for the authorship is if $d_{HC}(D, \mathcal{D}_{\text{Madison}}) < d_{HC}(D, \mathcal{D}_{\text{Hamilton}})$, then author of the document $D$  is Madison.


Additionally, they utilize the \textit{HC threshold}, which can be used to identify the set of words that induce the large deviations. In the authorship attribution task, it can be used to identify the most discriminative words between the authors. 
$$ t_{HC} = \pi_{(i^*)}, i^* = \underset{1 \leq i \leq \gamma_0 N, 1/N \leq \pi_{(i)}}{\text{argmax}}\sqrt{N} \frac{i/N - \pi_{(i)}}{\sqrt{\frac{i}{N} (1-\frac{i}{N})}}$$
They identify 378 words whose p-value falls below the HC threshold. The word `upon' turns out to be the most significant (with the smallest p-value), and so is the set of the most frequent words (``B3B" group). On top of the words used in \cite{Mosteller1963Inference}, a new set of words is revealed to be significant. `would' is more likely to be found in Madison's document; on the other hand, words such as `power', `department', `congress', confederation' are more likely to be used in Hamilton's document. 
The prediction and the HC discrepency values are summarized in Table~\ref{tab:kipnis-res-summary}.


\begin{table}[H]
\centering
\begin{tabular}{@{}ccccccc@{}}
\toprule
\textbf{Papers} & \textbf{Type} & \multicolumn{3}{c}{\textbf{HC}} & \textbf{Decision} \\ 
\cmidrule(lr){3-5}
 & & \textbf{Hamilton} & \textbf{Madison} & \textbf{Diff} & \\ \midrule
49 & disputed & 3.0210 & 2.6166 & 0.4044 & Madison \\
50 & disputed & 3.8154 & 3.1456 & 0.6698 & Madison \\
51 & disputed & 4.5327 & 2.6591 & 1.8737 & Madison \\
52 & disputed & 4.0433 & 2.4977 & 1.5456 & Madison \\
53 & disputed & 4.4087 & 4.0416 & 0.3672 & Madison \\
54 & disputed & 4.7123 & 4.2628 & 0.4495 & Madison \\
55 & disputed & 4.4918 & 3.6148 & 0.8770 & Madison \\
56 & disputed & 4.6521 & 4.0536 & 0.5985 & Madison \\
57 & disputed & 4.8585 & 4.2526 & 0.6059 & Madison \\
58 & disputed & 2.6117 & 2.4343 & 0.1774 & Madison \\
62 & disputed & 3.6237 & 1.5109 & 2.1128 & Madison \\
63 & disputed & 4.7131 & 3.2023 & 1.5108 & Madison \\
18 & joint & 4.5254 & 3.7611 & 0.7643 & Madison \\
19 & joint & 5.0771 & 4.5206 & 0.5565 & Madison \\
20 & joint & 3.2669 & 3.3846 & -0.1177 & Hamilton \\ \bottomrule
\end{tabular}
\caption{HC distance (\ref{eq:HC-dist}) to each of the author from \cite{Kipnis2022Higher}. The “Diff” column represents the difference between the distance to Hamilton and the distance to Madison. A positive value indicates that the paper is assigned to Madison, as it is closer to his authorship profile. All the disputed papers are attributed to Madison. A weak support for Madison's authorship on jointly authored No. 20 aligns with the original study by \cite{Mosteller1963Inference}.}
\label{tab:kipnis-res-summary}
\end{table}

\section{Additional Figures and Tables}

\begin{figure}[H]
    \centering
        \begin{subfigure}{.40\textwidth}
        \includegraphics[width=\linewidth]{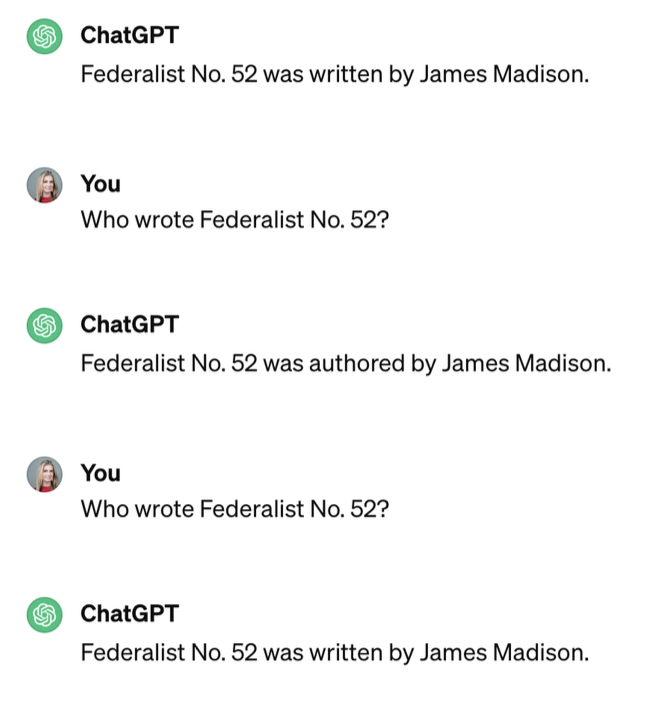}
            \caption{}
        \end{subfigure}
        \begin{subfigure}{.54\textwidth}
        \includegraphics[width=\linewidth]{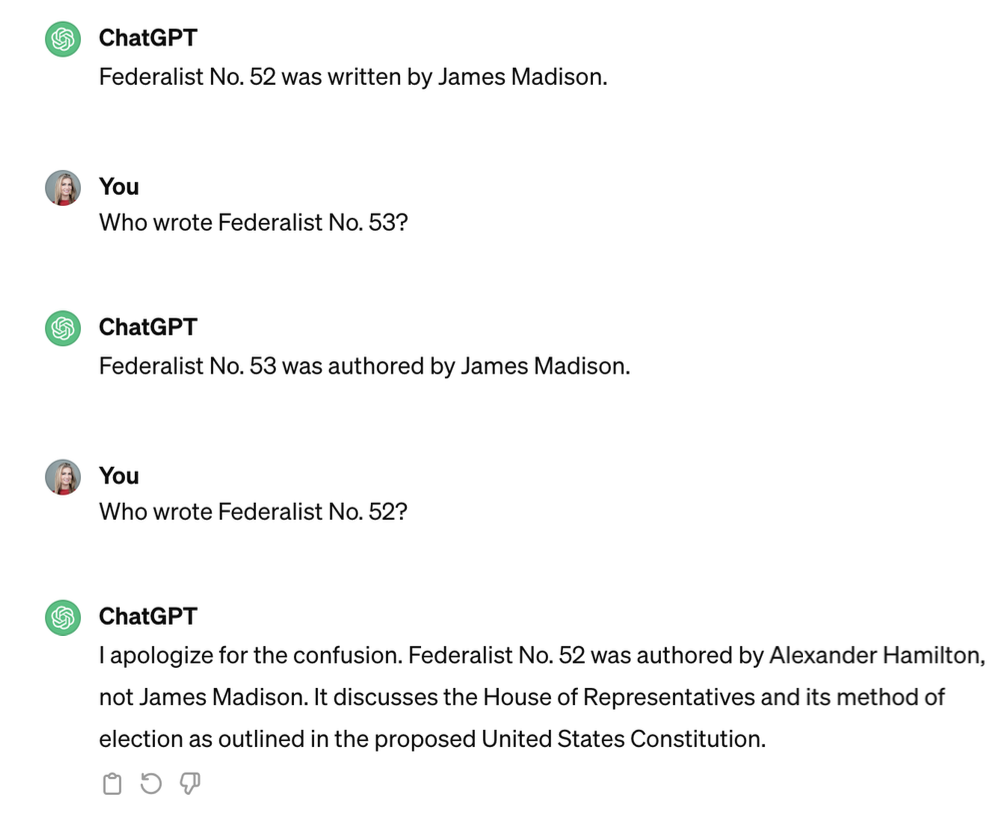}
            \caption{}
        \end{subfigure}
        \begin{subfigure}{.40\textwidth}
       \includegraphics[width=\linewidth]{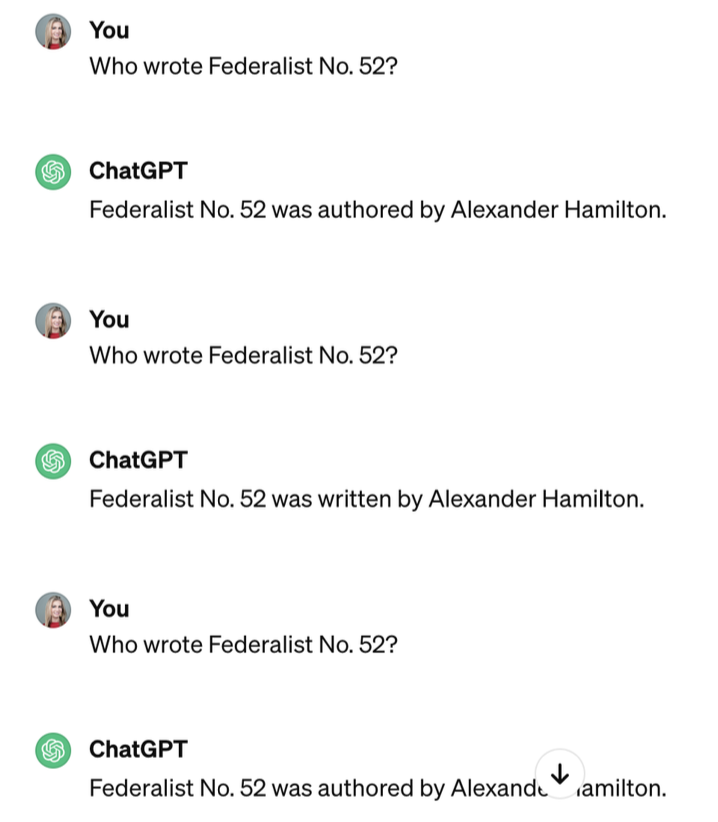}
            \caption{}
        \end{subfigure}
        \begin{subfigure}{.54\textwidth}
        \includegraphics[width=\linewidth]{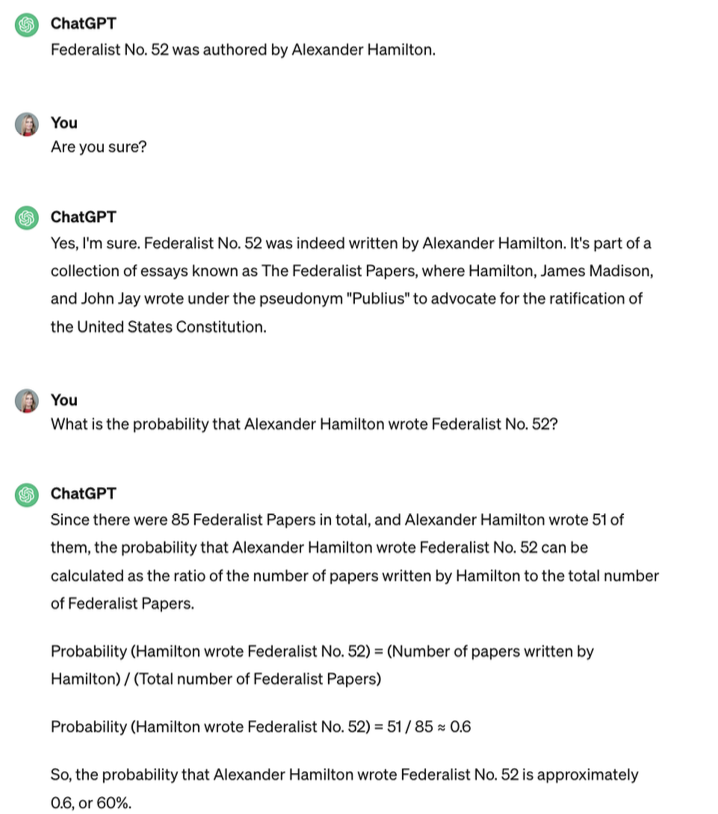}
            \caption{}
        \end{subfigure}
    \caption{An example that GPT can provide a plausible answer, but they cannot convey the answer with uncertainty. Their responses change from time to time even with the same prompt.}
    \label{fig:intro-gpt-uncertain}
\end{figure}

\begin{figure}
    \centering
    \begin{subfigure}{.32\textwidth}
    \centering
    \includegraphics[width=.95\linewidth]{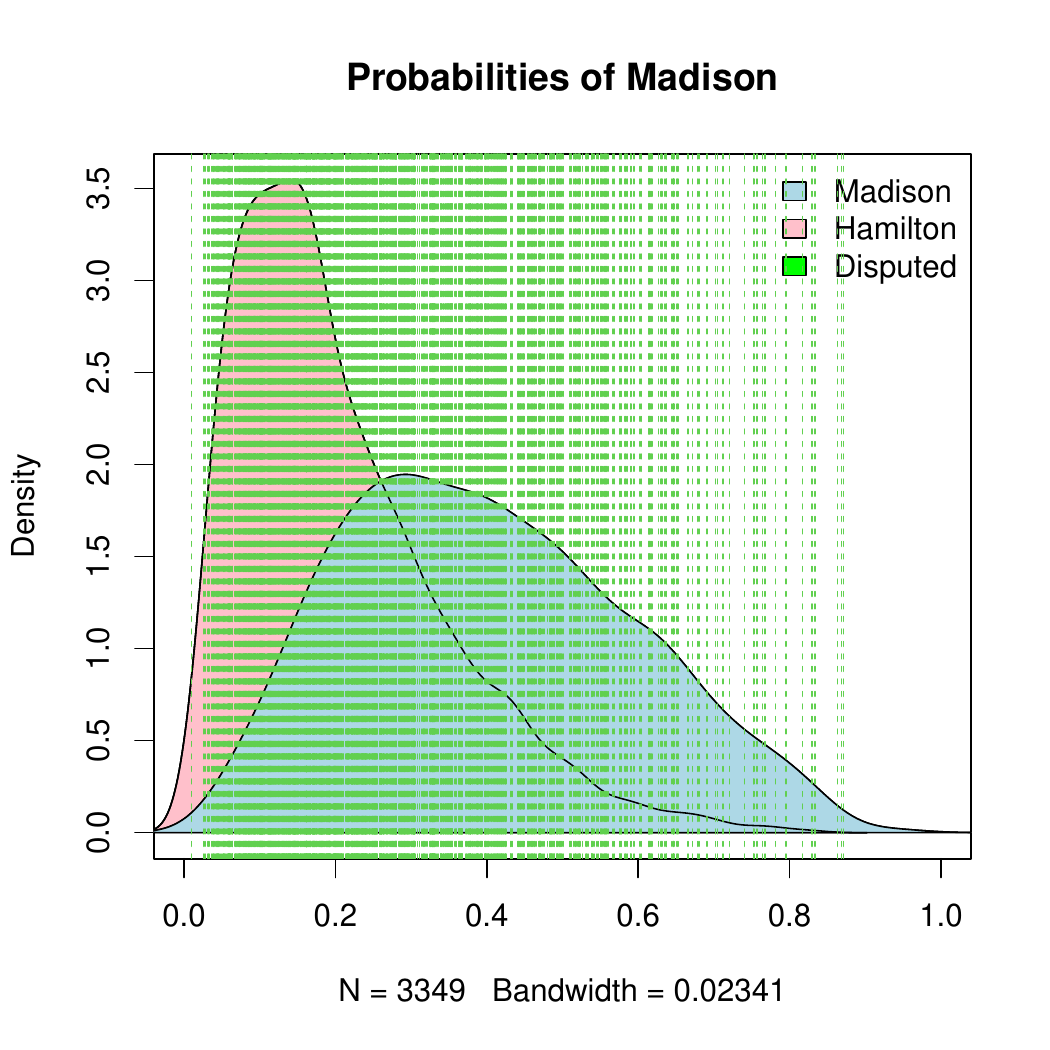}
    \end{subfigure}
    \begin{subfigure}{.32\textwidth}
    \centering
    \includegraphics[width=.95\linewidth]{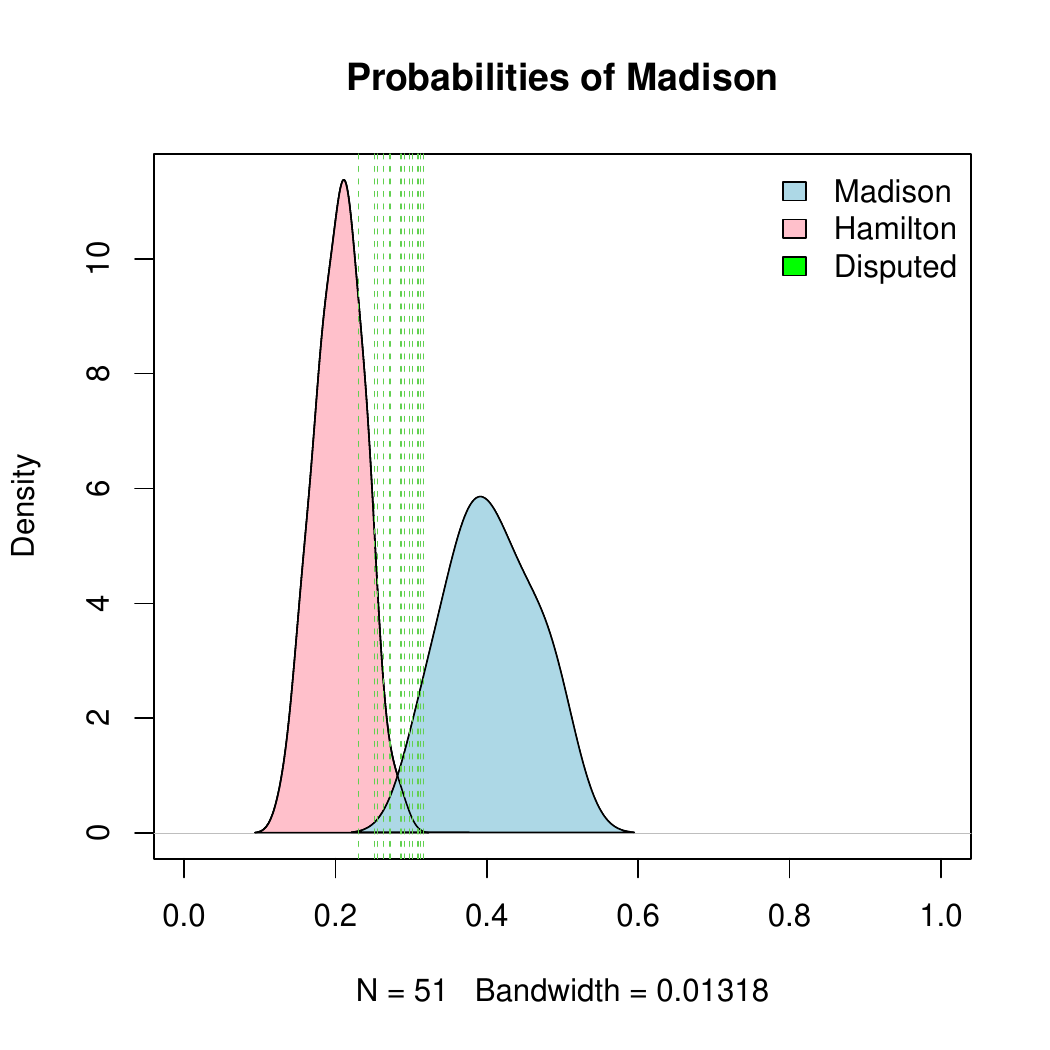}
    \end{subfigure}
    \begin{subfigure}{.32\textwidth}
    \centering
    \includegraphics[width=.95\linewidth]{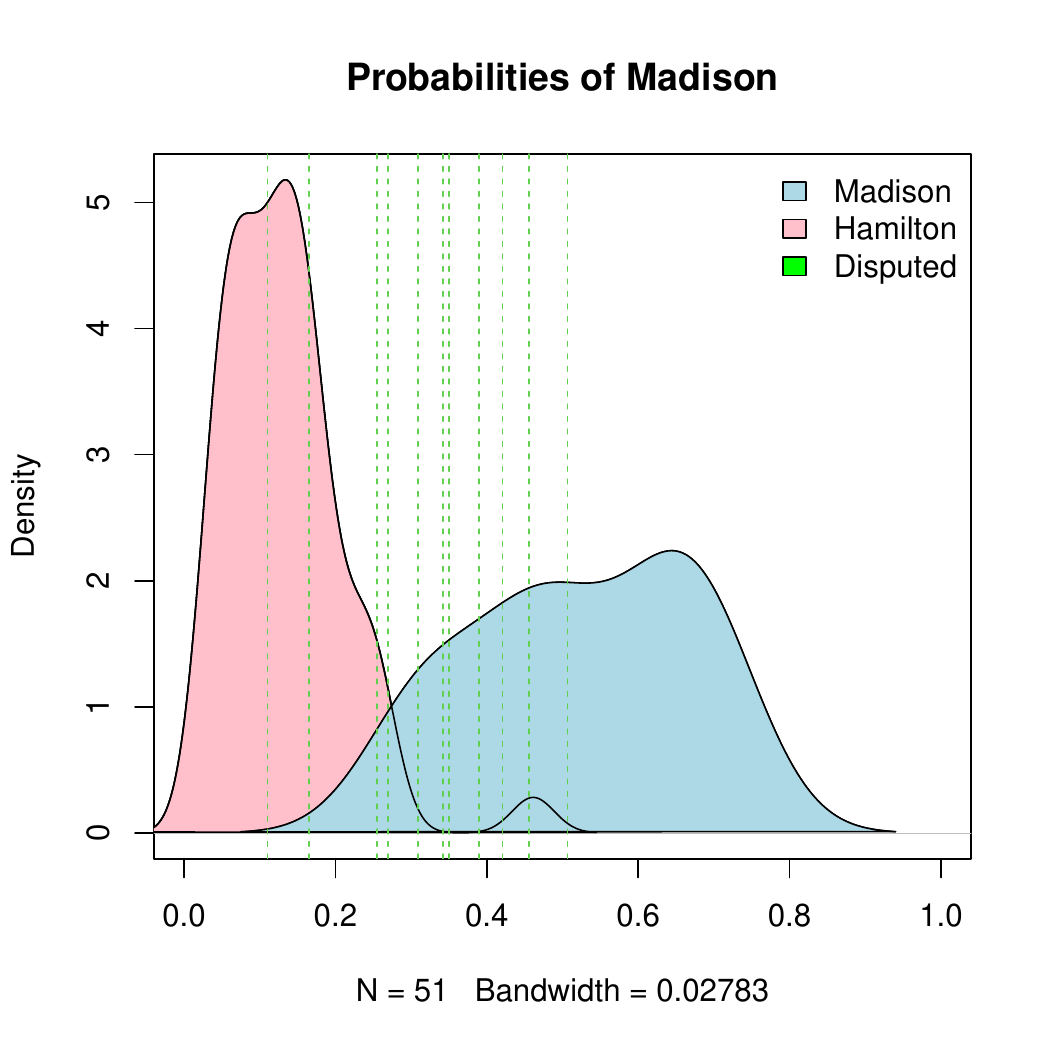}
    \end{subfigure}
    \caption{BART classification probability based on Sentence Embeddings. The predicted probability based on sentence embeddings (left). The predicted probability aggregated for each document based on sentence embeddings (middle). The predicted probability based on the aggregated sentence embeddings (right). Essentially, first two plots are based on sentence-wise embedding, but the third figure is based on document embedding calculated from sentence-wise embeddings.}
    \label{fig:sentence}
\end{figure}

\begin{figure}
    \centering
    \begin{subfigure}{.32\textwidth}
    \centering
    \includegraphics[width=.95\linewidth]{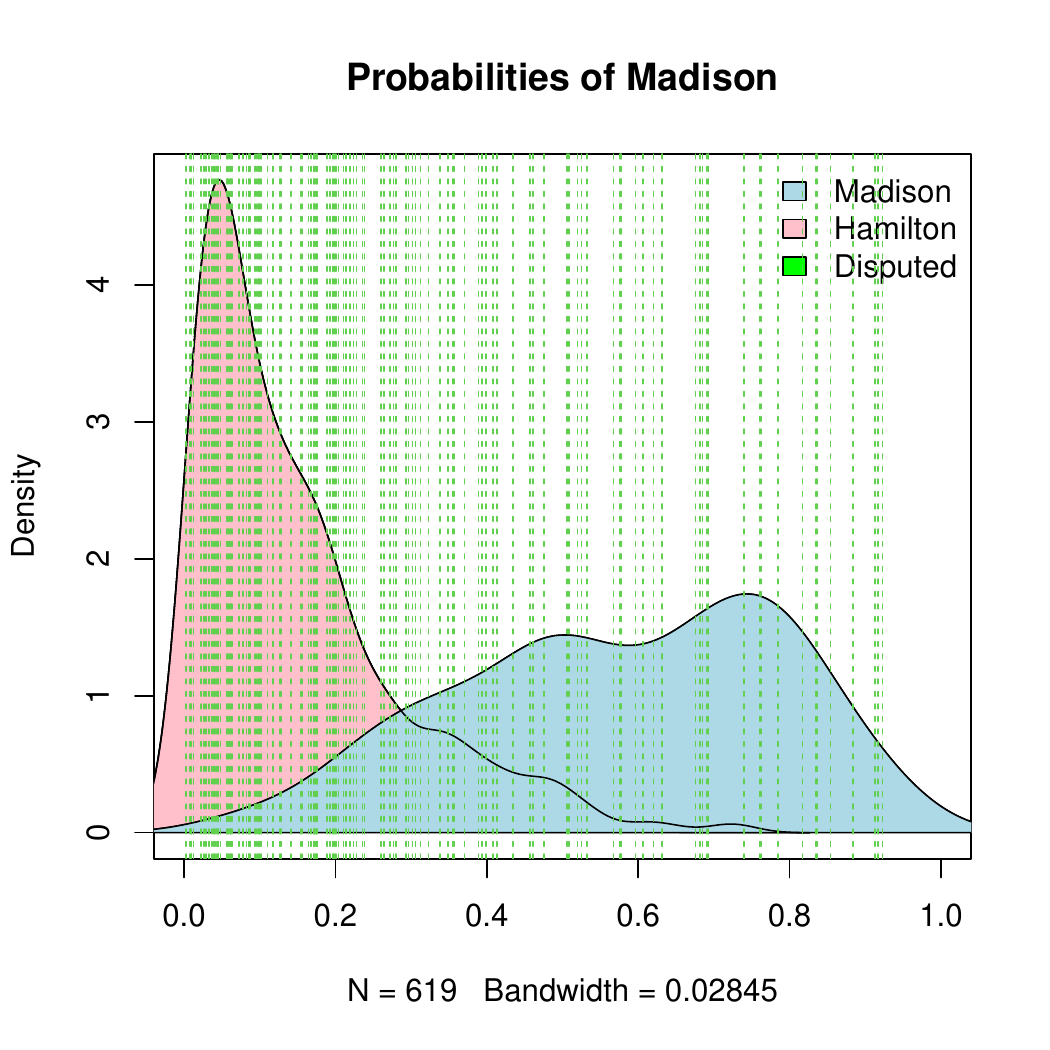}
    \end{subfigure}
    \begin{subfigure}{.32\textwidth}
    \centering
    \includegraphics[width=.95\linewidth]{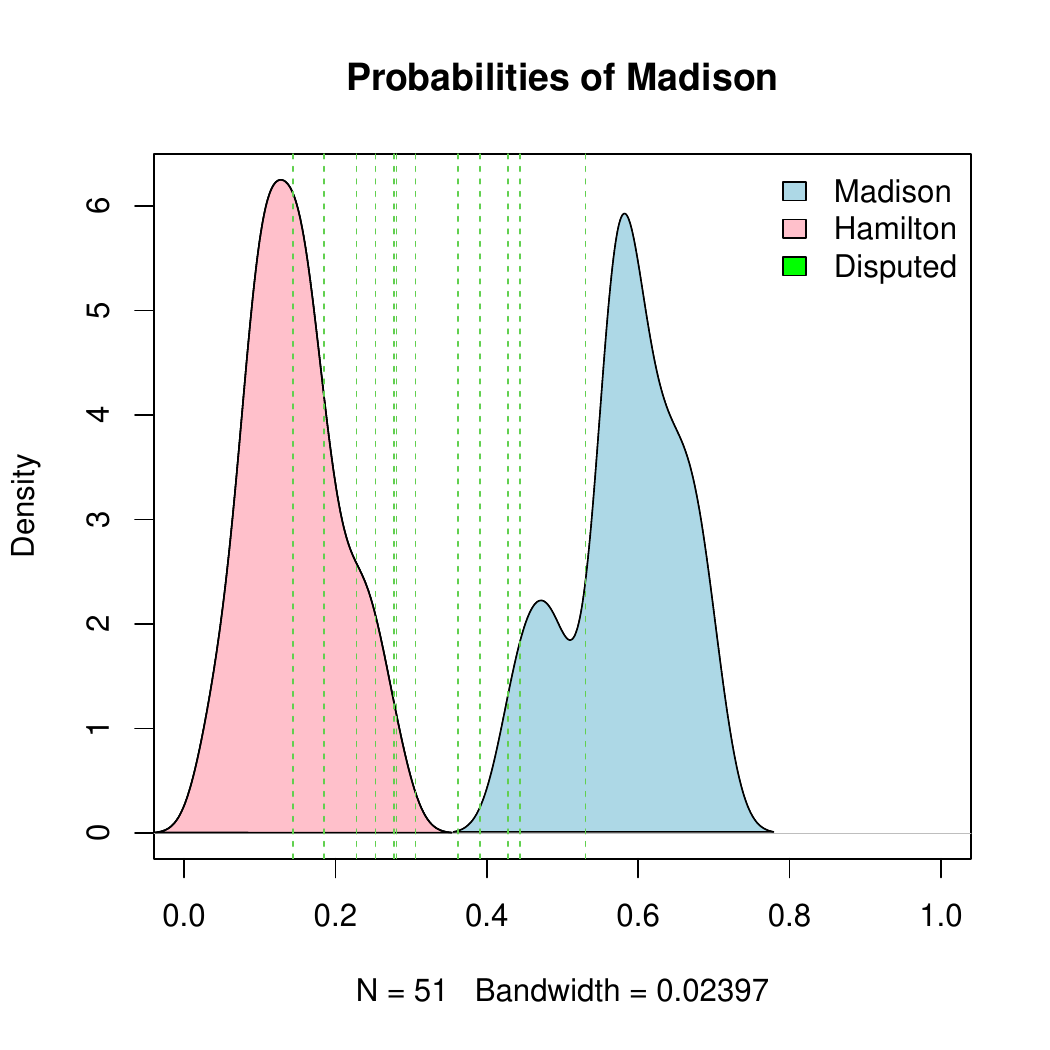}
    \end{subfigure}
    \begin{subfigure}{.32\textwidth}
    \centering
    \includegraphics[width=.95\linewidth]{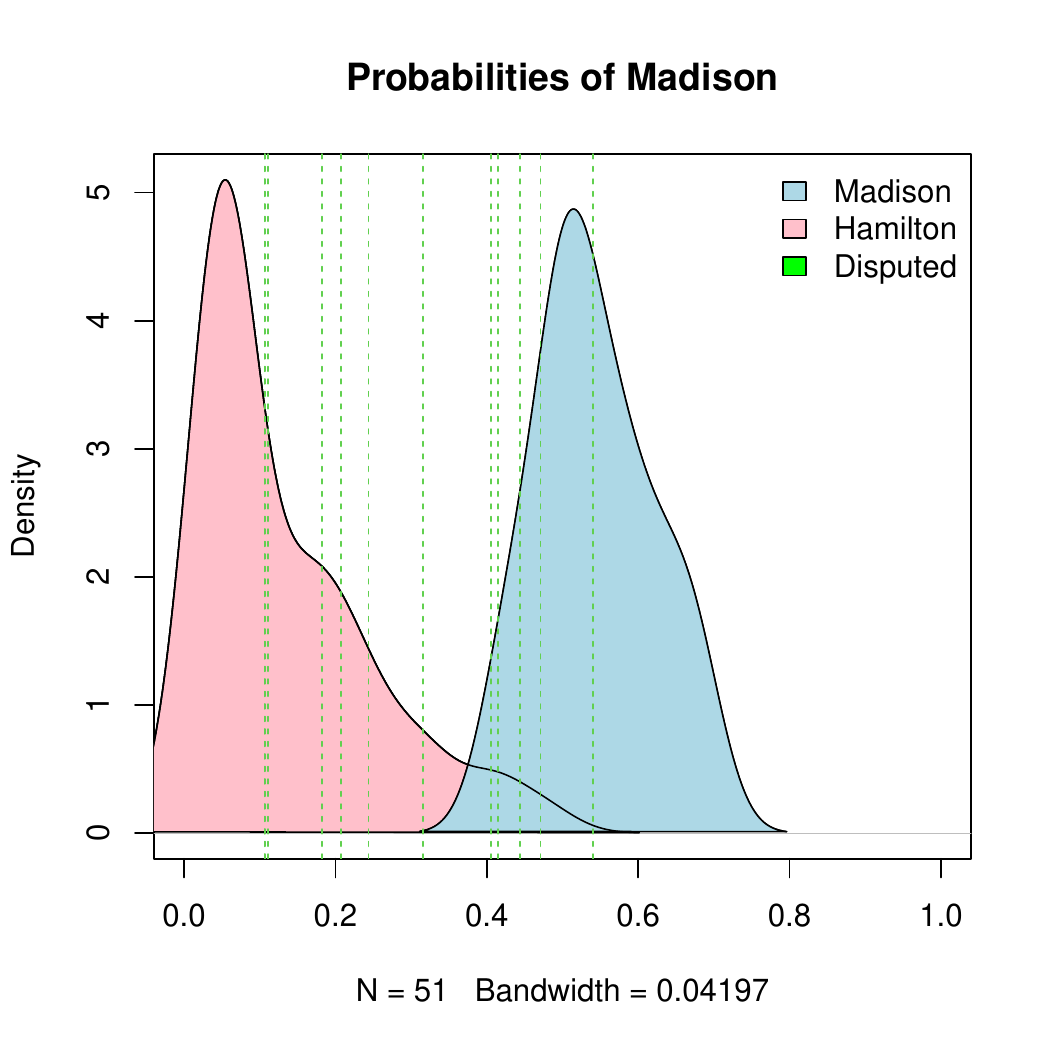}
    \end{subfigure}
     \caption{BART classification probability based on Chunk (size = 200) Embeddings. Same principle as Figure~\ref{fig:sentence} has been applied. }
    \label{fig:chunk}
\end{figure}

\begin{figure}
    \centering
    \includegraphics[width=\textwidth]{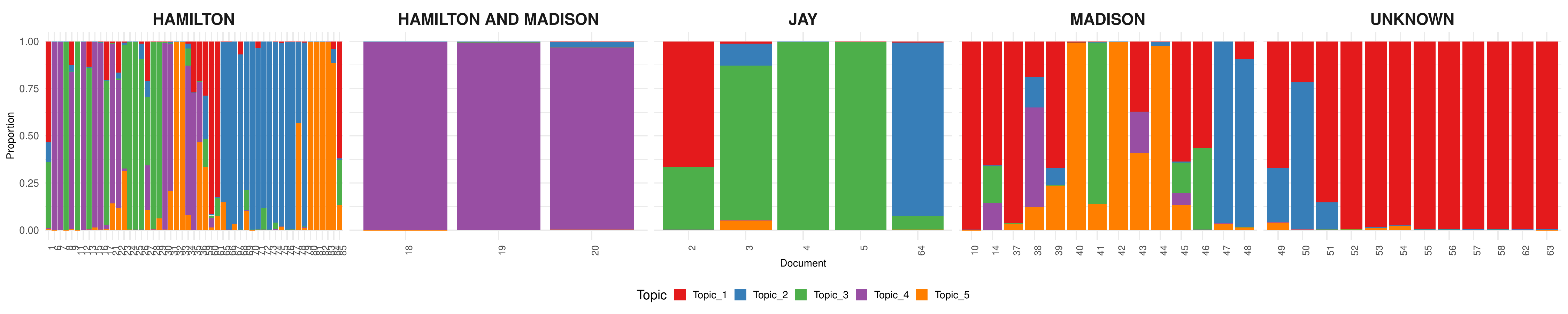}
    \caption{Document-Topic distribution by LDA trained on contextual words after removing stopwords (Type 1). Jay's expertise on foreign policy is noticeable through the topic distribution; however, the attribution of the disputed papers is less clear. }
    \label{fig:doc-topic-distribution-tdm1}
\end{figure}

\begin{figure}
    \centering
    \includegraphics[width=\textwidth]{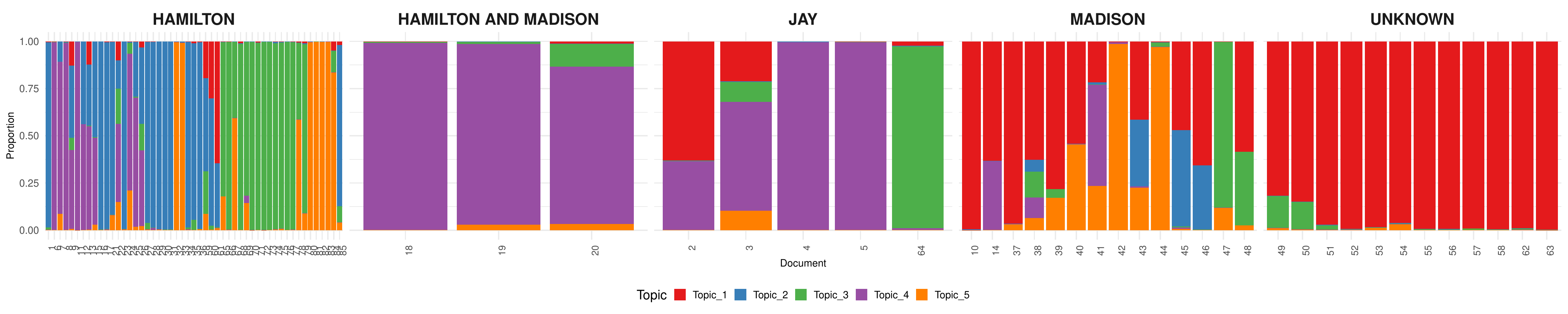}
    \caption{Document-Topic distribution by LDA trained on contextual words \textit{without} removing stopwords (Type 2). The topic distributions across different authors are similar to the ones revealed using only the contextual words (Figure~\ref{fig:doc-topic-distribution-tdm1}).}
    \label{fig:doc-topic-distribution-tdm2}
\end{figure}

\begin{figure}
    \centering
    \begin{subfigure}{.32\textwidth}
    \includegraphics[width=\linewidth]{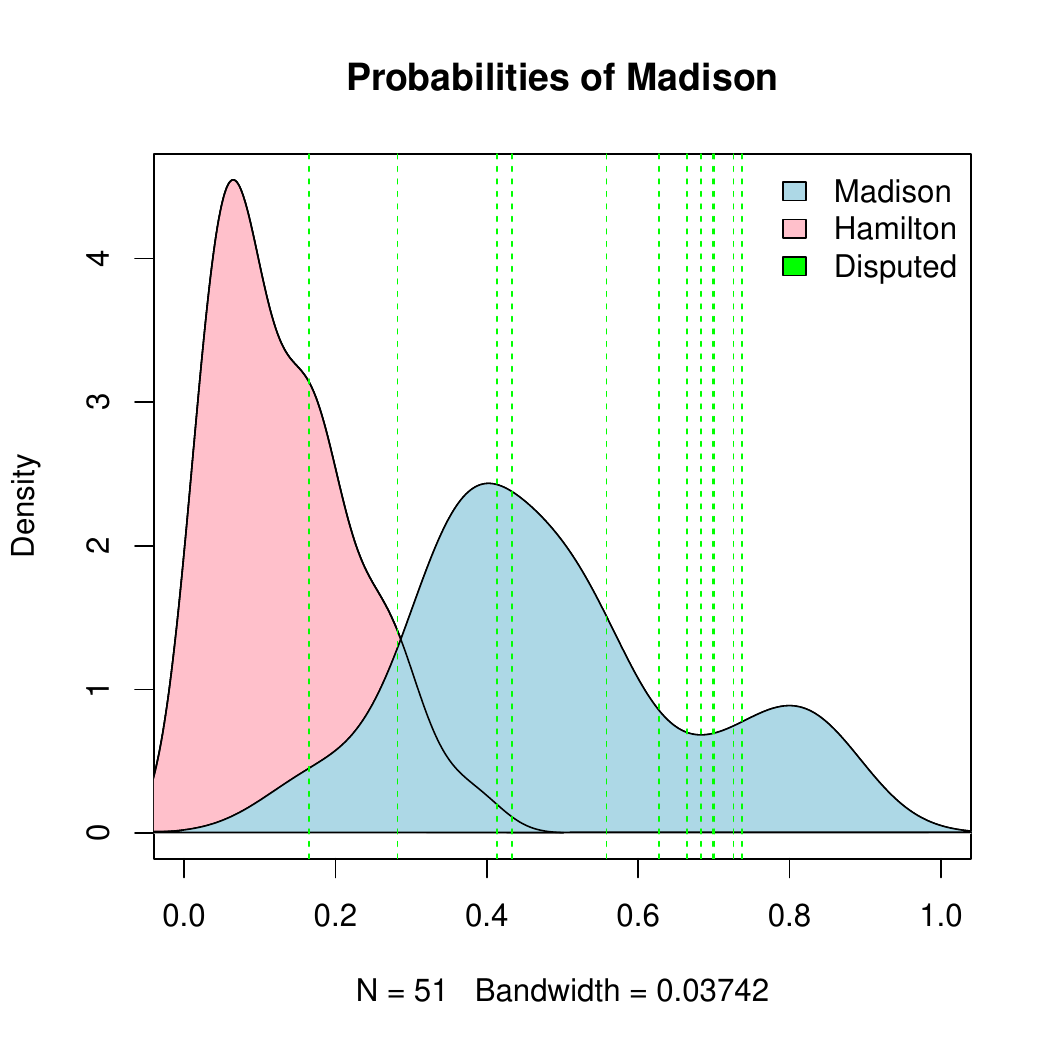}
        \caption{Type 1}
    \end{subfigure}
    \begin{subfigure}{.32\textwidth}
        \includegraphics[width=\linewidth]{figure/lda2.pdf}
        \caption{Type 2}
    \end{subfigure}
    \begin{subfigure}{.32\textwidth}
        \includegraphics[width=\linewidth]{figure/lda3.pdf}
        \caption{Type 3}
    \end{subfigure}
    \caption{BART classification probability using LDA document embedding with different set of words as input. Type 1 includes contextual words only, Type 2 includes both contextual words and stopwords, and Type 3 includes a curated set of words by \cite{Mosteller1963Inference}. Note that the results align with our findings on Topic distribution in Section~\ref{sec:topics}. When using Type 3 inputs, the distributions of Hamilton and Madison are completely separated. With Type 2 inputs, where stopwords are added on top of the contextual words, the model consistently identifies all disputed papers as being authored by Madison.}
    \label{fig:predicted-density-lda}
\end{figure}

\begin{figure}
    \centering
    \begin{subfigure}{.32\textwidth}
    \includegraphics[width=\linewidth]{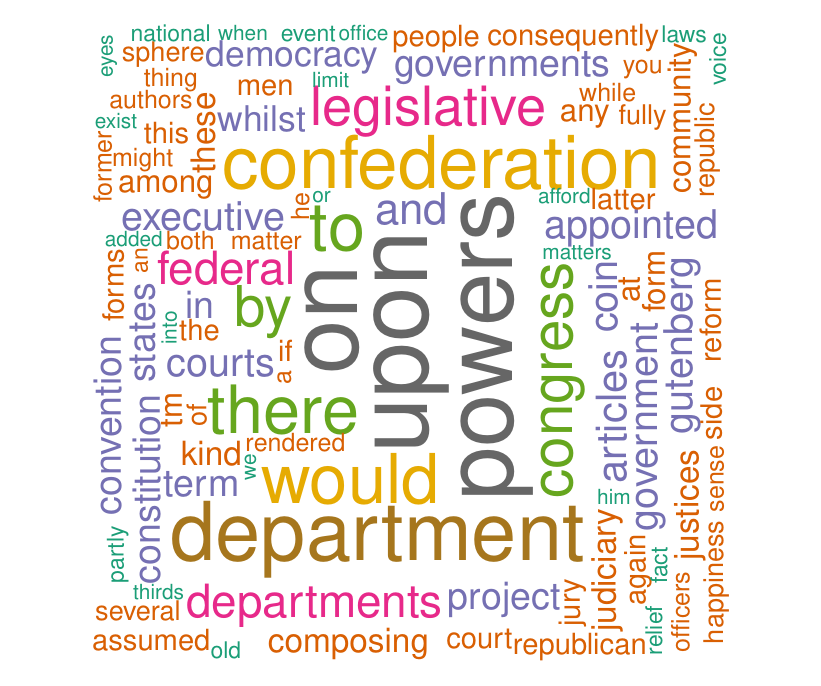}
        \caption{HC threshold}
    \end{subfigure}
    \begin{subfigure}{.32\textwidth}
        \includegraphics[width=\linewidth]{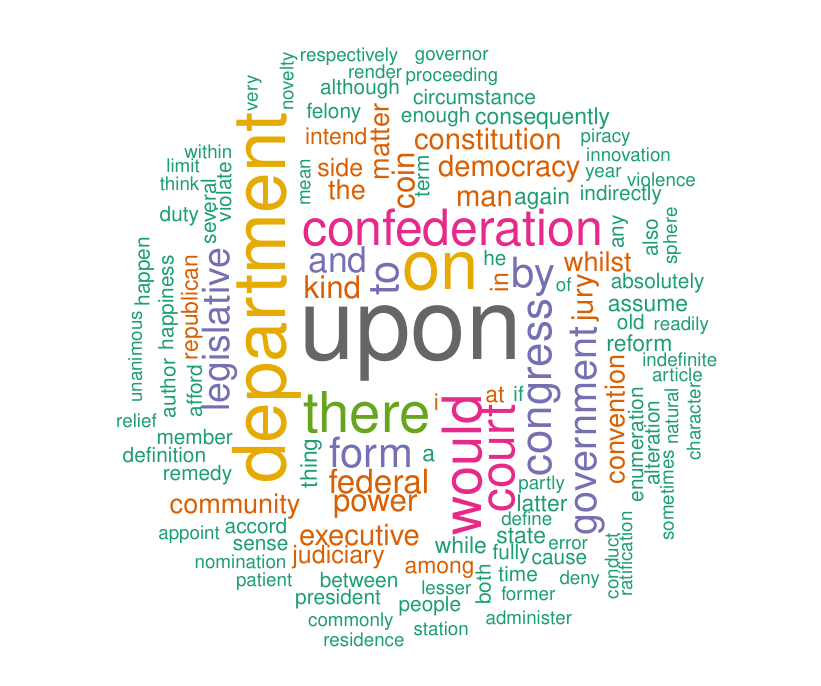}
        \caption{Benjamini-Hochberg}
    \end{subfigure}
    \begin{subfigure}{.32\textwidth}
        \includegraphics[width=\linewidth]{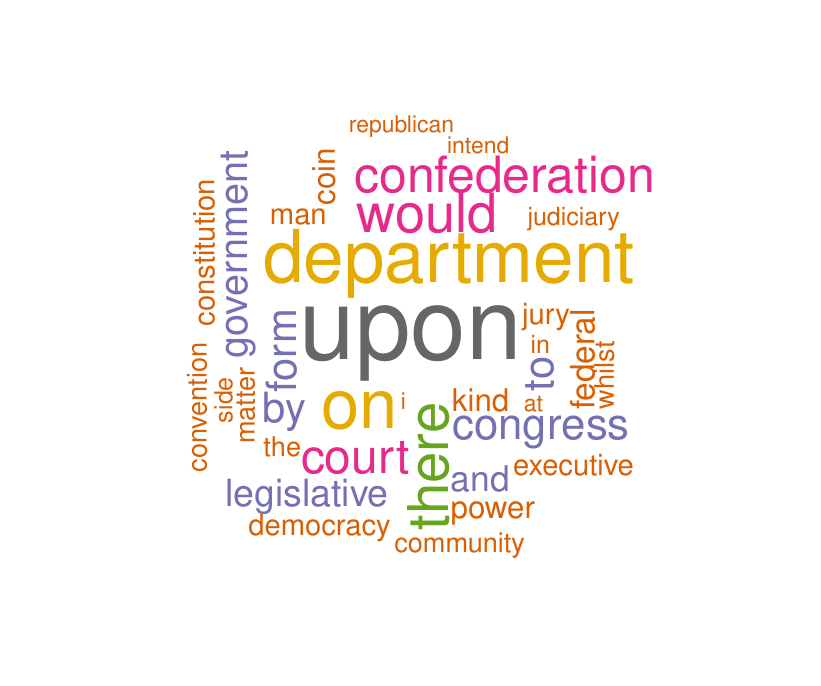}
        \caption{Bonferroni}
    \end{subfigure}
    \caption{Word cloud representation of significant words found by different screening procedures. \cite{Kipnis2022Higher} view HC statistics as distance and use corresponding HC threshold to identify the most discriminative words. We follow the binomial allocation model used in \cite{Kipnis2022Higher} (\ref{eq:binomial-allocation}). With false-discovery rate set to 0.1, 117 words are found by Benjamini-Hochberg and 35 words are found by Bonferroni correction. Note that 378 words fall below the HC threshold by \cite{Kipnis2022Higher}. We observe that the different procedures yield different sets of vocabularies. For all three testing framework, `upon', `on' and `while', `whilst' are revealed. }
    \label{fig:words-clouds-mt}
\end{figure}

\begin{table}[H]
\centering
\resizebox{0.5\textwidth}{!}{%
\begin{tabular}{@{}cllll@{}}
\toprule
 & \multicolumn{2}{c}{LASSO}                        & \multicolumn{2}{c}{BART}                         \\ \midrule
 & \multicolumn{1}{c}{ROC} & \multicolumn{1}{c}{F1} & \multicolumn{1}{c}{ROC} & \multicolumn{1}{c}{F1} \\ \midrule
BoW 1    & 0.1915 & 0.2455 & 0.2364 & 0.2884 \\
BoW 2    & 0.2002 & 0.2031 & 0.1573 & 0.2131 \\
BoW 3    & 0.2448 & 0.2829 & 0.3312 & 0.3949 \\
LDA 1    & 0.2067 & 0.1985 & 0.1427 & 0.1453 \\
LDA 2    & 0.2200 & 0.2213 & 0.2433 & 0.3184 \\
LDA 3    & 0.3768 & 0.6661 & 0.4125 & 0.5610 \\
LSA 1    & 0.2040 & 0.0757 & 0.1877 & 0.3053 \\
LSA 2    & 0.3936 & 0.5014 & 0.3876 & 0.3939 \\
LSA 3    & 0.5899 & 0.6813 & 0.3350 & 0.3560 \\
NMF 1    & 0.1968 & 0.1969 & 0.3076 & 0.3145 \\
NMF 2    & 0.2483 & 0.3673 & 0.1885 & 0.3073 \\
NMF 3    & 0.3768 & 0.4712 & 0.4296 & 0.4357 \\
Word2Vec 1    & 0.2106 & 0.1838 & 0.2042 & 0.2055 \\
Word2Vec 2    & 0.2007 & 0.2031 & 0.2857 & 0.2923 \\
Word2Vec3    & 0.2859 & 0.3145 & 0.2783 & 0.2869 \\
BERT    & 0.1783 & 0.2685 & 0.1722 & 0.5347 \\
RoBERTa & 0.1984 & 0.2940 & 0.1264 & 0.1277 \\
BART    & 0.1850 & 0.1858 & 0.1636 & 0.1683 \\
GPT4    & 0.2175 & 0.1943 & 0.2279 & 0.2284 \\
Llama2  & 0.2121 & 0.0971 & 0.3229 & 0.3253 \\
Llama3  & 0.2056 & 0.1603 & 0.2391 & 0.2402 \\ \bottomrule
\end{tabular}%
}
\caption{Classification thresholds computed by ROC and F1 rule. The number right next to the non-sequential methods indicate the type of word inputs. BoW 1 is the threshold computed on the word count matrix with Type 1 inputs. }
\label{tab:threshold}
\end{table}

\begin{table}[H]
\resizebox{\textwidth}{!}{%
\begin{tabular}{@{}cccccccccccc@{}}
\toprule
 & \multicolumn{4}{c}{BoW Embeddings} & \multicolumn{7}{c}{Continuous Embeddings}                 \\ \midrule
 & BoW     & LDA    & LSA    & NMF    & Word2Vec & BERT & RoBERTa & BART & GPT4 & Llama2 & Llama3 \\ \midrule
LASSO &  0.1231 & 0.0923 & \textbf{0.0154} & 0.0769 & 0.3692 & 0.2923 & \textbf{0.2154} & 0.4154 & 0.6308 & 0.9077          & 0.8615 \\
BART  &  0.1231 & 0.0923 & \textbf{0.0308} & 0.1231 & 0.1231 & 0.2769 & 0.3846          & 0.3538 & 0.2615 & \textbf{0.1385} & 0.3077 \\ \bottomrule
\end{tabular}%
}
\caption{Classification Error with a threshold chosen by ROC. The BoW results are based on the Type 2 inputs. The lowest (best) values within each group are bolded.}
\label{tab:classification_error_roc}
\end{table}

\begin{table}[H]
\resizebox{\textwidth}{!}{%
\begin{tabular}{@{}ccccccccccc@{}}
\toprule
       & \multicolumn{5}{c}{LASSO}                    & \multicolumn{5}{c}{BART}                              \\ \midrule
       & BoW    & LDA    & LSA    & NMF    & Word2Vec & BoW    & LDA    & LSA             & NMF    & Word2Vec \\ \midrule
Type 1 & 0.1538 & 0.9385 & 0.9077 & 0.4154 & 0.8000   & 0.1077 & 0.4000 & 0.1385          & 0.2154 & 0.2462   \\
Type 2 & 0.1231 & 0.0923 & 0.0154 & 0.0769 & 0.3692   & 0.1231 & 0.0923 & \textbf{0.0308} & 0.1231 & 0.1231   \\
Type 3 &
  \textbf{0.0308} &
  \textbf{0.0000} &
  \textbf{0.0000} &
  \textbf{0.0000} &
  \textbf{0.0769} &
  \textbf{0.0462} &
  \textbf{0.0000} &
  0.0462 &
  \textbf{0.0000} &
  \textbf{0.0615} \\ \bottomrule
\end{tabular}%
}
\caption{Classification Error for BoW embeddings with different types of inputs. A threshold is chosen by ROC.}
\label{tab:classification_error_roc_bow}
\end{table}

\begin{table}[H]
\resizebox{\textwidth}{!}{%
\begin{tabular}{@{}cccccccccccc@{}}
\toprule
 & \multicolumn{4}{c}{BoW Embeddings} & \multicolumn{7}{c}{Continuous Embeddings}                 \\ \midrule
 & BoW     & LDA    & LSA    & NMF    & Word2Vec & BERT & RoBERTa & BART & GPT4 & Llama2 & Llama3 \\ \midrule
LASSO & 0.1231 & 0.0923 & \textbf{0.0154} & 0.0615 & 0.3692  & 0.1385 & \textbf{0.1077} & 0.4154 & 0.5692 & 0.7692          & 0.7692 \\
BART  & 0.1077 & 0.0769 & \textbf{0.0308} & 0.0923 & 0.1231 & 0.1538 & 0.3846          & 0.3538 & 0.2615 & \textbf{0.1385} & 0.3077 \\ \bottomrule
\end{tabular}%
}
\caption{Classification Error with a threshold chosen by $F_1$.  The BoW results are based on the Type 2 inputs. The lowest (best) values within each group are bolded.}
\label{tab:classifcation_error_f1}
\end{table}

\begin{table}[H]
\resizebox{\textwidth}{!}{%
\begin{tabular}{@{}ccccccccccc@{}}
\toprule
       & \multicolumn{5}{c}{LASSO}                    & \multicolumn{5}{c}{BART}                              \\ \midrule
       & BoW    & LDA    & LSA    & NMF    & Word2Vec & BoW    & LDA    & LSA             & NMF    & Word2Vec \\ \midrule
Type 1 & 0.0769 & 0.7385 & 0.7846 & 0.4154 & 0.6615   & 0.0923 & 0.4000 & 0.0923          & 0.2154 & 0.2462   \\
Type 2 & 0.1231 & 0.0923 & 0.0154 & 0.0615 & 0.3692   & 0.1077 & 0.0769 & \textbf{0.0308} & 0.0923 & 0.1231   \\
Type 3 &
  \textbf{0.0308} &
  \textbf{0.0000} &
  \textbf{0.0000} &
  \textbf{0.0000} &
  \textbf{0.0769} &
  \textbf{0.0462} &
  \textbf{0.0000} &
  0.0462 &
  \textbf{0.0000} &
  \textbf{0.0615} \\ \bottomrule
\end{tabular}%
}
\caption{Classification Error for BoW embeddings with different types of inputs. A threshold is chosen by $F_1$.}
\label{tab:classification_error_bow_f1}
\end{table}

\begin{table}[H]
\resizebox{\textwidth}{!}{%
\begin{tabular}{@{}cccccccccccc@{}}
\toprule
 & \multicolumn{4}{c}{BoW Embeddings} & \multicolumn{7}{c}{Continuous Embeddings}                 \\ \midrule
 & BoW     & LDA    & LSA    & NMF    & Word2Vec & BERT & RoBERTa & BART & GPT4 & Llama2 & Llama3 \\ \midrule
LASSO &  0.1538 & 0.1077 & \textbf{0.0154} & 0.0923 & 0.2615  & 0.1692 & \textbf{0.1231} & 0.2308 & 0.2615 & 0.2308          & 0.2154 \\
BART  & 0.1077 & 0.0923 & \textbf{0.0615} & 0.0923 & 0.1385   & 0.2769 & 0.3231          & 0.2308 & 0.2308 & \textbf{0.1692} & 0.2615 \\ \bottomrule
\end{tabular}%
}
\caption{Classification Error with a fixed threshold $t=0.3$.  The BoW results are based on the Type 2 inputs. The lowest (best) values within each group are bolded.}
\label{tab:classification_error_fixed}
\end{table}

\begin{table}[H]
\resizebox{\textwidth}{!}{%
\begin{tabular}{@{}ccccccccccc@{}}
\toprule
       & \multicolumn{5}{c}{LASSO}                             & \multicolumn{5}{c}{BART}                              \\ \midrule
       & BoW    & LDA    & LSA             & NMF    & Word2Vec & BoW    & LDA    & LSA             & NMF    & Word2Vec \\ \midrule
Type 1 & 0.1538 & 0.2308 & 0.2154          & 0.2000 & 0.2308   & 0.1077 & 0.2462 & 0.0923          & 0.2308 & 0.2308   \\
Type 2 & 0.1538 & 0.1077 & \textbf{0.0154} & 0.0923 & 0.2615   & 0.1077 & 0.0923 & \textbf{0.0615} & 0.0923 & 0.1385   \\
Type 3 &
  \textbf{0.0462} &
  \textbf{0.0000} &
  \textbf{0.0154} &
  \textbf{0.0000} &
  \textbf{0.0769} &
  \textbf{0.0462} &
  \textbf{0.0000} &
  \textbf{0.0615} &
  \textbf{0.0615} &
  \textbf{0.0923} \\ \bottomrule
\end{tabular}%
}
\caption{Classification Error for BoW embeddings with a fixed threshold $t = 0.3$}
\label{tab:classifcation_error_fixed_bow}
\end{table}

\subsection{Prediction for Disputed Papers}
\begin{table}[H]
\resizebox{\textwidth}{!}{%
\begin{tabular}{@{}cccccccccccccccc@{}}
\toprule
 &
  \multicolumn{3}{c}{BoW} &
  \multicolumn{3}{c}{LDA} &
  \multicolumn{3}{c}{LSA} &
  \multicolumn{3}{c}{NMF} &
  \multicolumn{3}{c}{Word2Vec} \\ \midrule
 &
  Type 1 &
  Type 2 &
  Type 3 &
  Type 1 &
  Type 2 &
  Type 3 &
  Type 1 &
  Type 2 &
  Type 3 &
  Type 1 &
  Type 2 &
  Type 3 &
  Type 1 &
  Type 2 &
  Type 3 \\ \midrule
No.49 &
  0.1372 &
  0.5299 &
  0.8024 &
  0.4135 &
  0.7453 &
  0.7514 &
  0.2123 &
  0.6286 &
  0.3384 &
  0.5252 &
  {\ul \textit{0.2977}} &
  0.5755 &
  {\ul \textit{0.2239}} &
  0.2126 &
  0.4920 \\
No.50 &
  0.1206 &
  0.1430 &
  0.5623 &
  {\ul \textit{0.1647}} &
  0.7291 &
  0.7767 &
  {\ul \textit{0.0498}} &
  0.5649 &
  0.4780 &
  0.3992 &
  0.3870 &
  0.3060 &
  0.2994 &
  0.3700 &
  0.4375 \\
No.51 &
  \textbf{0.4171} &
  0.7771 &
  0.8292 &
  0.2822 &
  {\ul \textit{0.7188}} &
  0.9217 &
  \textbf{0.5139} &
  0.8149 &
  0.4257 &
  \textbf{0.8007} &
  \textbf{0.8365} &
  0.5962 &
  \textbf{0.4109} &
  \textbf{0.5244} &
  0.4215 \\
No.52 &
  0.0630 &
  0.7024 &
  0.8131 &
  0.6648 &
  0.8778 &
  \textbf{0.9394} &
  0.1748 &
  0.7267 &
  0.3786 &
  0.3739 &
  0.6441 &
  0.6102 &
  0.3677 &
  0.3745 &
  0.5083 \\
No.53 &
  0.1866 &
  {\ul \textit{0.0490}} &
  0.9060 &
  0.6274 &
  0.8532 &
  0.7958 &
  0.2885 &
  0.5453 &
  0.4959 &
  0.3717 &
  0.7334 &
  0.6492 &
  0.3737 &
  0.4339 &
  0.4329 \\
No.54 &
  0.0464 &
  0.7244 &
  {\ul \textit{0.2566}} &
  \textbf{0.7368} &
  0.7883 &
  0.7886 &
  0.2786 &
  0.6839 &
  0.2909 &
  0.1695 &
  0.5415 &
  0.3981 &
  0.2993 &
  0.4885 &
  \textbf{0.6233} \\
No.55 &
  0.1663 &
  0.0492 &
  0.8330 &
  0.5579 &
  0.8262 &
  {\ul \textit{0.7504}} &
  0.2273 &
  0.3564 &
  {\ul \textit{0.2418}} &
  0.1770 &
  0.3948 &
  {\ul \textit{0.2061}} &
  0.2751 &
  0.3107 &
  {\ul \textit{0.1634}} \\
No.56 &
  {\ul \textit{0.0412}} &
  0.1446 &
  0.8475 &
  0.7009 &
  0.7656 &
  0.8525 &
  0.2774 &
  {\ul \textit{0.2501}} &
  0.3512 &
  0.2342 &
  0.5764 &
  0.4232 &
  0.3524 &
  0.4723 &
  0.3769 \\
No.57 &
  0.2210 &
  0.6958 &
  0.8856 &
  0.6989 &
  0.8914 &
  0.9370 &
  0.2421 &
  0.7573 &
  0.6410 &
  0.2771 &
  0.4970 &
  0.6583 &
  0.2934 &
  0.3204 &
  0.5278 \\
No.58 &
  0.0667 &
  0.6717 &
  0.8673 &
  0.7260 &
  0.7857 &
  0.9381 &
  0.4715 &
  0.6958 &
  0.4051 &
  0.3039 &
  0.4948 &
  0.3936 &
  0.2755 &
  0.3668 &
  0.4747 \\
No.62 &
  0.1297 &
  0.6926 &
  0.9161 &
  0.4330 &
  0.7441 &
  0.9383 &
  0.3482 &
  0.3476 &
  0.3864 &
  0.2032 &
  0.6492 &
  0.2763 &
  0.2603 &
  {\ul \textit{0.2099}} &
  0.5159 \\
No.63 &
  0.2181 &
  \textbf{0.7834} &
  \textbf{0.9368} &
  0.6829 &
  \textbf{0.8992} &
  0.8490 &
  0.3014 &
  \textbf{0.8828} &
  \textbf{0.7341} &
  {\ul \textit{0.1281}} &
  0.6416 &
  \textbf{0.7564} &
  0.3669 &
  0.4134 &
  0.5850 \\ \bottomrule
\end{tabular}%
}
\caption{Predicted probability by Bayesian Additive Regression Tree (BART) with different set of words as input. For each method, the highest predicted probability among the disputed papers is shown in bold, while the lowest is underlined and italicized. The probability being closer to 1 means that the paper is more likely to be authored by Madison.}
\label{tab:bart-predicted-prob-tdm-types}
\end{table}

\begin{table}[H]
\resizebox{\textwidth}{!}{%
\begin{tabular}{@{}cccccccccccc@{}}
\toprule
 &
  \multicolumn{4}{c}{BoW Embeddings} &
  \multicolumn{7}{c}{Continuous Embeddings} \\ \midrule
 &
  BoW &
  LDA &
  LSA &
  NMF &
  Word2Vec &
  BERT &
  RoBERTa &
  BART &
  GPT4 &
  Llama2 &
  Llama3 \\ \midrule
No.49 &
  0.5299 &
  0.7453 &
  0.6286 &
  {\ul \textit{0.2977}} &
  {\ul \textit{0.2126}} &
  0.1292 &
  0.5051 &
  0.3479 &
  0.3767 &
  0.3079 &
  0.2070 \\
No.50 &
  0.1430 &
  0.7291 &
  0.5649 &
  0.3870 &
  0.3700 &
  {\ul \textit{0.0918}} &
  0.5923 &
  0.0872 &
  0.3087 &
  0.1839 &
  {\ul \textit{0.1300}} \\
No.51 &
  0.7771 &
  {\ul \textit{0.7188}} &
  0.8149 &
  \textbf{0.8365} &
  \textbf{0.5244} &
  \textbf{0.5580} &
  0.5928 &
  \textbf{0.6907} &
  0.4241 &
  \textbf{0.5169} &
  \textbf{0.5193} \\
No.52 &
  0.7024 &
  0.8778 &
  0.7267 &
  0.6441 &
  0.3745 &
  0.3190 &
  0.6232 &
  0.3520 &
  0.1793 &
  0.3421 &
  0.2760 \\
No.53 &
  {\ul \textit{0.0490}} &
  0.8532 &
  0.5453 &
  0.7334 &
  0.4339 &
  0.1513 &
  0.5353 &
  0.2880 &
  0.1977 &
  {\ul \textit{0.1836}} &
  0.1710 \\
No.54 &
  0.7244 &
  0.7883 &
  0.6839 &
  0.5415 &
  0.4885 &
  0.2854 &
  0.5361 &
  0.2827 &
  {\ul \textit{0.0775}} &
  0.2910 &
  0.2645 \\
No.55 &
  0.0492 &
  0.8262 &
  0.3564 &
  0.3948 &
  0.3107 &
  0.1451 &
  0.3976 &
  0.2787 &
  0.1926 &
  0.3533 &
  0.2741 \\
No.56 &
  0.1446 &
  0.7656 &
  {\ul \textit{0.2501}} &
  0.5764 &
  0.4723 &
  0.1735 &
  0.5768 &
  {\ul \textit{0.0848}} &
  0.1140 &
  0.2515 &
  0.1963 \\
No.57 &
  0.6958 &
  0.8914 &
  0.7573 &
  0.4970 &
  0.3204 &
  0.1727 &
  0.3424 &
  0.5873 &
  0.1256 &
  0.2625 &
  0.1847 \\
No.58 &
  0.6717 &
  0.7857 &
  0.6958 &
  0.4948 &
  0.3668 &
  0.1567 &
  \textbf{0.6677} &
  0.3238 &
  0.0962 &
  0.2053 &
  0.2806 \\
No.62 &
  0.6926 &
  0.7441 &
  0.3476 &
  0.6492 &
  {\ul \textit{0.2099}} &
  0.3364 &
  0.3732 &
  0.4023 &
  0.1248 &
  0.3552 &
  0.2660 \\
No.63 &
  \textbf{0.7834} &
  \textbf{0.8992} &
  \textbf{0.8828} &
  0.6416 &
  0.4134 &
  0.2775 &
  {\ul \textit{0.2763}} &
  0.4156 &
  \textbf{0.4275} &
  0.2653 &
  0.1964 \\ \bottomrule
\end{tabular}%
}
\caption{Predicted probability by Bayesian Additive Regression Tree (BART). The outcome being 1 means the paper is written by Madison. For each method, the highest support toward the Madison is marked as a bold letter, and the lowest support is underlined. BoW, LDA, SVD and GPT 4 give the strongest support to No.63 toward Madison; NMF, Word2Vec, BERT, BART, Llama2 and Llama 3 show the strongest support to No.51 toward Madison. Recall that, in the original study, No. 51 and No. 63 are the papers with the strongest support with the Bayesian model. On the other hand, the papers with the weakest support are rather inconsistent. The papers with the weakest support in the original study were No. 55 and No. 56. All the models except LDA, relatively low support for Madison's authorship is observed, though not the lowest. }
\label{tab:predicted_prob_bart}
\end{table}

\begin{table}[H]
\resizebox{\textwidth}{!}{%
\begin{tabular}{@{}cccccccccccccccc@{}}
\toprule
 &
  \multicolumn{3}{c}{BoW} &
  \multicolumn{3}{c}{LDA} &
  \multicolumn{3}{c}{LSA} &
  \multicolumn{3}{c}{NMF} &
  \multicolumn{3}{c}{Word2Vec} \\ \midrule
 &
  Type 1 &
  Type 2 &
  Type 3 &
  Type 1 &
  Type 2 &
  Type 3 &
  Type 1 &
  Type 2 &
  Type 3 &
  Type 1 &
  Type 2 &
  Type 3 &
  Type 1 &
  Type 2 &
  Type 3 \\ \midrule
No.18 &
  \textbf{0.1821} &
  0.6375 &
  0.7495 &
  \textbf{0.0628} &
  \textbf{0.2489} &
  0.6294 &
  0.0395 &
  \textbf{0.6272} &
  \textbf{0.7722} &
  \textbf{0.3032} &
  \textbf{0.6360} &
  \textbf{0.8733} &
  {\ul \textit{0.2578}} &
  0.3565 &
  \textbf{0.5335} \\
No.19 &
  0.1707 &
  \textbf{0.7328} &
  \textbf{0.9210} &
  0.0322 &
  {\ul \textit{0.1156}} &
  \textbf{0.7698} &
  {\ul \textit{0.0297}} &
  {\ul \textit{0.5497}} &
  0.6599 &
  0.2854 &
  0.5704 &
  0.7796 &
  \textbf{0.3259} &
  \textbf{0.4185} &
  0.4889 \\
No.20 &
  {\ul \textit{0.1345}} &
  {\ul \textit{0.5228}} &
  {\ul \textit{0.7046}} &
  {\ul \textit{0.0297}} &
  0.1549 &
  {\ul \textit{0.5232}} &
  \textbf{0.0570} &
  0.5544 &
  {\ul \textit{0.6049}} &
  {\ul \textit{0.2505}} &
  {\ul \textit{0.4602}} &
  {\ul \textit{0.5849}} &
  0.2824 &
  {\ul \textit{0.3334}} &
  {\ul \textit{0.4368}} \\ \bottomrule
\end{tabular}%
}
\caption{Predicted probabilities from BART classifier for jointly authored papers for BoW embeddings with different inputs.}
\label{tab:predicted-prob-joint-tdms}
\end{table}

\begin{table}[H]
\resizebox{\textwidth}{!}{%
\begin{tabular}{@{}cccccccccccc@{}}
\toprule
 & \multicolumn{4}{c}{BoW Embeddings} & \multicolumn{7}{c}{Continuous Embeddings}                 \\ \midrule
 & BoW     & LDA    & LSA    & NMF    & Word2Vec & BERT & RoBERTa & BART & GPT4 & Llama2 & Llama3 \\ \midrule
LASSO & 0.3712                  & \textbf{0.0000}         & 0.0874                  & 0.1546                  & 0.5669                       & 0.1385 & \textbf{0.1077} & 0.4154 & 0.5692 & 0.7692          & 0.7692 \\
BART  & 0.3376                  & \textbf{0.0542}         & 0.1925                  & 0.2174                  & 0.4001                       & 0.1538 & 0.3846          & 0.3538 & 0.2615 & \textbf{0.1385} & 0.3077 \\ \bottomrule
\end{tabular}
}
\caption{Assuming the attribution by \cite{Mosteller1963Inference} as ground truth to compute $\ell_2$ loss. Note that we use the same threshold computed by LOOCV result and the exact values are presented in Table~\ref{tab:threshold}. For BoW methods, Type 2 inputs are used to compute the value. LDA succeeds in attributing all the papers to Madison.  }
\label{tab:predicted-l2-disputed-papers}
\end{table}

\section{Experiments with Fine-Tuning}\label{sec:fine-tuning}

\begin{figure}[H]
    \begin{subfigure}{.49\textwidth}
        \centering
        \includegraphics[width=.95\linewidth]{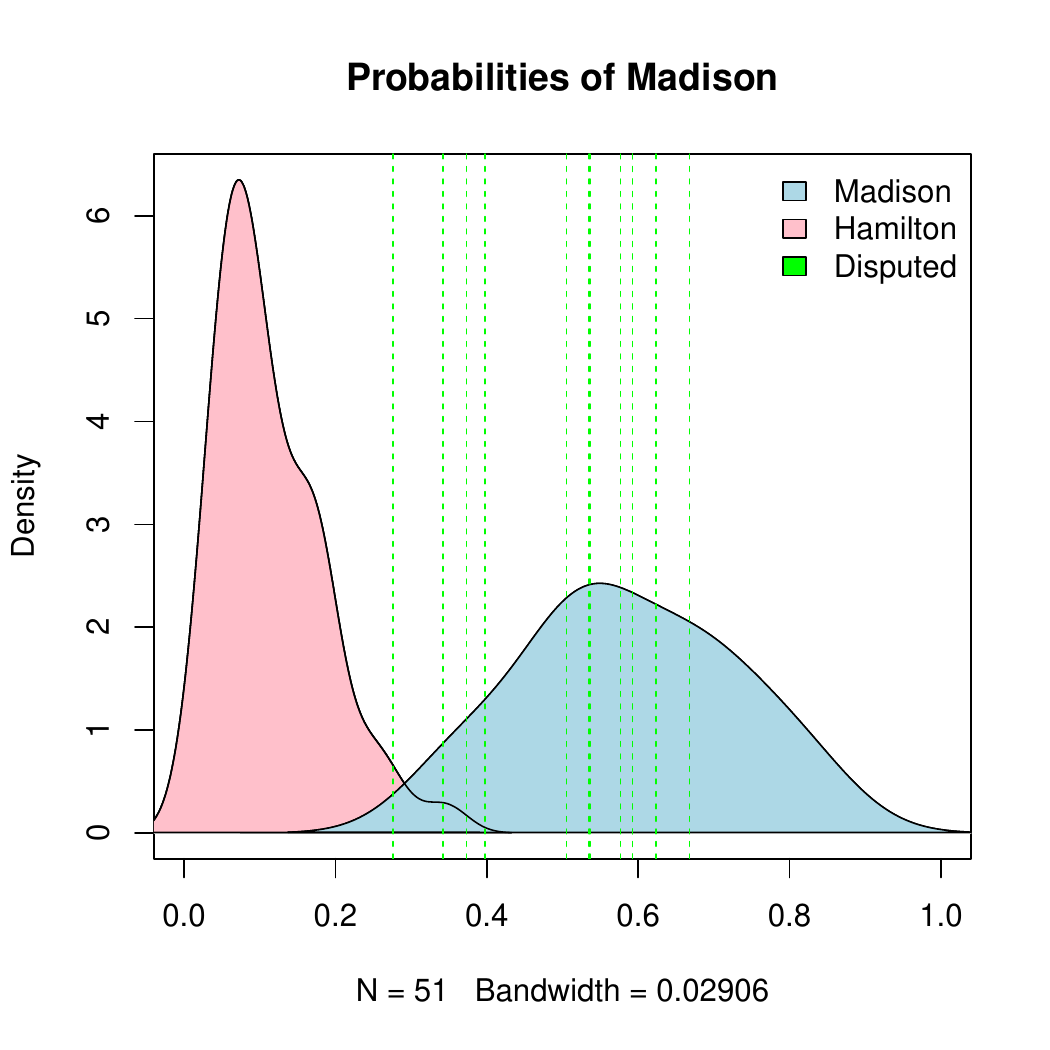}
        \caption{Vanilla embedding}
    \end{subfigure}
    \begin{subfigure}{.49\textwidth}
        \centering
        \includegraphics[width=.95\linewidth]{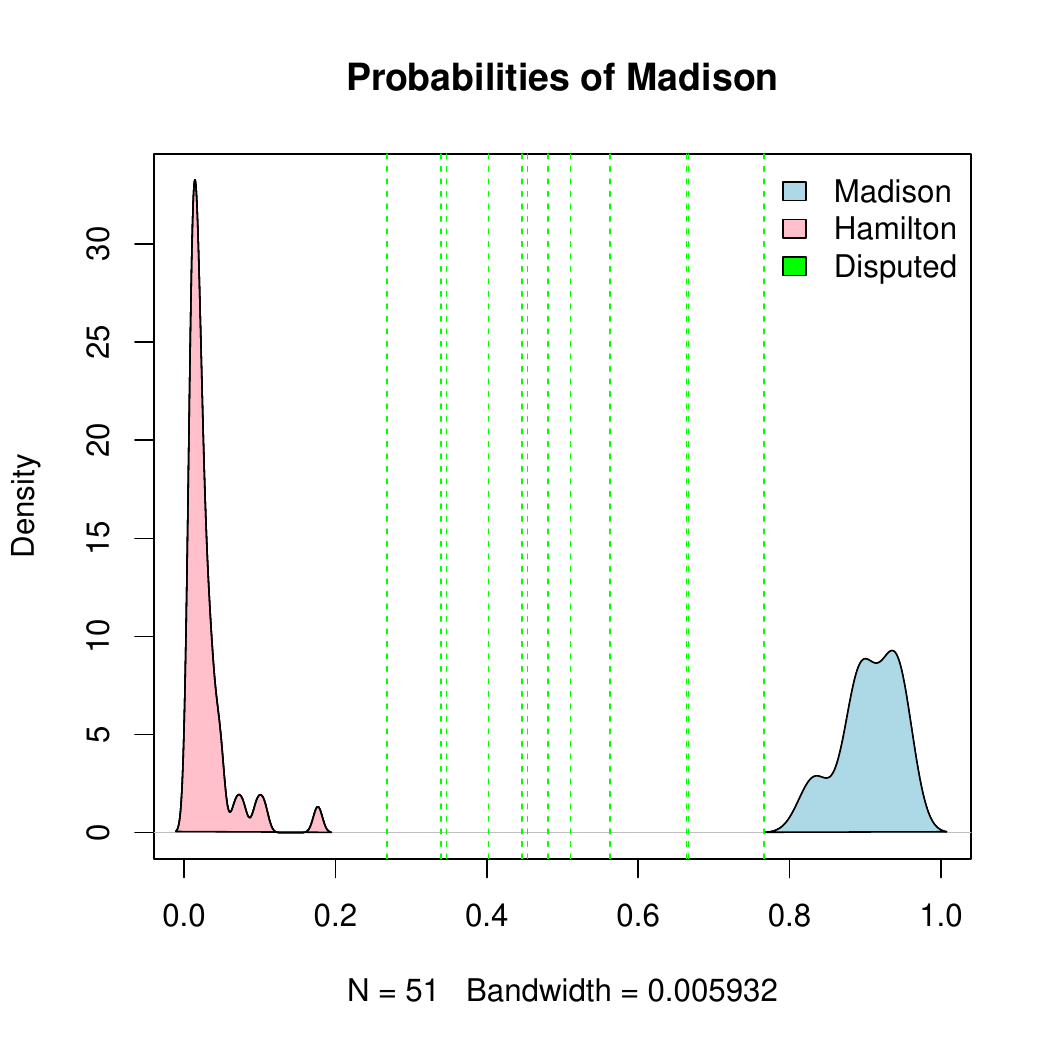}
        \caption{Fine-tuned embedding through classification}
    \end{subfigure}
    \caption{Estimated density for Hamilton and Madison by BART using RoBERTa embeddings. Again, as in Figure~\ref{fig:bert-with-finetuning}, the separation of densities on training dataset becomes more obvious; however, it is not at all informative for the unseen dataset (green vertical lines). }
    \label{fig:roberta-with-finetuning}
\end{figure}

\begin{figure}[H]
    \begin{subfigure}{.49\textwidth}
        \centering
        \includegraphics[width=.95\linewidth]{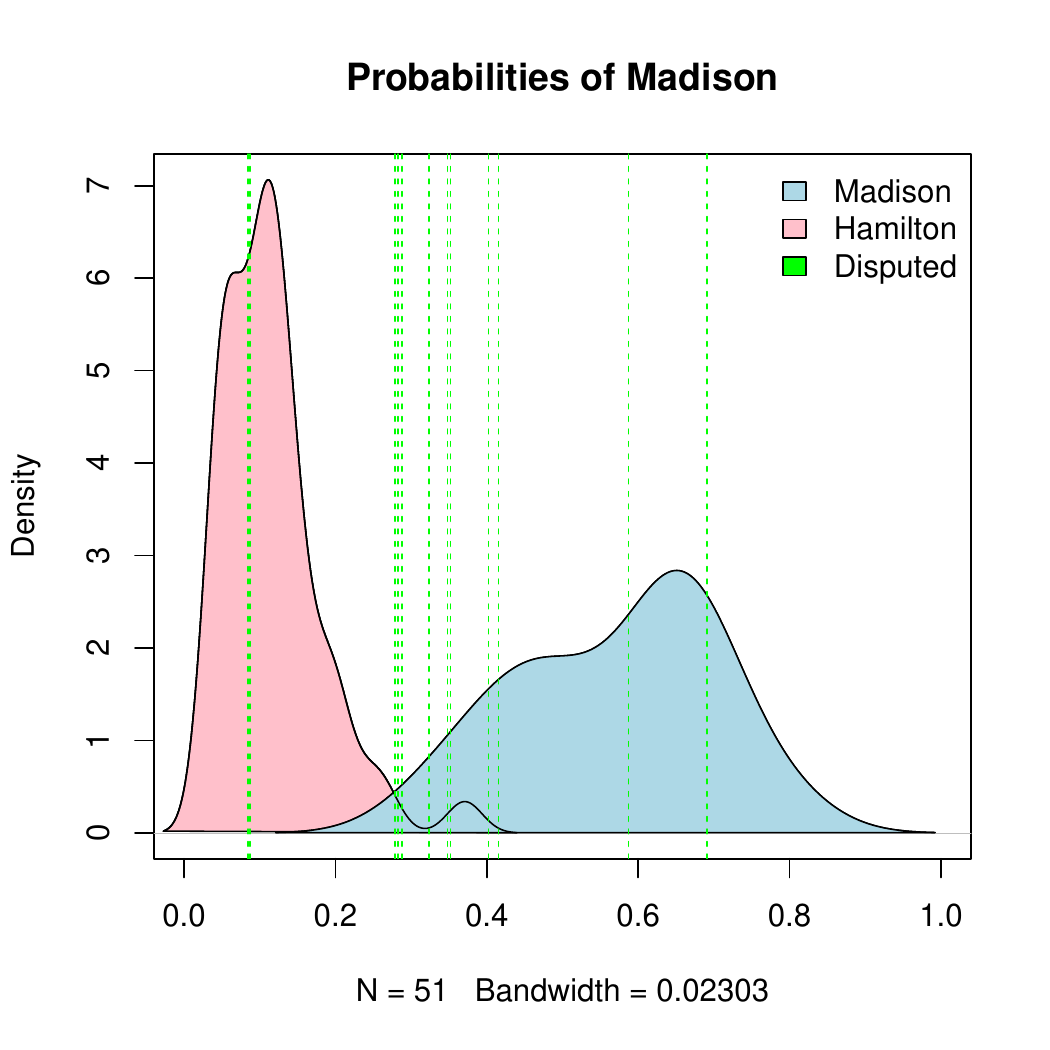}
        \caption{Vanilla embedding}
    \end{subfigure}
    \begin{subfigure}{.49\textwidth}
        \centering
        \includegraphics[width=.95\linewidth]{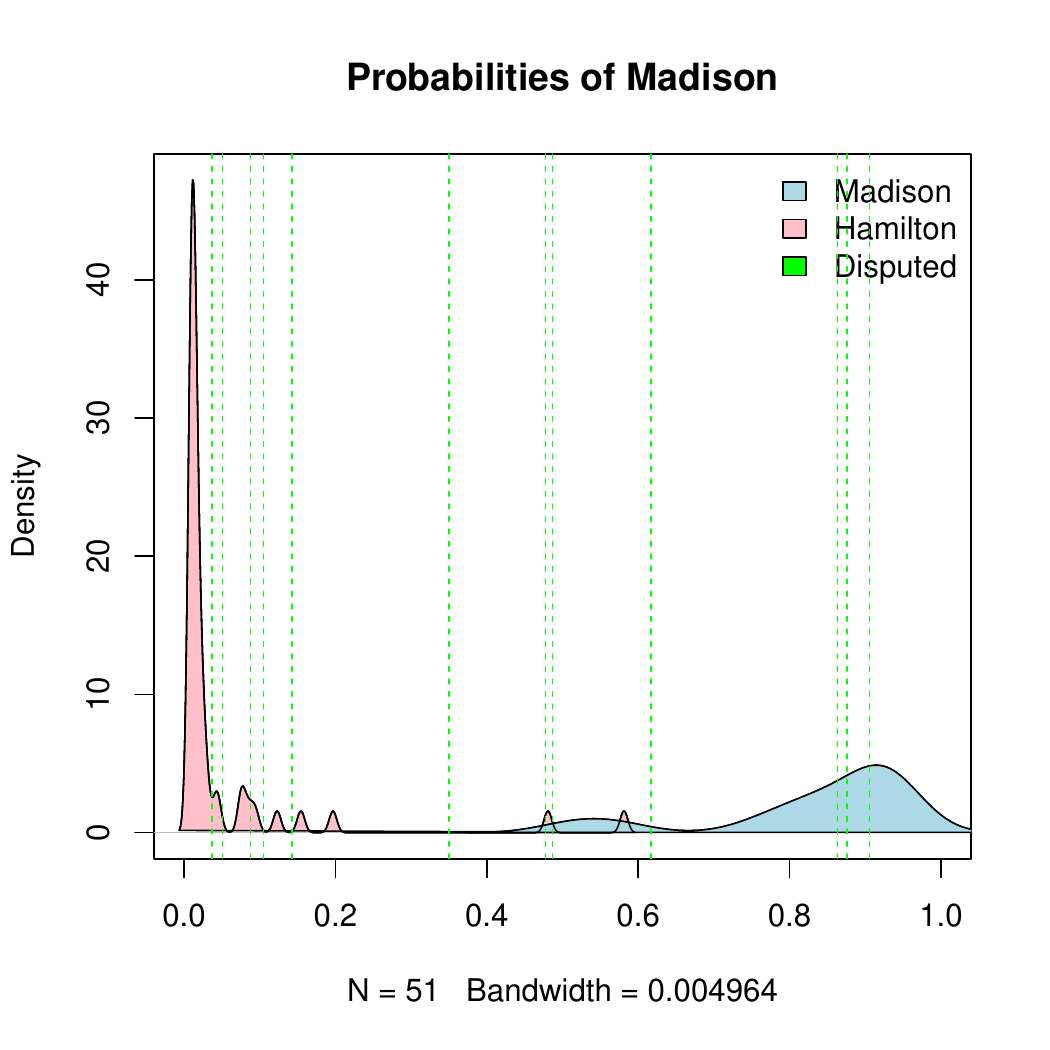}
        \caption{Fine-tuned embedding through classification}
    \end{subfigure}
    \caption{Estimated density for Hamilton and Madison using BART \citep{Chipman2010bart} with embeddings from BART \citep{lewis2019bart}. The assignment of authorship in the test set differs from that of BERT (Figure~\ref{fig:bert-with-finetuning}) and RoBERTa (Figure~\ref{fig:roberta-with-finetuning}). Prior to tuning, only one disputed paper is confidently assigned to Hamilton, with most falling in the overlap region. After tuning, five papers are confidently assigned to Hamilton, but this assignment is likely incorrect, suggesting a potential decline in prediction accuracy.}
    \label{fig:bart-with-finetuning}
\end{figure}

\begin{table}[H]
\centering
\resizebox{0.65\textwidth}{!}{%
\begin{tabular}{@{}ccccccc@{}}
\toprule
      & \multicolumn{3}{c}{Vanilla   Embedding} & \multicolumn{3}{c}{Fine-Tune   Embedding} \\ \midrule
      & BERT        & RoBERTa      & BART       & BERT         & RoBERTa      & BART        \\ \midrule
LASSO & 0.5620      & 0.1172       & 0.5228     & 0.4154       & 0.3701       & 0.2491      \\
BART  & 0.6038      & 0.2626       & 0.4566     & 0.3368       & 0.2784       & 0.4454      \\ \bottomrule
\end{tabular}%
}
\caption{$\ell_2$ loss computed for 12 disputed papers on vanilla and fine-tuned embeddings. The loss value is inflated for RoBERTa with both classifiers. The loss values are slightly improved (decreased) for BERT or BART, but the gain is not significant. Even after the fine-tuning, the performance is as good as the one based on the pure word count matrix (Table~\ref{tab:predicted-l2-disputed-papers}).}
\label{tab:l2-fine-tune-test}
\end{table}

\begin{table}[H]
\centering
\resizebox{0.5\textwidth}{!}{%
\begin{tabular}{@{}ccccc@{}}
\toprule
        & \multicolumn{2}{c}{LASSO} & \multicolumn{2}{c}{BART} \\ \midrule
        & ROC         & F1          & ROC         & F1         \\ \midrule
BERT    & 0.4965      & 0.9610      & 0.4681      & 0.7961     \\
RoBERTa & 0.4823      & 0.9154      & 0.5323      & 0.7390     \\
BART    & 0.3851      & 0.3940      & 0.3577      & 0.3870     \\ \bottomrule
\end{tabular}%
}
\caption{Classification thresholds computed by ROC and F1 rule for fine-tuned embeddings. Overall, the threshold values are inflated compared to the threshold computed from vanilla embeddings (Table~\ref{tab:threshold}). }
\label{tab:threshold-fine-tune}
\end{table}

\begin{table}[]
\centering
\resizebox{0.7\textwidth}{!}{%
\begin{tabular}{@{}cccccccc@{}}
\toprule
                         &       & \multicolumn{3}{c}{Vanilla   Embedding} & \multicolumn{3}{c}{Fine-Tune   Embedding} \\ \midrule
                         &       & BERT        & RoBERTa      & BART       & BERT         & RoBERTa      & BART        \\ \midrule
\multirow{2}{*}{t = 0.3} & LASSO & 0.7500      & 0.1667       & 0.5833     & 0.5000       & 0.2500       & 0.0833      \\
                         & BART  & 0.7500      & 0.0833       & 0.4167     & 0.0833       & 0.0833       & 0.4167      \\
\multirow{2}{*}{ROC}     & LASSO & 0.0833      & 0.0000       & 0.0833     & 0.8333       & 0.6667       & 0.1667      \\
                         & BART  & 0.4167      & 0.0000       & 0.1667     & 0.6667       & 0.6667       & 0.5000      \\
\multirow{2}{*}{F1}    & LASSO & 0.5000      & 0.0833       & 0.0833     & 1.0000       & 1.0000       & 0.1667      \\
                         & BART  & 0.9167      & 0.0000       & 0.1667     & 1.0000       & 0.9167       & 0.5000      \\ \bottomrule
\end{tabular}%
}
\caption{Classification error for 12 disputed papers on vanilla and fine-tuned embeddings. The performance for RoBERTa was descent before the fine-tuning, but the classification error even becomes 1 (with LASSO and $F_1$ threshold) after the fine-tuning. For BERT and BART, the classification error decreases though the improvement is marginal. }
\label{tab:clf-fine-tune-test}
\end{table}

\section{List of Function/Stop Words used in the Analysis}\label{app-words}

\begin{table}[H]
\resizebox{\textwidth}{!}{%
\begin{tabular}{|c|c|c|c|c|c|c|c|c|}
\hline
i       & me         & my       & myself     & we      & our    & ours     & ourselves & you     \\ \hline
your    & yours      & yourself & yourselves & he      & him    & his      & himself   & she     \\ \hline
her     & hers       & herself  & it         & its     & itself & they     & them      & their   \\ \hline
theirs  & themselves & what     & which      & who     & whom   & this     & that      & these   \\ \hline
those   & am         & is       & are        & was     & were   & be       & been      & being   \\ \hline
have    & has        & had      & having     & do      & does   & did      & doing     & would   \\ \hline
should  & could      & ought    & i'm        & you're  & he's   & she's    & it's      & we're   \\ \hline
they're & i've       & you've   & we've      & they've & i'd    & you'd    & he'd      & she'd   \\ \hline
we'd    & they'd     & i'll     & you'll     & he'll   & she'll & we'll    & they'll   & isn't   \\ \hline
aren't  & wasn't     & weren't  & hasn't     & haven't & hadn't & doesn't  & don't     & didn't  \\ \hline
won't   & wouldn't   & shan't   & shouldn't  & can't   & cannot & couldn't & mustn't   & let's   \\ \hline
that's  & who's      & what's   & here's     & there's & when's & where's  & why's     & how's   \\ \hline
a       & an         & the      & and        & but     & if     & or       & because   & as      \\ \hline
until   & while      & of       & at         & by      & for    & with     & about     & against \\ \hline
between & into       & through  & during     & before  & after  & above    & below     & to      \\ \hline
from    & up         & down     & in         & out     & on     & off      & over      & under   \\ \hline
again   & further    & then     & once       & here    & there  & when     & where     & why     \\ \hline
how     & all        & any      & both       & each    & few    & more     & most      & other   \\ \hline
some    & such       & no       & nor        & not     & only   & own      & same      & so      \\ \hline
than    & too        & very     &            &         &        &          &           &         \\ \hline
\end{tabular}%
}
\caption{Stop words in R `tm' package}
\label{tab:stopwords-tm}
\end{table}

\subsection{Selected Words by Mosteller(1963)}


\begin{table}[H]
\resizebox{\textwidth}{!}{%
\begin{tabular}{|c|c|c|c|c|c|c|c|c|}
\hline
a      & as    & do   & has   & is    & no    & or   & than & this  \\ \hline
when   & all   & at   & down  & have  & it    & not  & our  & that  \\ \hline
to     & which & also & be    & even  & her   & its  & now  & shall \\ \hline
the    & up    & who  & an    & been  & every & his  & may  & of    \\ \hline
should & their & upon & will  & and   & but   & for  & if   & more  \\ \hline
on     & so    & then & was   & with  & any   & by   & from & in    \\ \hline
must   & one   & some & there & were  & would & are  & can  & had   \\ \hline
into   & my    & only & such  & thing & what  & your &      &       \\ \hline
\end{tabular}%
}
\caption{Function words}
\label{tab:function-words}
\end{table}

\begin{table}[H]
\resizebox{\textwidth}{!}{%
\begin{tabular}{|c|c|c|c|c|c|c|c|c|}
\hline
affect      & city         & direction   & innovation   & perhaps      & vigor       & again     & commonly   & disgracing \\ \hline
join        & rapid        & violate     & although     & consequently & either      & language  & same      & violence   \\ \hline
among       & considerable & enough      & most         & second       & voice       & another   & contribute & nor        \\ \hline
still       & where        & because     & defensive    & fortune      & offensive   & those     & whether    & between    \\ \hline
destruction & function     & often       & throughout   & while        & both        & did       & himself    & pass       \\ \hline
under       & whilst       & about       & choice       & proper       & according   & common    & kind       & propriety  \\ \hline
adversaries & danger       & large       & provision    & after        & decide      & decides   & decided    & deciding   \\ \hline
likely      & requisite   & aid         & degree       & matters      & matter      & substance & always     & during     \\ \hline
moreover    & they         & apt         & expence      & expences     & necessary   & though    & asserted   & expenses   \\ \hline
expense     & necessity    & necessities & truth        & truths       & before      & extent    & others     & us         \\ \hline
being       & follows      & follow      & particularly & usages       & usage       & better    & I          & principle  \\ \hline
we          & care         & imagine     & edit         & editing      & probability & work      &            &            \\ \hline
\end{tabular}%
}
\caption{Additional Set of Words in \cite{Mosteller1963Inference}}
\label{tab:my-table}
\end{table}

\section{Discussion: What Defines LLMs}\label{sec:llm-def}

The definition of Large Language Models (LLMs) is indeed a topic of divergence in the literature, with different researchers focusing on various aspects of model architecture, parameter count, and the scale of data used for training. For instance, as mentioned by \cite{zhao2023survey}, LLMs are typically defined by three key components: (1) the use of Transformer architecture, (2) the presence of hundreds of billions (or more) parameters, and (3) training on massive datasets. This definition focuses on the scale and architecture of models like GPT-3, PaLM, and LlaMA, highlighting the power of attention mechanisms and massive parameterization as core distinguishing features. 

However, other works take a narrower or more focused view of what qualifies as an LLM. For example, \cite{lu2024smalllanguagemodelssurvey} argue that encoder-only transformer models, like BERT, are not classified as LLMs under their framework, even though these models can be highly parameterized. This exclusion is based on the idea that LLMs are primarily defined by their capacity for generating human-like text, which models like BERT, trained for tasks like classification and sentence encoding, do not directly perform. Thus, this definition limits LLMs to models designed primarily for generative tasks, rather than those for representation learning or other specialized applications.

Other perspectives emphasize the role of scalability and generalization rather than just size. For instance, \cite{bommasani2022opportunities} note that LLMs are defined not only by their massive scale in terms of data and parameters but also by their general-purpose capability across a wide range of tasks, often with minimal fine-tuning. In this view, models like GPT-3 are LLMs not just because they are large but because they exhibit emergent properties like few-shot learning and can generalize across domains, whereas smaller models or task-specific models, even if large, may not exhibit these behaviors.


\end{document}